\documentclass[lettersize,journal]{IEEEtran}
\usepackage{amsmath,amsfonts}
\usepackage{algorithmic}
\usepackage{algorithm}
\usepackage{array}
\usepackage[caption=false,font=normalsize,labelfont=sf,textfont=sf]{subfig}
\usepackage{textcomp}
\usepackage{stfloats}
\usepackage{url}
\usepackage{verbatim}
\usepackage{graphicx}
\usepackage{cite}
\usepackage{color}
\usepackage{amssymb}
\usepackage{mathrsfs}
\usepackage{varwidth}
\usepackage{multirow}
\usepackage{arydshln}
\usepackage{booktabs}
\usepackage{xcolor}

\usepackage[breaklinks,colorlinks]{hyperref}

\begin{document}

\title{Underwater Organism Color Enhancement via Color Code Decomposition, Adaptation and Interpolation}

\author{Xiaofeng Cong, Jing Zhang,~\IEEEmembership{Senior Member},~IEEE, Yeying Jin, Junming Hou, Yu Zhao, Jie Gui,~\IEEEmembership{Senior Member},~IEEE,  James Tin-Yau Kwok,~\IEEEmembership{Fellow},~IEEE, Yuan Yan Tang,~\IEEEmembership{Life Fellow},~IEEE
        \thanks{
                 (X. Cong and J. Zhang contributed equally to this work.) (Corresponding author: J. Gui.)

                 X. Cong and Y. Zhao are with the School of Cyber Science and Engineering, Southeast University, Nanjing 210000, China (e-mail: cxf\_svip@163.com, zyzzustc@gmail.com).

                 J. Zhang is with the School of Computer Science, The University of Sydney, Camperdown, NSW 2050, Australia (e-mail:
                 jing.zhang1@sydney.edu.au).

                 Y. Jin is with Department of Electrical and Computer Engineering (ECE), National University of Singapore (NUS), Singapore 119260 (e-mail: e0178303@u.nus.edu).

                 J. Hou is with the State Key Laboratory of Millimeter Waves, School of Information Science and Engineering, Southeast University, Nanjing 210096, China (e-mails: junming\_hou@seu.edu.cn).

                J. Gui is with the School of Cyber Science and Engineering, Southeast University and with Purple Mountain Laboratories, Nanjing 210000, China (e-mail: guijie@seu.edu.cn).

                J. Kwok is with the Department of Computer Science and Engineering, The Hong Kong University of Science and Technology, Hong Kong 999077, China. (e-mail: jamesk@cse.ust.hk).

                Y. Tang is with the Department of Computer and Information Science, University of Macau, Macau 999078, China (e-mail: yytang@um.edu.mo).

}
}

\markboth{Journal of \LaTeX\ Class Files,~Vol.~14, No.~8, August~2021}%
{Shell \MakeLowercase{\textit{et al.}}: A Sample Article Using IEEEtran.cls for IEEE Journals}

\makeatletter
\def\ps@IEEEtitlepagestyle{
  \def\@oddfoot{\mycopyrightnotice}
  \def\@evenfoot{}
}
\def\mycopyrightnotice{
  {\footnotesize
  \begin{minipage}{\textwidth}
  \centering
  Copyright~\copyright~20xx IEEE. Personal use of this material is permitted. \\ 
   However, permission to use this material for any other purposes must be obtained from the IEEE by sending a request to pubs-permissions@ieee.org.
  \end{minipage}
  }
}

\maketitle

\begin{abstract}
        Underwater images often suffer from quality degradation due to absorption and scattering effects. Most existing underwater image enhancement algorithms produce a single, fixed-color image, limiting user flexibility and application. To address this limitation, we propose a method called \textit{ColorCode}, which enhances underwater images while offering a range of controllable color outputs. Our approach involves recovering an underwater image to a reference enhanced image through supervised training and decomposing it into color and content codes via self-reconstruction and cross-reconstruction. The color code is explicitly constrained to follow a Gaussian distribution, allowing for efficient sampling and interpolation during inference. ColorCode offers three key features: 1) color enhancement, producing an enhanced image with a fixed color; 2) color adaptation, enabling controllable adjustments of long-wavelength color components using guidance images; and 3) color interpolation, allowing for the smooth generation of multiple colors through continuous sampling of the color code. Quantitative and visual evaluations on popular and challenging benchmark datasets demonstrate the superiority of ColorCode over existing methods in providing diverse, controllable, and color-realistic enhancement results. The source code is available at \href{https://github.com/Xiaofeng-life/ColorCode}{ColorCode}.
\end{abstract}


\section{Introduction}
\label{sec:introduction}
\begin{figure}
        \centering
        \includegraphics[width=8.8cm,height=7.6cm]{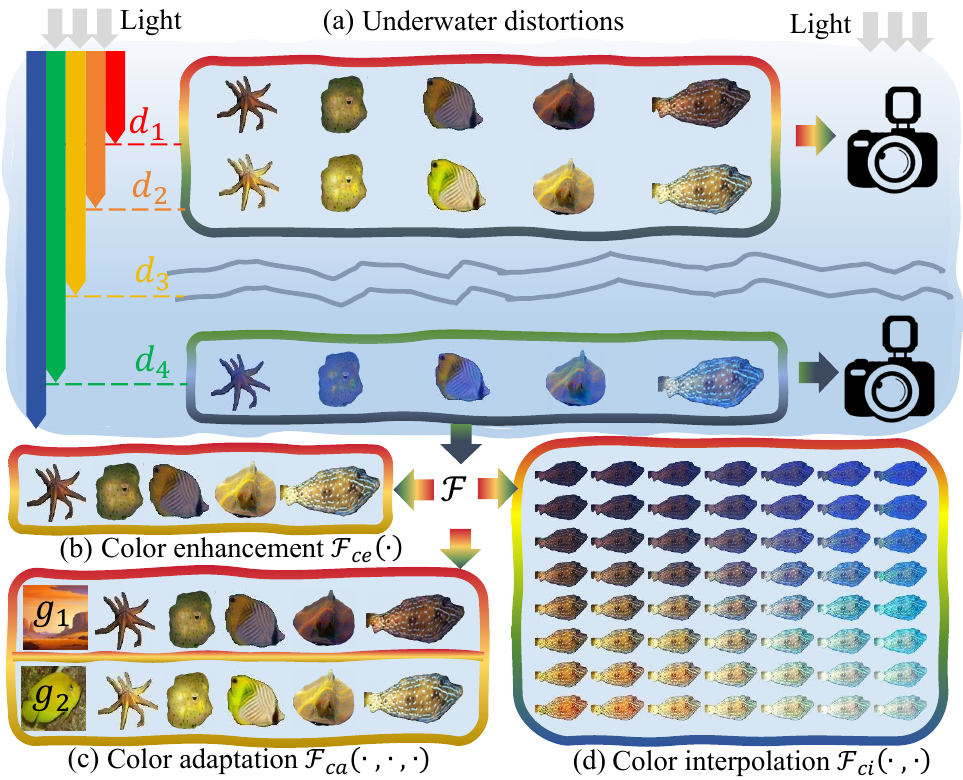}
        \caption{
        As the water depth increases from $d_{1}$ to $d_{4}$, light of various wavelengths gradually diminishes, causing underwater organisms to shift toward blue-green or blue hues, which appear dull, as shown in (a). To address this, we propose ColorCode, which offers three key features: (b) color enhancement $\mathcal{F}_{ce}(\cdot)$, (c) color adaptation $\mathcal{F}_{ca}(\cdot, \cdot, \cdot)$ using natural guidance $g_{1}$ or underwater guidance $g_{2}$, and (d) color interpolation $\mathcal{F}_{ci}(\cdot, \cdot)$ through sampling different color codes.
        }
        \label{fig:motivation_of_this_paper}
\end{figure}

\IEEEPARstart{U}{nderwater} images often suffer from color distortion due to the absorption and scattering of light \cite{wang2022semantic,fabbri2018enhancing,liu2019underwater,gonzalez2024dgd,esmaeilzehi2024dmml}. As light penetrates water, different wavelengths attenuate at varying rates—red and yellow wavelengths fade first, while green and blue persist longer. Consequently, underwater images typically appear blue-green or blue \cite{li2019underwater,qi2022sguie,huang2023contrastive}, as illustrated in Fig.~\ref{fig:motivation_of_this_paper}(a). This attenuation diminishes the vibrant colors of underwater organisms, reducing the aesthetic richness of their appearance \cite{yin2024unsupervised,khan2024spectroformer,wang2024multi,wang2017deep}.

To enhance the quality of distorted underwater images, numerous underwater image enhancement (UIE) algorithms have been developed. These algorithms can be categorized into two types based on their approach: deep learning-based UIE (DL-UIE) \cite{yan2022attention,islam2020fast,chen2021underwater,fu2022underwater,naik2021shallow} and non-deep learning-based UIE (NDL-UIE) \cite{li2016single,peng2017underwater,carlevaris2010initial}. NDL-UIE algorithms often rely on various prior assumptions, physical models, and digital image processing techniques, including IBLA \cite{peng2017underwater}, MIP \cite{carlevaris2010initial}, UDCP \cite{drews2016underwater}, ULAP \cite{song2018rapid}, SMBLOTMOP \cite{song2020enhancement}, MLLE \cite{zhang2022underwater}, and HLRP \cite{zhuang2022underwater}. However, in complex degraded scenes, NDL-UIE methods may produce results with color shifts \cite{cong2024underwater}.

To address the limitations of NDL-UIE, several advanced DL-UIE algorithms have been developed. These include networks that utilize various computational operations, such as naive convolution (UWNet \cite{naik2021shallow}), attention mechanisms (ADMNNet \cite{yan2022attention}), Transformer modules (U-Trans \cite{peng2023u}), Fourier transformations (UHD \cite{wei2022uhd}), and Wavelet decomposition (UIE-WD \cite{ma2022wavelet}). Additionally, different learning strategies have been explored, including adversarial learning (FUnIEGAN \cite{islam2020fast}), rank learning (URanker \cite{guo2023underwater}), contrastive learning (Semi-UIR \cite{huang2023contrastive}), and reinforcement learning (HPUIE-RL \cite{song2024hierarchical}). Auxiliary information like semantics (SGUIE \cite{qi2022sguie}) and depth (Joint-ID \cite{yang2023joint}) have also been used to enhance UIE model performance. Furthermore, various studies have explored interesting issues, such as embedding physical models (USUIR \cite{fu2022unsupervised}, GUPDM \cite{mu2023generalized}), applying different color spaces (UGIF-Net \cite{zhou2023ugif}, TCTL-Net \cite{li2023tctl}), and distinguishing water types (SCNet \cite{fu2022underwater}, DAL \cite{uplavikar2019all}). These efforts have led to impressive results in both full-reference and no-reference evaluation metrics.

Recent studies \cite{chen2022domain,kim2021pixel,cong2024underwater} highlight that, \textit{beyond improving quantitative metrics, offering users a variety of enhanced outputs is also advantageous for UIE tasks}. This is particularly important because the reference images used by UIE algorithms are not flawless \cite{cong2024underwater}. Even in the absence of mud or suspended particles, water molecules inevitably cause color shifts in underwater images. Thus, the images we collect are always affected by these shifts.

To date, three UIE algorithms capable of producing diversified outputs have been developed. UIESS \cite{chen2022domain} utilizes domain adversarial training to separate images into content and style spaces. By manipulating the latent space, UIESS can generate enhancement outputs at varying levels. PWAE \cite{kim2021pixel} creates a latent space that represents style, enabling the generation of style-diverse results by selecting guidance images with different styles. In the preliminary work \cite{cong2024underwater}, we proposed CECF, which focuses on color diversity rather than style, treating water as the background and underwater organisms as the foreground. CECF uses guidance with long-wavelength colors to adjust the colors of these organisms, compensating for the lack of such colors in the water itself. However, the latent spaces in UIESS, PWAE, and CECF do not allow for continuous semantic results through sampling and interpolation. As illustrated in Fig.~\ref{fig:motivation_of_this_paper}, our proposed method, ColorCode, simultaneously achieves color enhancement, color adaptation, and color interpolation:

\begin{figure}
        \small
        \centering

        \includegraphics[width=4.3cm,height=4.3cm]{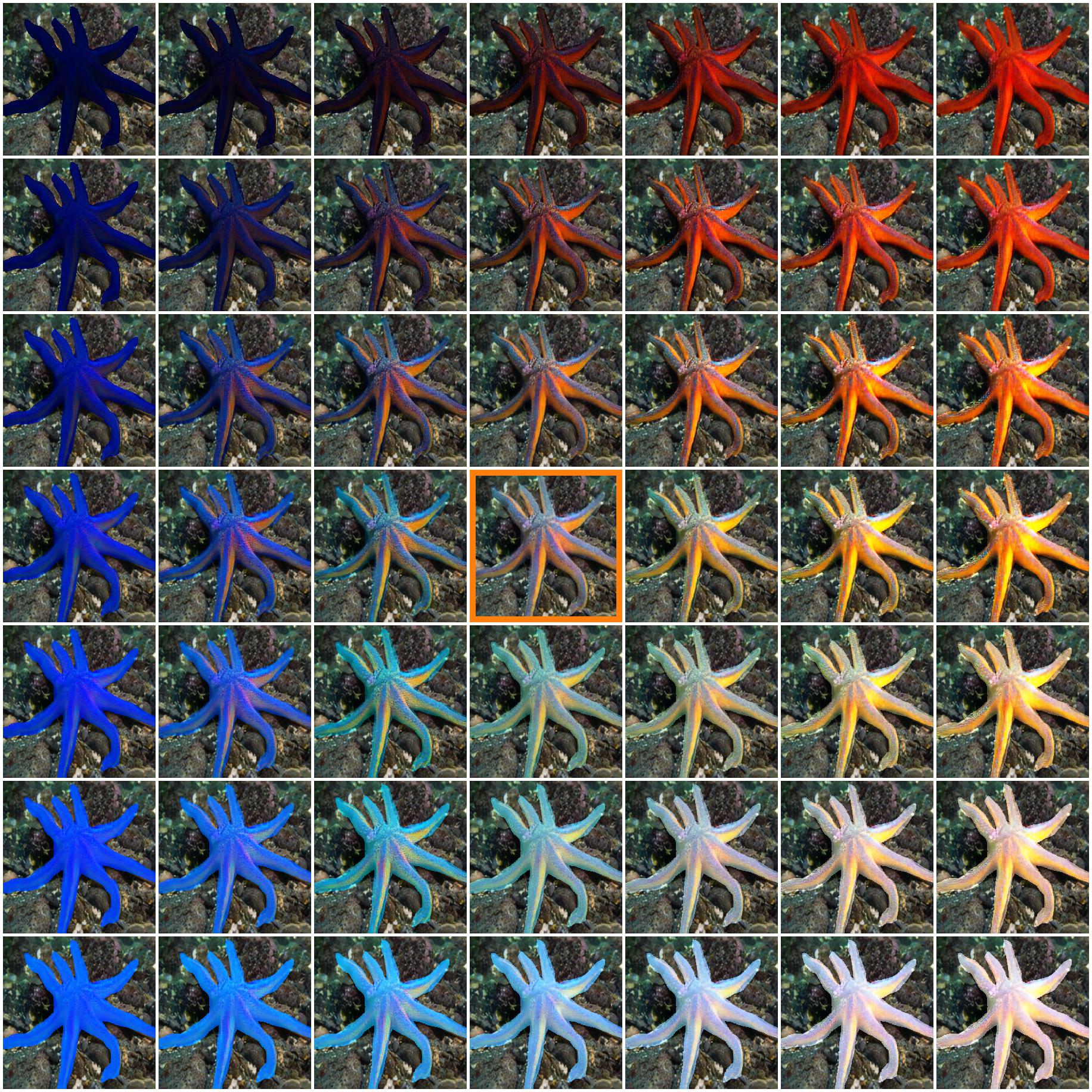}
        \includegraphics[width=4.3cm,height=4.3cm]{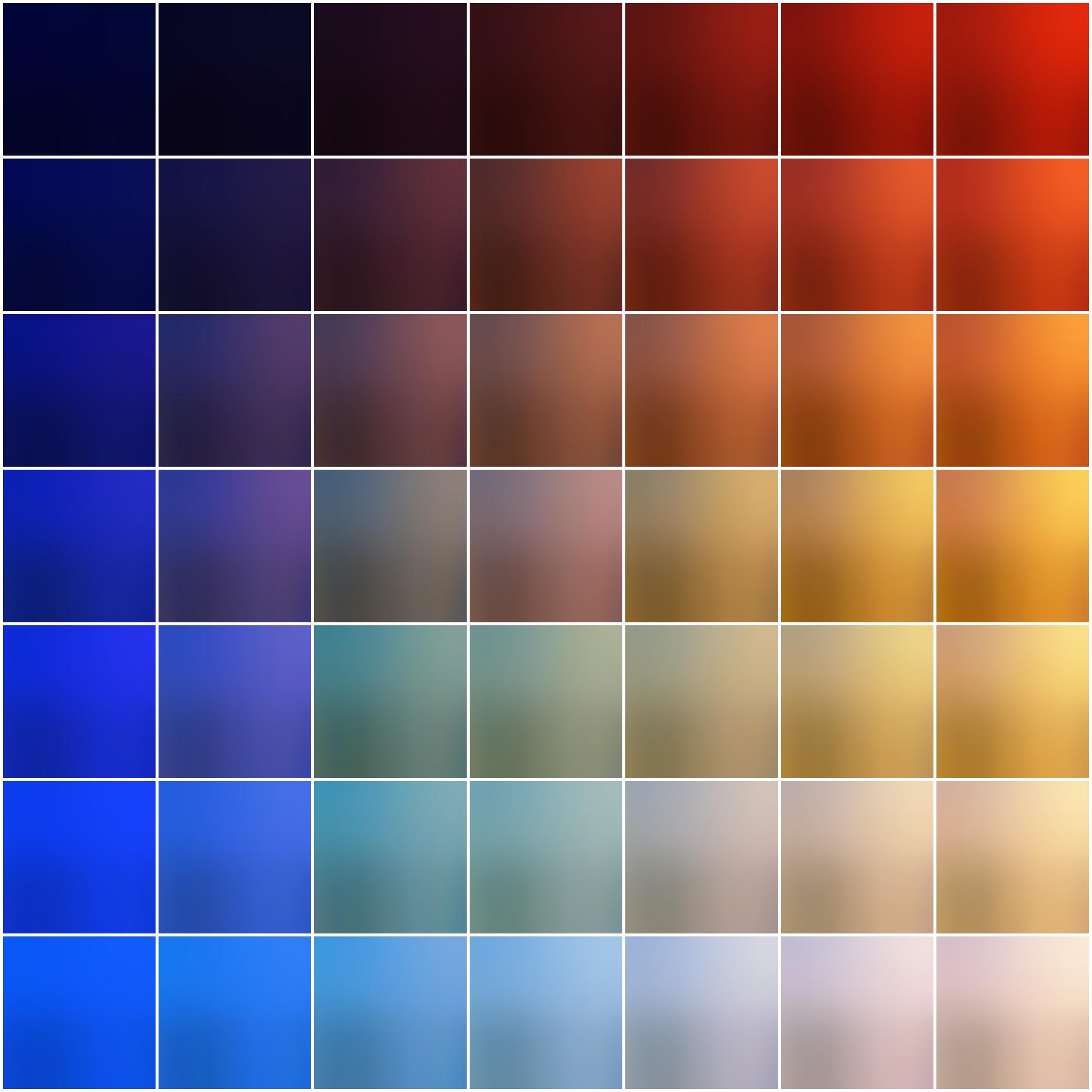}
        {(a) The interpolation result obtained by \textbf{style} code.}
        \vspace{0.03cm} 

        \includegraphics[width=4.3cm,height=4.3cm]{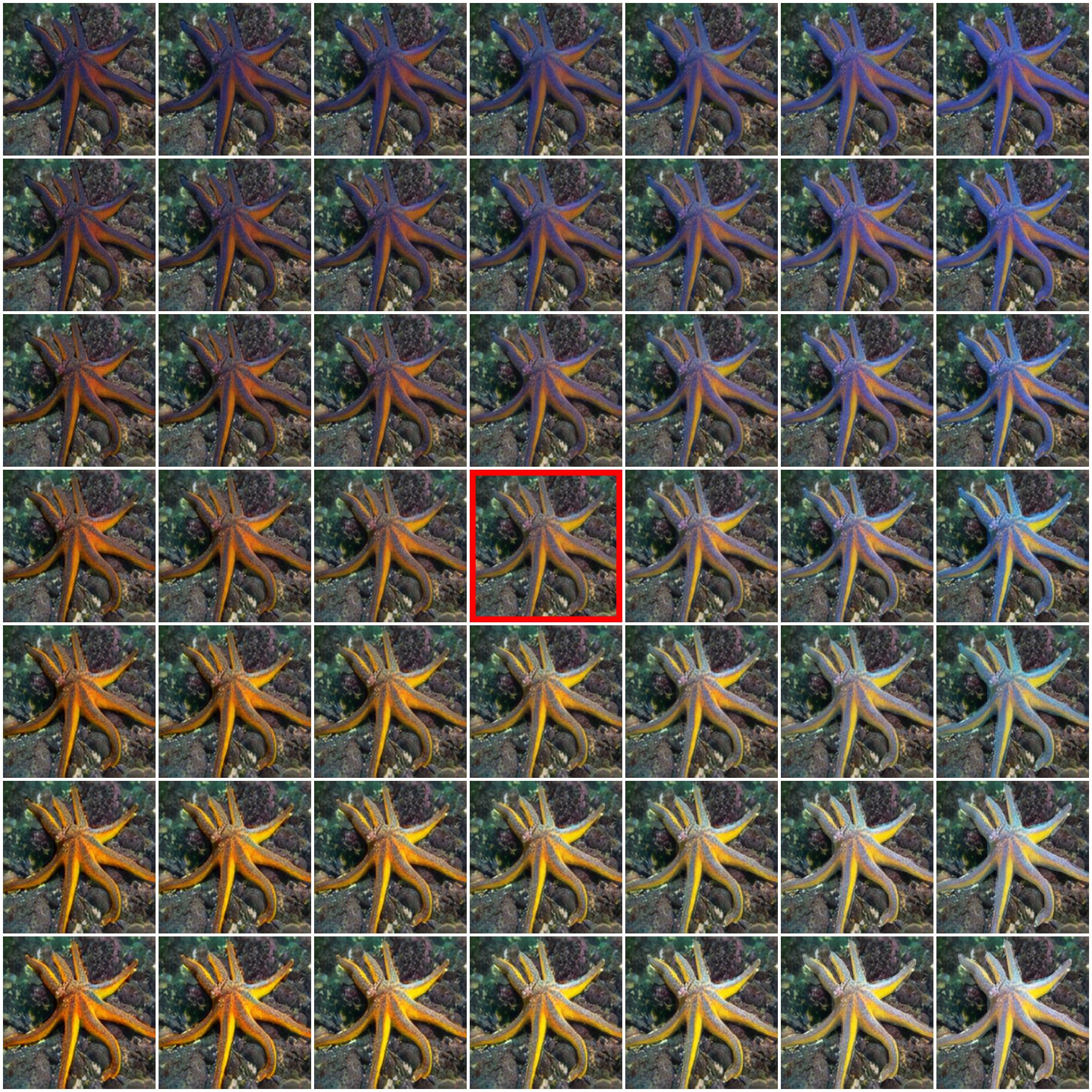}
        \includegraphics[width=4.3cm,height=4.3cm]{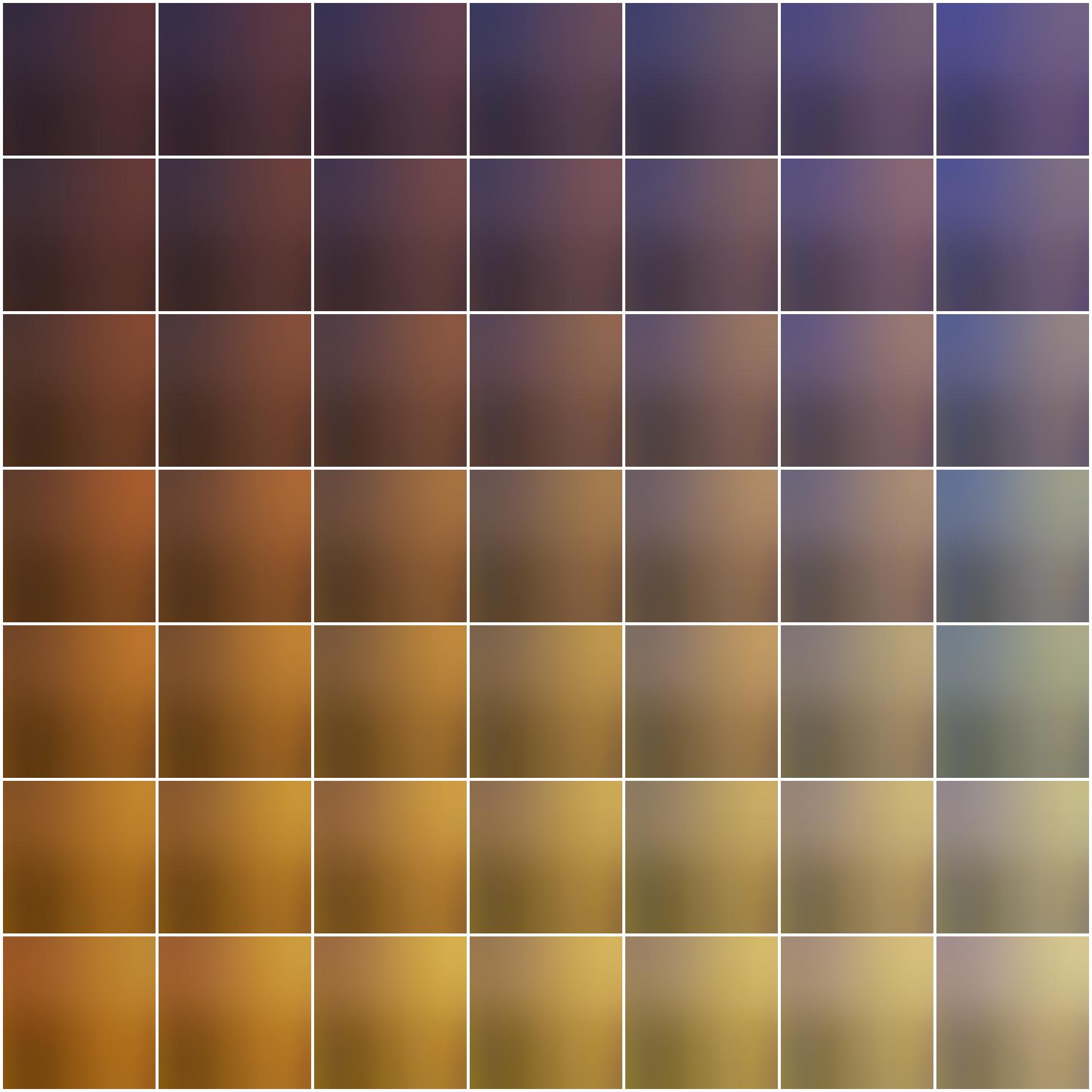}
        {(b) The interpolation result obtained by \textbf{color} code.}
        \vspace{0.03cm} 

        \includegraphics[width=2.1cm,height=2.1cm]{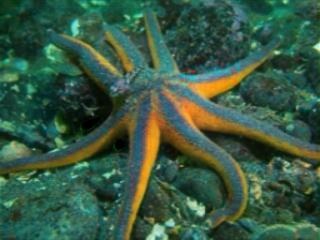}
        \includegraphics[width=2.1cm,height=2.1cm]{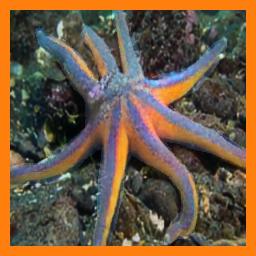}
        \includegraphics[width=2.1cm,height=2.1cm]{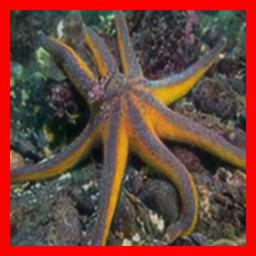}
        \includegraphics[width=2.1cm,height=2.1cm]{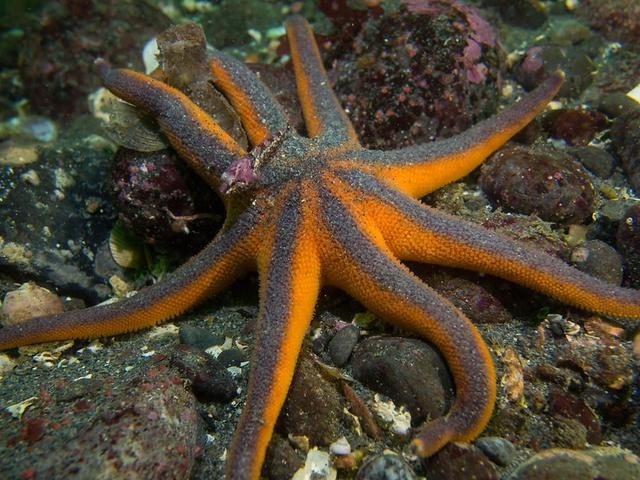}
        \leftline{\hspace{0.4cm} Distortion \hspace{0.7cm} \textbf{style} code \hspace{0.6cm} \textbf{color} code \hspace{0.7cm} Reference}
        {(c) Comparison of the center of interpolated colors.}

        \caption{Comparisons of results obtained by style code and color code. The color blocks represent the main colors of the organisms in the  images. The acquisition of color block is achieved by calculating the center of mass of the organism and the surrounding main color.}
        \label{fig:visualization_of_main_color_map_of_color_code_and_style_code}
\end{figure}

\begin{figure*}
        \small
        \centering
        \leftline{\hspace{1cm} $x_{1}$  \hspace{1.5cm}   $y_{1}$  \hspace{1.4cm} $p-x_{1}$  \hspace{1.3cm} $p-y_{1}$ 
        \hspace{1.7cm} $x_{2}$ \hspace{1.5cm}  $y_{2}$  \hspace{1.5cm} $p-x_{2}$  \hspace{1.1cm}  $p-y_{2}$}
        \includegraphics[width=2.1cm,height=1.9cm]{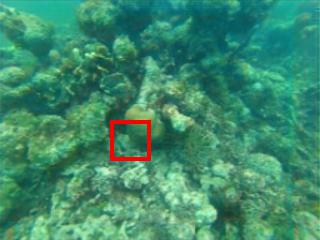}
        \includegraphics[width=2.1cm,height=1.9cm]{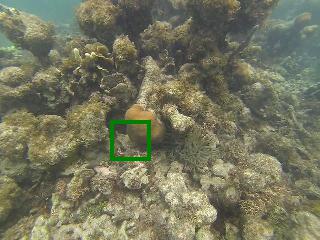}
        \includegraphics[width=2.1cm,height=1.9cm]{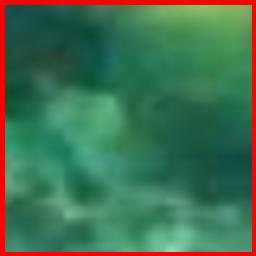}
        \includegraphics[width=2.1cm,height=1.9cm]{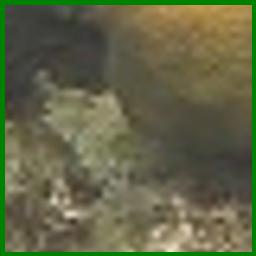}
        \includegraphics[width=0.2cm,height=1.9cm]{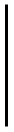}
        \includegraphics[width=2.1cm,height=1.9cm]{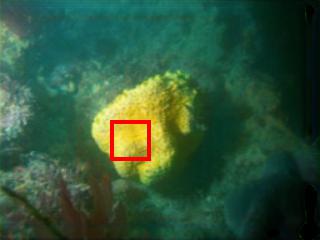}
        \includegraphics[width=2.1cm,height=1.9cm]{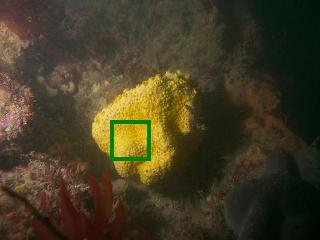}
        \includegraphics[width=2.1cm,height=1.9cm]{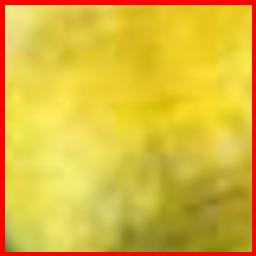}
        \includegraphics[width=2.1cm,height=1.9cm]{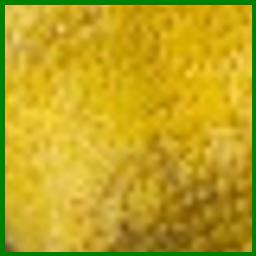}
        
        \vspace{0.1cm}

        \includegraphics[width=2.1cm,height=1.9cm]{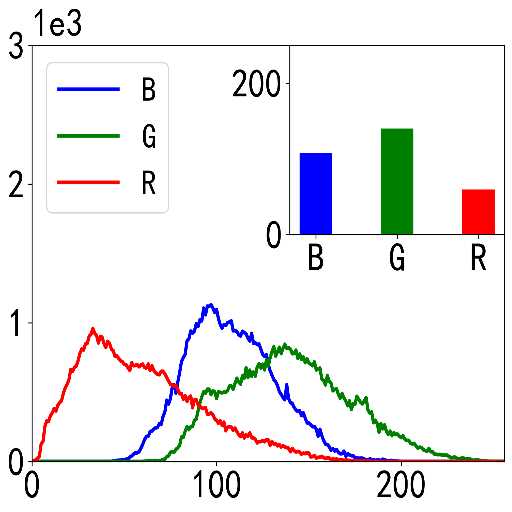}
        \includegraphics[width=2.1cm,height=1.9cm]{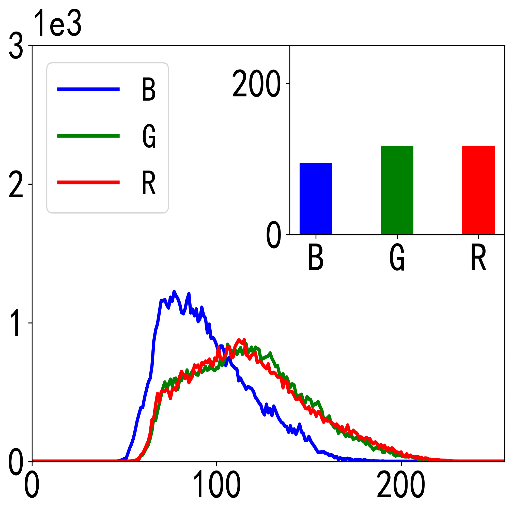}
        \includegraphics[width=2.1cm,height=1.9cm]{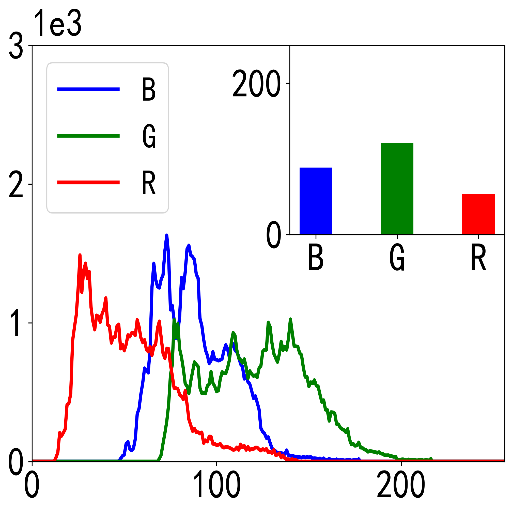}
        \includegraphics[width=2.1cm,height=1.9cm]{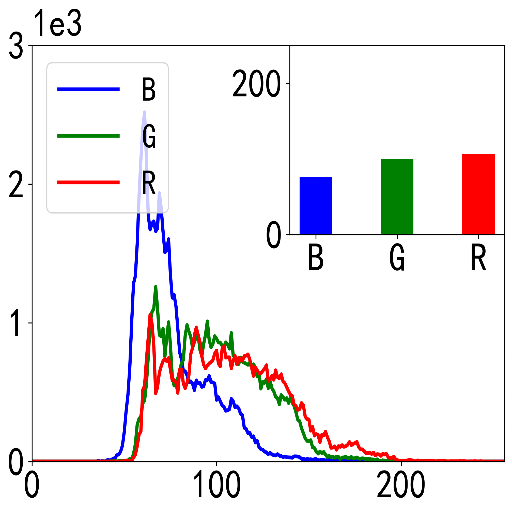}
        \includegraphics[width=0.2cm,height=1.9cm]{00_figures/materials/black_vertical_line.PNG}
        \includegraphics[width=2.1cm,height=1.9cm]{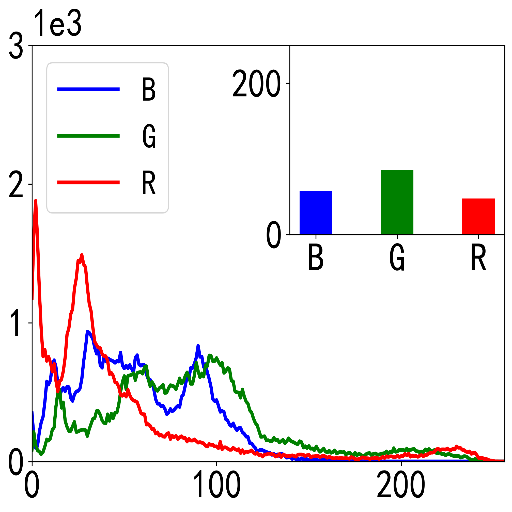}
        \includegraphics[width=2.1cm,height=1.9cm]{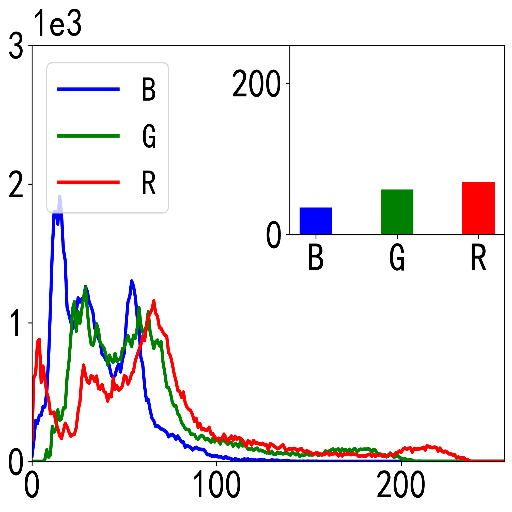}
        \includegraphics[width=2.1cm,height=1.9cm]{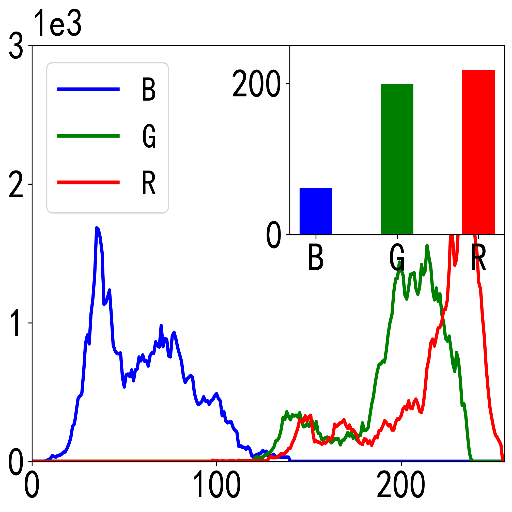}
        \includegraphics[width=2.1cm,height=1.9cm]{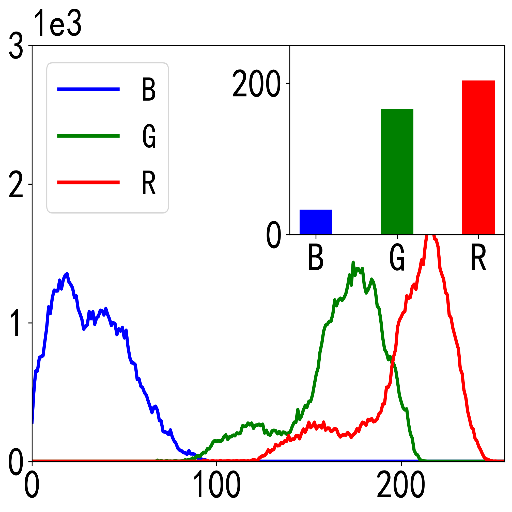}
        
        \leftline{\hspace{3cm} (a) Global distortion. \hspace{6cm} (b) Non-global distortion.}

        \caption{Observation of approximately invariance of hue of long-wavelength colors during enhancement. $x_1$ and $x_2$ represent two underwater images with different distortion levels, where $y_1$ and $y_2$ mean the corresponding reference images. $p-$ denotes the enlarged patch. The curve figure stands for the pixel histogram, while the bar figure shows the average pixel value of the red (R), green (G) and blue (B) channel. The images in (a) and (b) are from \cite{islam2020fast}. In (a), there is no long-wavelength color. In (b), the long-wavelength color (yellow) is preserved during the enhancement from $x_{2}$ to $y_{2}$.}
        \label{fig:illustration_of_the_invariant_of_the_hue_of_long_wavelength_colors}
\end{figure*}

\begin{itemize}
        \item \textbf{Color Enhancement (CE)}: For a distorted underwater image, this technique provides a deterministic enhanced image with reliable quantitative results, as demonstrated in Fig.~\ref{fig:motivation_of_this_paper}(b).
        \item \textbf{Color Adaptation (CA)}: Enhanced underwater images can be adjusted to match specific color guidance. For instance, the adaptation degree for images with long-wavelength colors can be modified through color code fusion, as shown in Fig.~\ref{fig:motivation_of_this_paper}(c).
        \item \textbf{Color Interpolation (CI)}: This process enables the generation of diverse colors by sampling different color codes, as illustrated in Fig.~\ref{fig:motivation_of_this_paper}(d).
\end{itemize}

\begin{figure*}
        \small
        \centering
        \includegraphics[width=2.9cm,height=1.5cm]{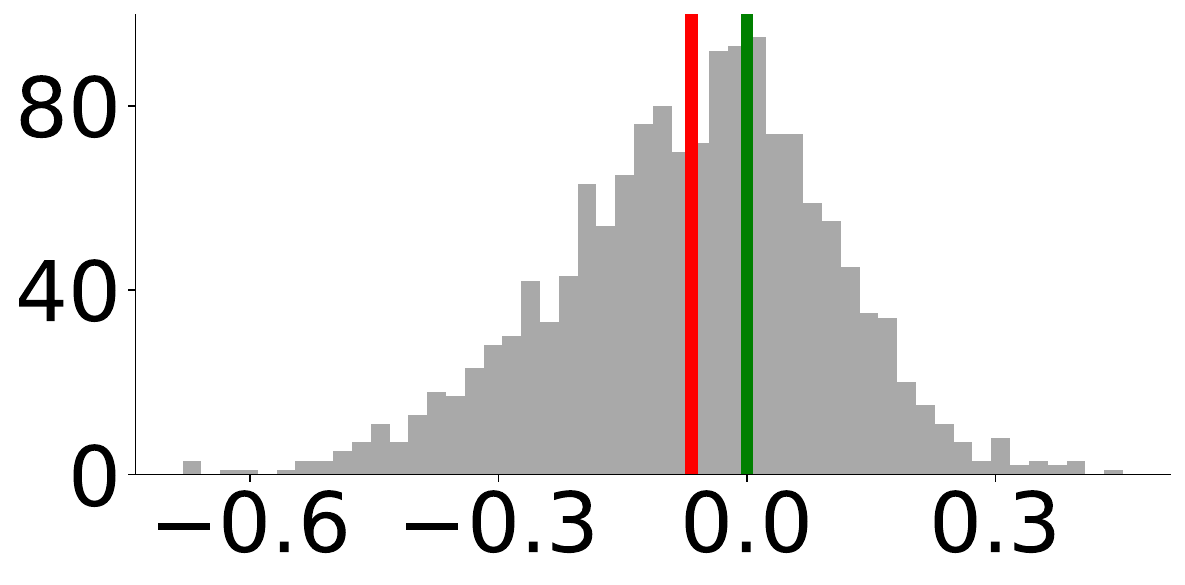}
        \includegraphics[width=2.9cm,height=1.5cm]{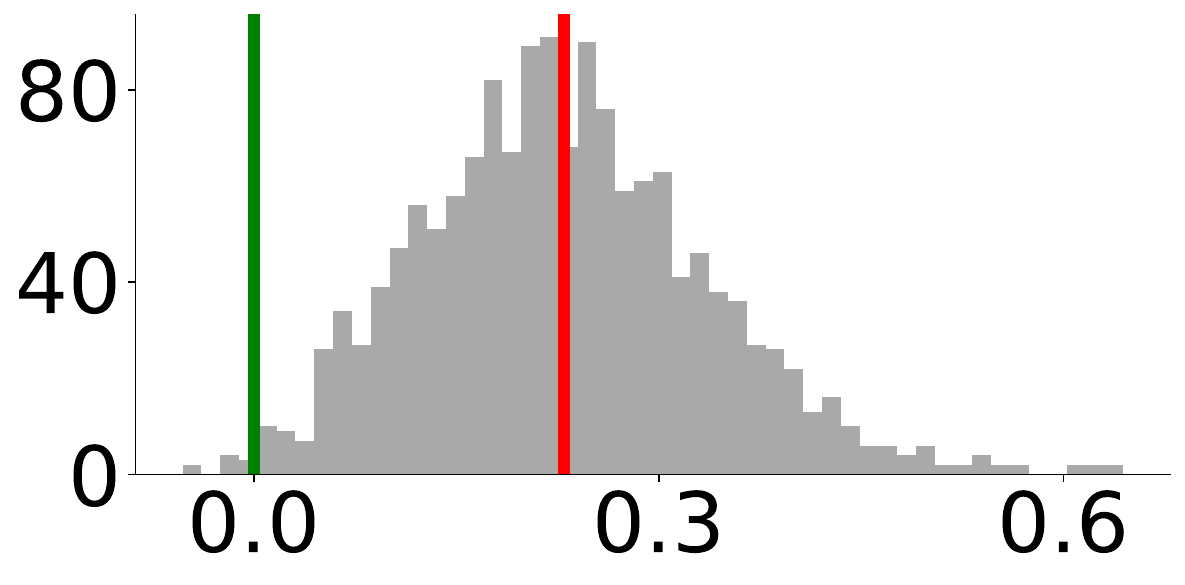}
        \includegraphics[width=2.9cm,height=1.5cm]{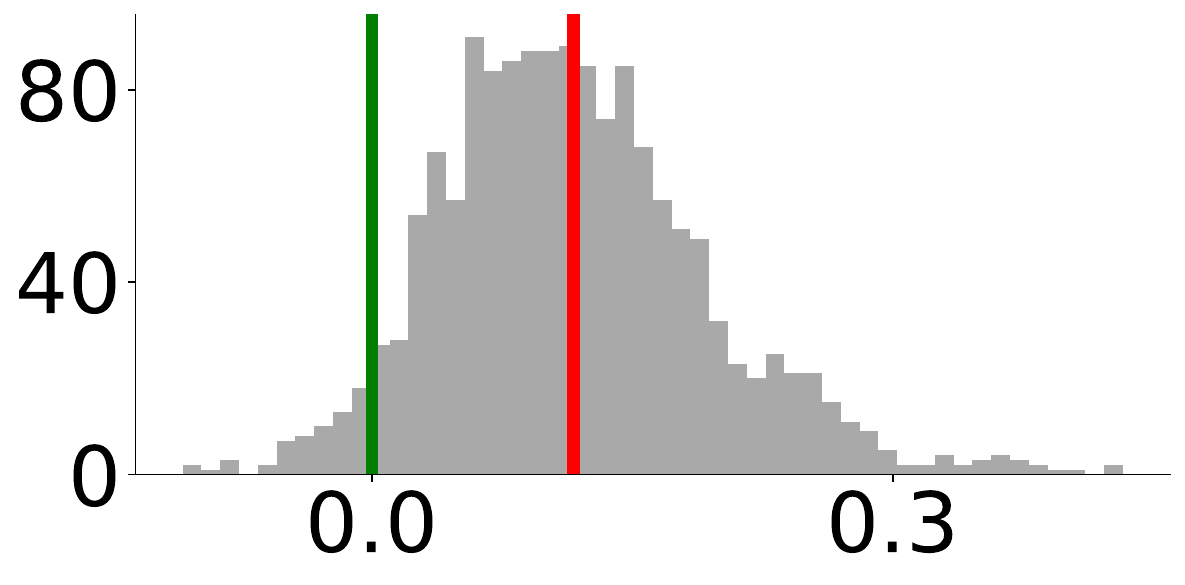}
        \includegraphics[width=2.9cm,height=1.5cm]{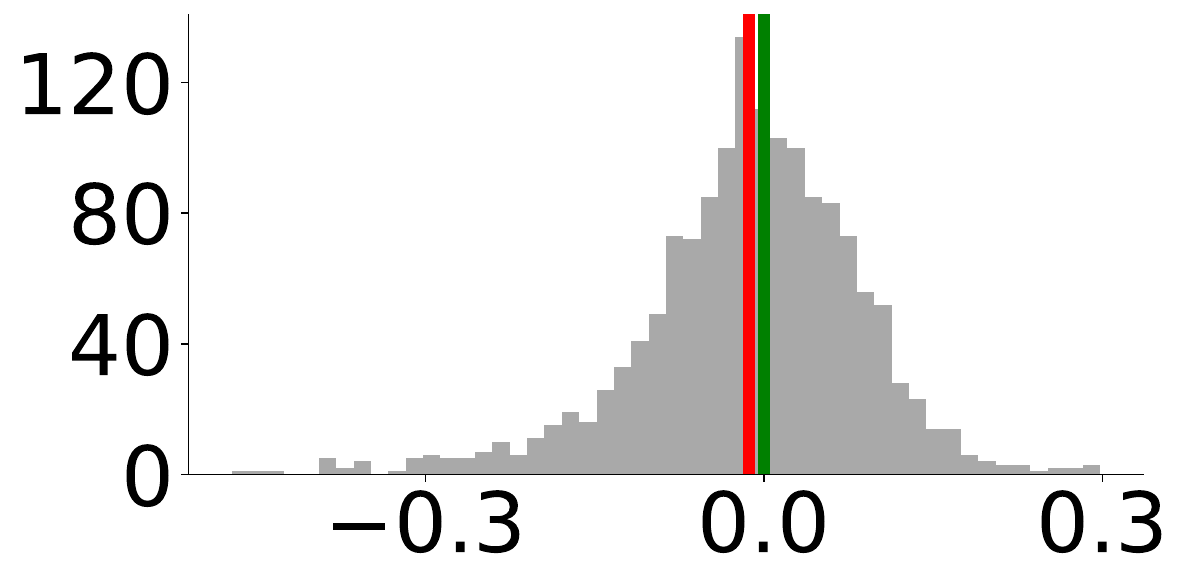}
        \includegraphics[width=2.9cm,height=1.5cm]{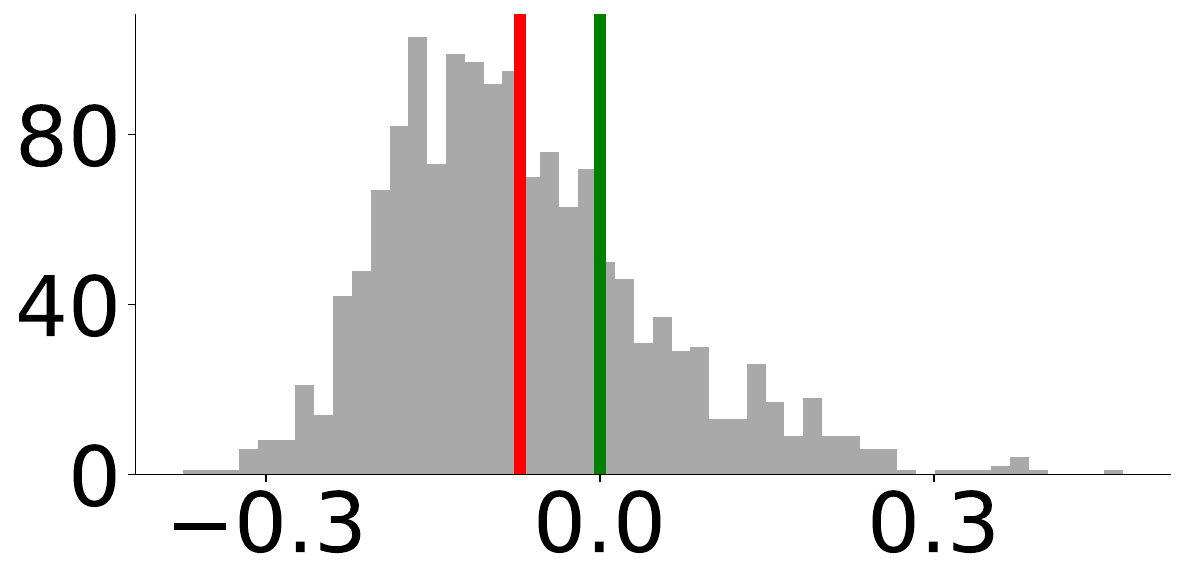}
        \includegraphics[width=2.9cm,height=1.5cm]{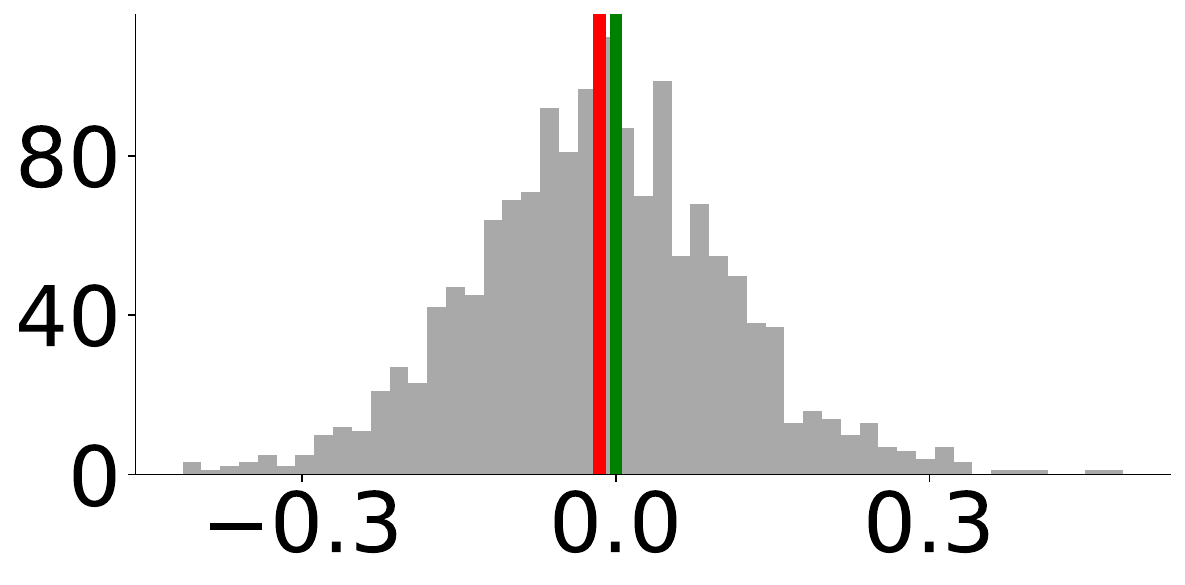}

        {(a) The distribution histograms without any constraints.}

        \includegraphics[width=2.9cm,height=1.5cm]{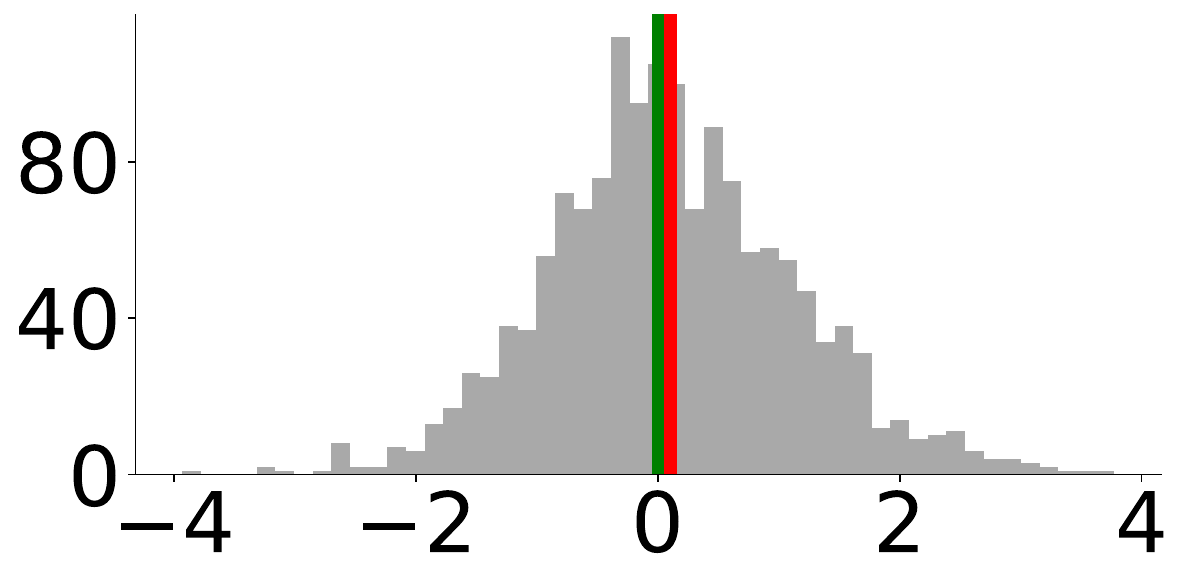}
        \includegraphics[width=2.9cm,height=1.5cm]{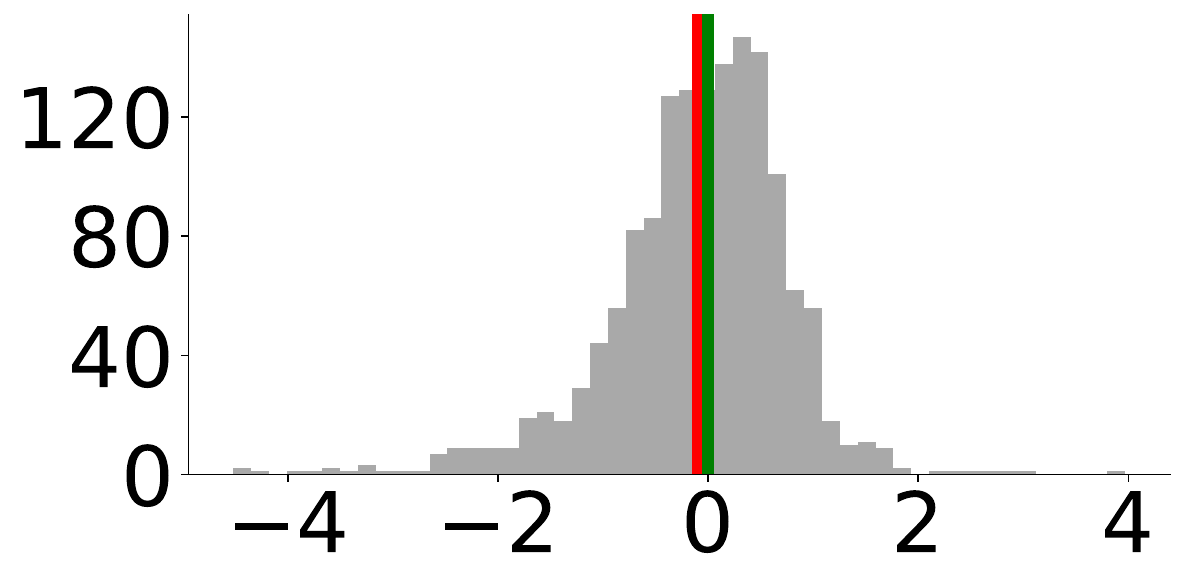}
        \includegraphics[width=2.9cm,height=1.5cm]{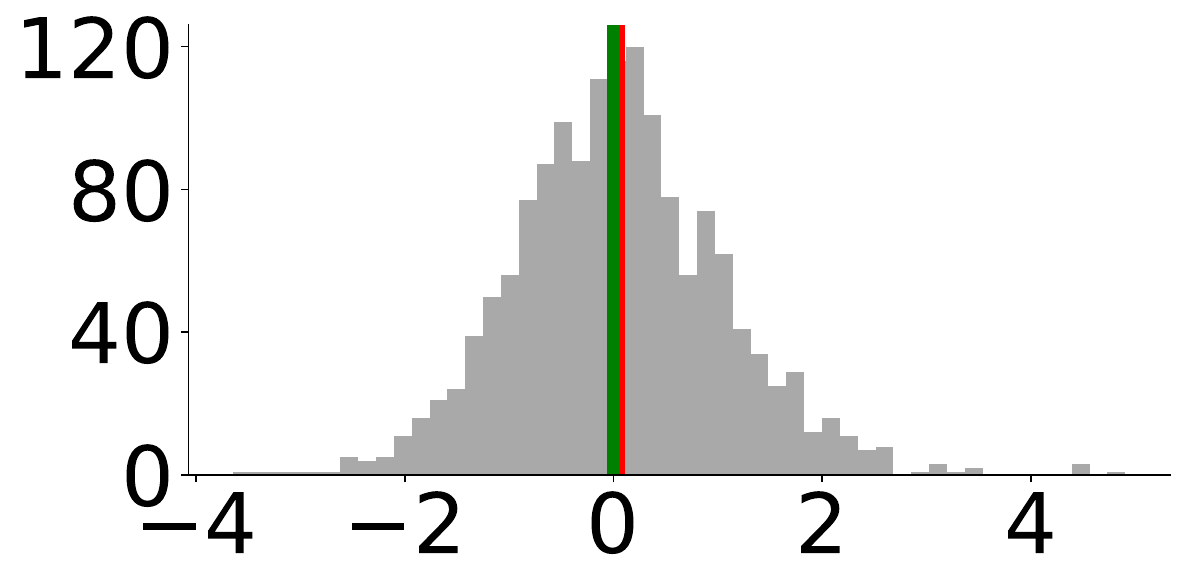}
        \includegraphics[width=2.9cm,height=1.5cm]{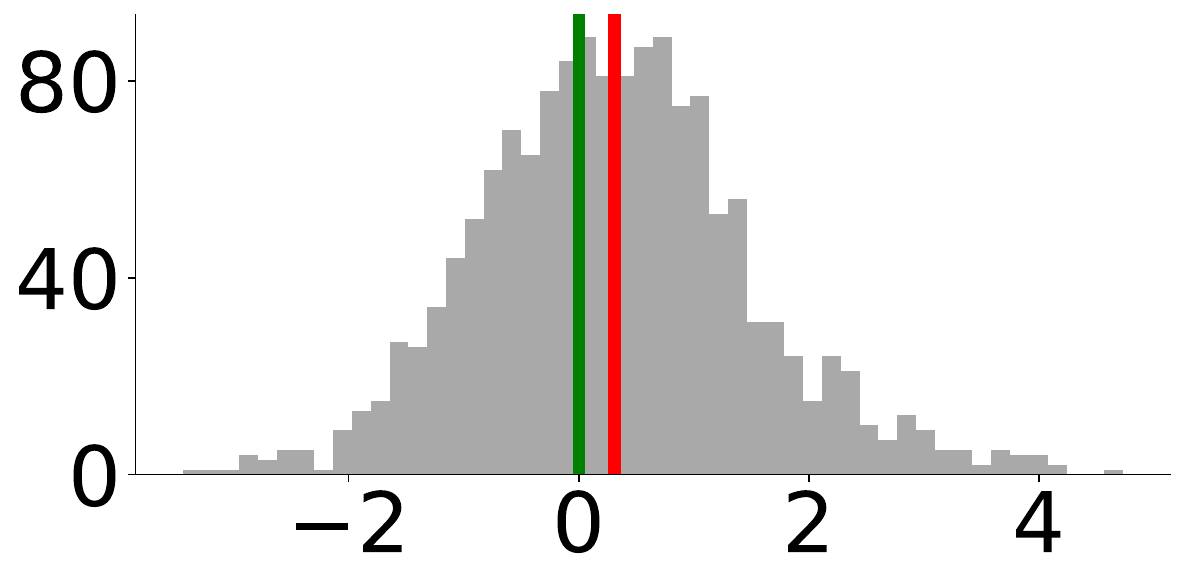}
        \includegraphics[width=2.9cm,height=1.5cm]{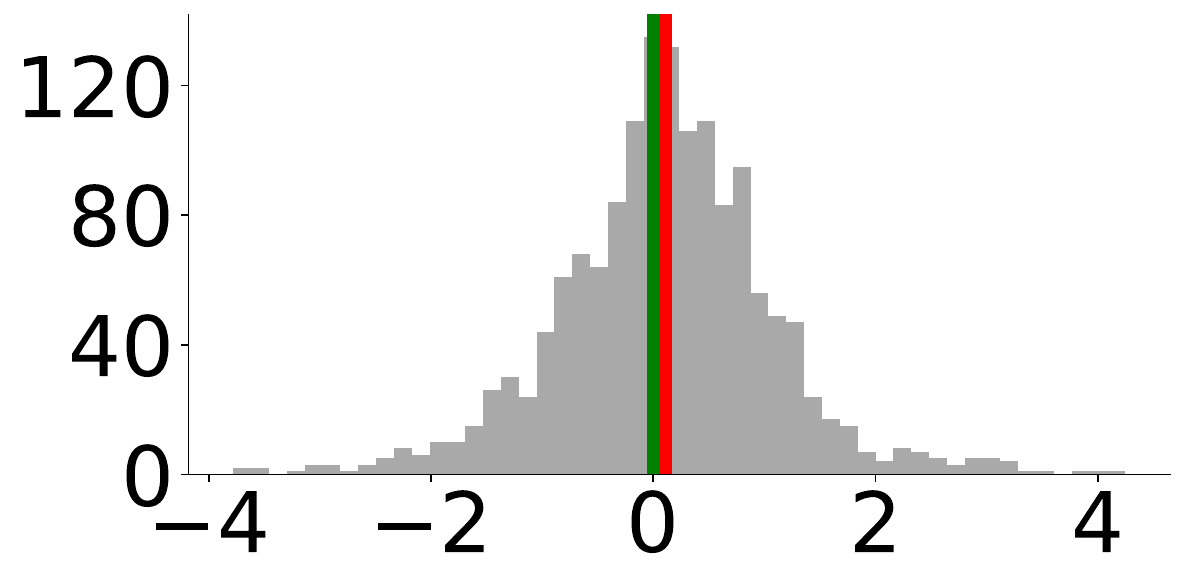}
        \includegraphics[width=2.9cm,height=1.5cm]{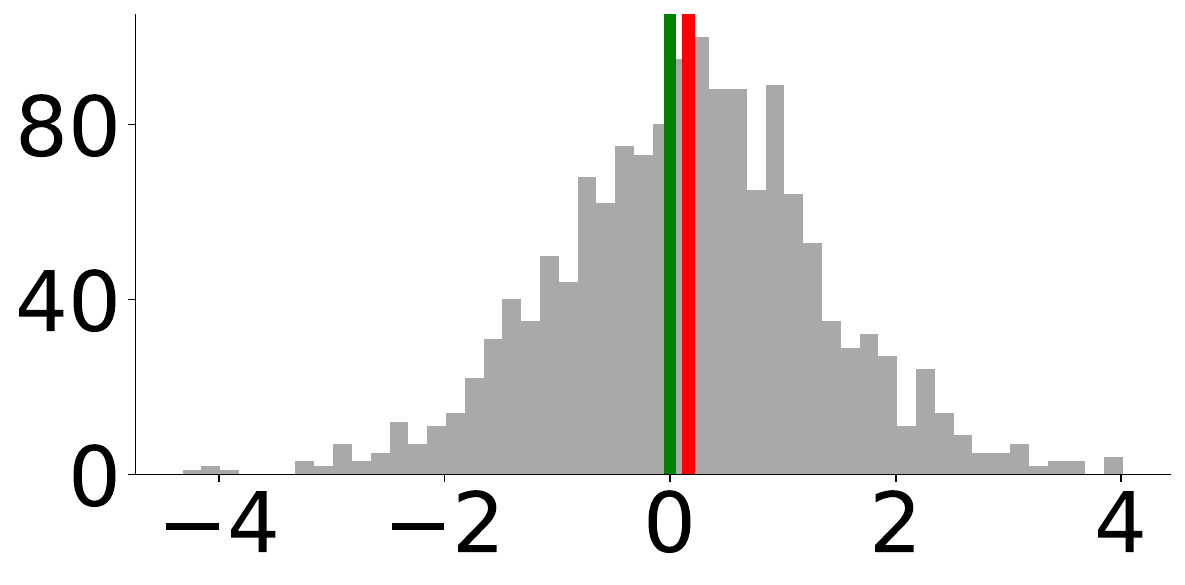}

        {(b) The distribution histograms under an explicit constraint.}

        \caption{The distribution histogram of each dimensions in color code, where the $x$ and $y$ axis coordinates represent the value range and amount, respectively. The red line represents the location of the mean of the data, while the green line denotes $x=0$.}
        \label{fig:color_code_without_any_constraints}
\end{figure*}

The acquisition of CE, CA, and CI capabilities depends on a fundamental concept named \textit{color code}. Our proposed method describes the color semantics of underwater organisms using the color code through a decomposition process. We expect that the color code used in CE, CA, and CI processes should have the following three key properties:
\begin{itemize}
    \item The primary aim of the underwater image enhancement network is to correct distorted colors. Thus, the color code should directly facilitate enhancement results that closely match reference images. \textit{This distinguishes the color code from the style code used in style transfer \cite{chen2022domain}. As illustrated in Fig.~\ref{fig:visualization_of_main_color_map_of_color_code_and_style_code}, the color code should align closely with the reference image, whereas the style code does not share this property.} 
    \item Due to light attenuation underwater, colors with longer wavelengths may be missing in distorted images. During enhancement, these long-wavelength colors should be preserved. \textit{As shown in Fig.~\ref{fig:illustration_of_the_invariant_of_the_hue_of_long_wavelength_colors}, these colors remain approximately invariant throughout the enhancement process. Thus, the color code should approximately preserve such long-wavelength colors.} We will discuss this in detail in Section \ref{subsec:color_guidance_by_the_image_with_long_wavelength_color}. 
    \item Without constraints, the dimensions of the color code may not follow the same distribution, as shown in Fig.~\ref{fig:color_code_without_any_constraints}(a). This lack of constraints impedes the continuous semantic sampling of the color code. \textit{Therefore, the color code should be constrained so that each dimension follows a uniform distribution, as depicted in Fig.~\ref{fig:color_code_without_any_constraints}(b).}
\end{itemize}

The acquisition of these three properties involves three key aspects. First, while color semantics cannot be directly obtained, image content can be separated \cite{huang2018multimodal,cong2024underwater}. In an enhancement network with two semantic codes, one code represents the content while the other represents the color information. Thus, the color code for fixed enhancement can be separated from the feature space. Second, when the color code effectively represents semantic color information, it naturally preserves long-wavelength color semantics, as supervised training maintains these colors in the reference image. This allows for the use of long-wavelength color guidance to adjust the color of enhanced underwater images through color code fusion. Third, by constraining the color code to a specific distribution through dynamic explicit latent space alignment \cite{dziugaite2015training}, richer color properties can be obtained through sampling and interpolation.

In summary, this paper makes three key contributions:
\begin{itemize}
    \item We proposed ColorCode, an underwater image color enhancement method that improves the color of underwater organisms.
    \item By leveraging the hue invariance of long-wavelength colors, the enhancement process can use natural or underwater images as guidance, adapting the colors of the enhanced images accordingly.
    \item By constraining color codes to adhere to a specific distribution during training, diverse color codes can be generated through interpolation at inference, enriching the colors of underwater organisms.
\end{itemize}
Quantitative and visual evaluations on popular and challenging benchmark datasets demonstrate that ColorCode outperforms existing methods in providing diverse, controllable, and realistic color enhancement results.


\section{Related Work}

\subsection{Non-deep Learning-based UIE}
Before the widespread adoption of deep learning \cite{rao2023deep,liu2023wsds} in computer vision tasks \cite{tang2023underwater,zhou2024hclr}, various non-deep learning-based underwater image enhancement (NDL-UIE) algorithms \cite{li2016single,peng2017underwater,carlevaris2010initial,drews2016underwater,song2018rapid,song2020enhancement,zhang2022underwater,zhuang2022underwater,wang2021leveraging} were developed. These methods relied on physical models, digital image processing techniques, and prior assumptions. For instance, MIP \cite{carlevaris2010initial} introduced a prior hypothesis for underwater depth estimation based on differences in color channels. ICSP \cite{hou2023non} proposed an illumination channel sparsity prior to account for non-uniform lighting in underwater environments, guiding variational calculations. SMBLOTMOP \cite{song2020enhancement} combined physical models and channel statistics priors to estimate background light and transmission maps. MLLE \cite{zhang2022underwater} introduced the minimal color loss principle, used alongside local contrast adjustment and color balancing. HLRP \cite{zhuang2022underwater} utilized a hyper-Laplacian reflectance prior, decomposing the complex enhancement task into simpler subproblems through statistical modeling. ULAP \cite{song2018rapid} employed an underwater light attenuation prior to build a depth estimation strategy for image restoration. Despite their advancements, NDL-UIE methods often lack generalization, as assumptions or priors effective in one scene may fail in others \cite{cong2024comprehensive,qi2023deep}.

\subsection{Deep Learning-based UIE}

Numerous deep learning-based UIE algorithms have been developed and validated in recent years \cite{zhang2024robust,zhou2024iacc,wu2024self,xie2024uveb,zhang2024atlantis,wang2024metalantis,bing2023domain,xue2023investigating,jiang2023perception,li2023ruiesr,yan2023uw,lu2023speed}, owing to the availability of large training datasets. We summarize the existing UIE algorithms following the framework in \cite{cong2024comprehensive}.

\textbf{Network architecture.} UWNet \cite{naik2021shallow} proposes a lightweight enhancement framework using simple convolutions. ADMNNet \cite{yan2022attention} introduces an attention-guided dynamic multibranch block to capture diverse features, while U-Trans \cite{peng2023u} combines channel-wise multi-scale feature fusion with spatial-wise global feature modeling based on Transformer architecture \cite{khan2024spectroformer}. WFI2-Net \cite{zhao2023wavelet} integrates spatial and frequency domain information using the Fast Fourier Transform to build a dual-domain feature reconstruction module. AutoEnhancer \cite{tang2022autoenhancer} explores neural architecture search for designing UIE networks.

\textbf{Training strategy.} FUnIEGAN \cite{islam2020fast} employs adversarial training with conditional constraints, while URanker \cite{guo2023underwater} applies rank learning to train and evaluate UIE models. TACL \cite{liu2022twin} utilizes contrastive learning, treating distorted, clear, and enhanced images as negative, positive, and anchor samples. HPUIE-RL \cite{song2024hierarchical} adopts reinforcement learning to design a reward function for fine-tuning the model.

\textbf{Color space.} UColor \cite{li2021underwater} extracts and fuses features from RGB, HSV, and LAB color spaces via a multi-color encoder. UGIF-Net \cite{zhou2023ugif} proposes a color-guided strategy that utilizes six channels of color information from RGB and HSV spaces.

\textbf{Domain adaptation}. TUDA \cite{wang2023domain} constructs image-, feature-, and output-level adaptations to enable knowledge transfer between synthetic and real-world samples. TSDA \cite{jiang2022two} migrates in-air image dehazing techniques to UIE tasks, incorporating style transfer and domain adaptation.

\textbf{Physical model.} USUIR \cite{fu2022unsupervised} estimates global background light, transmission maps, and scene radiance through an unsupervised enhancement process using a re-degradation scheme and homology constraint. GUPDM \cite{mu2023generalized} proposes a dynamic model guided by physical knowledge, simulating various underwater image types by adjusting physical parameters to improve adaptability.

\textbf{Auxiliary task}. SGUIE \cite{qi2022sguie} integrates semantic segmentation for better local feature extraction, and TFUIE \cite{yu2023task} separates segmentation task-friendly content features and task-unfriendly distortion features through an unsupervised feature disentanglement framework. WaterFlow \cite{zhang2023waterflow} designs a detection perception module that balances enhancement quality and object detection performance. CCMSR-Net \cite{qi2023deep} breaks down the UIE process into color correction and visibility enhancement, using Multiscale Retinex theory for color correction.

While these methods have made notable progress, recent studies \cite{kim2021pixel,chen2022domain} propose a new objective that extends beyond a single enhancement result. This new perspective will be discussed further in Section \ref{subsec:the_difference_from_similar_studies}.

\subsection{The Difference from Similar Studies}
\label{subsec:the_difference_from_similar_studies}

The most related studies to ours include PWAE \cite{kim2021pixel}, UIESS \cite{chen2022domain}, and CECF \cite{cong2024underwater}, which aim to produce diverse enhancement results rather than a single fixed one. The key differences between these methods and our ColorCode are summarized as follows:

\begin{itemize}

\item \textbf{PWAE} develops a method for semantic manipulation in latent space, allowing style changes in enhanced images through guidance. However, (i) style is a broad concept, whereas our ColorCode specifically targets color information, and (ii) PWAE's latent space interpolation may not yield diverse results for certain semantics.

\item \textbf{UIESS} offers adjustable enhancement through a multi-domain training process aimed at disentangling content and style. However, (i) its primary goal is to regulate enhancement levels rather than color semantics, and (ii) its latent space lacks semantic constraints.

\item \textbf{CECF} can adjust the colors of underwater organisms using guidance from long-wavelength colors. However, since CECF does not constrain its latent space, it cannot produce diverse color results through latent space interpolation.
\end{itemize}


\section{Method}
\label{sec:methods}

\begin{figure*}
        \small
        \centering
        \includegraphics[width=18cm,height=7cm]{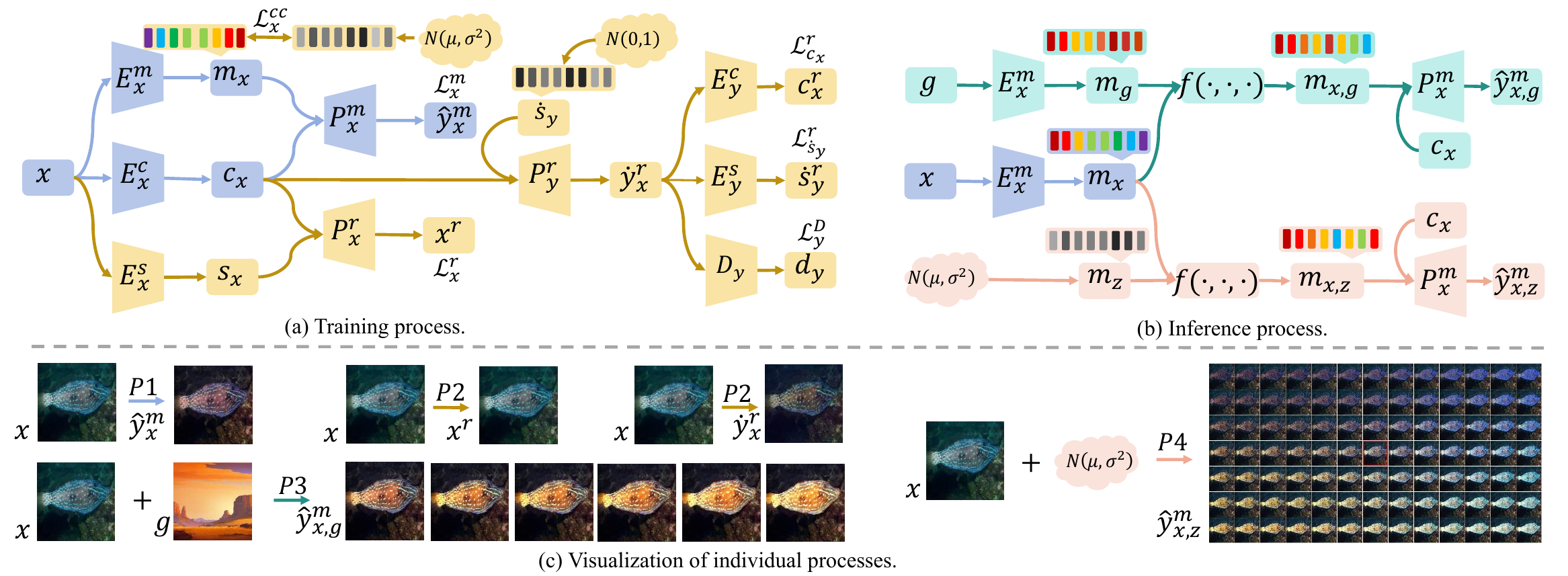}
        \caption{The overall training and inference process of the proposed ColorCode. The $P1$, $P2$, $P3$, and $P4$ stand for the enhancement (Section \ref{subsec:enhancement_process}), decomposition (Section \ref{subsec:extraction_of_the_content_code} and Section \ref{subsec:explicit_constraint_on_the_color_code}), color adaptation (Section \ref{subsec:color_guidance_by_the_image_with_long_wavelength_color}), and color code interpolation (Section \ref{subsec:color_interpolation_by_sampling_color_codes}) processes, respectively. For the meanings of other symbols, please refer to Section \ref{sec:methods}.}
        \label{fig:network_structure}
\end{figure*}

To facilitate the reading of this paper, symbols and their meanings are provided in Table \ref{tab:notations_table}. The overall pipeline of the proposed ColorCode is shown in Fig. \ref{fig:network_structure}, which consists of four main processes, namely (i) a supervised training process \textit{P1} to learn the deterministic enhanced image, (ii) a decomposition process \textit{P2} to decompose color and content, (iii) a color adaptation process \textit{P3} by guidance images with long-wavelength colors, and (iv) an interpolating process \textit{P4} to obtain underwater organisms with diverse colors by continuous color codes. These four processes are introduced in Sections \ref{subsec:enhancement_process}, \ref{subsec:extraction_of_the_content_code} $\&$ \ref{subsec:explicit_constraint_on_the_color_code}, \ref{subsec:color_guidance_by_the_image_with_long_wavelength_color} and \ref{subsec:color_interpolation_by_sampling_color_codes}, respectively.

\begin{table}
        \footnotesize
        \centering
      
        \renewcommand{\arraystretch}{1.2}
      
        \caption{Notations and corresponding meanings.}
        \label{tab:notations_table}
        
        \begin{tabular}{cc}
          \hline   
          Notations                                                           & Meanings \\
          \hline   
          
          $H, W$                                                              & the height, width of an image \\
          $\{x, y\}$ $\in \mathbb{R}^{H \times W \times 3}$                   & a distortion, reference image \\
          $\mathcal{X}$, $\mathcal{Y}$                                        & the distribution of $x$ and $y$    \\
          $g \in \mathbb{R}^{H \times W \times 3}$                            & a guidance with long-wavelength colors \\   
          $\hat{y}_{x}^{m} \in \mathbb{R}^{H \times W \times 3}$              & the deterministic enhanced image \\

          \hdashline                    
   
          $\{m, s\} \in \mathbb{R}^{1 \times K_{m}}$                          & the color, style code \\
          $m_{x}$, $m_{g}$                                                    & the color code for $x$, $g$ \\
          $s_{x}$, $s_{y}$                                                    & the style code for $x$, $y$ \\
          $c \in \mathbb{R}^{K_{c} \times \frac{H}{4} \times \frac{W}{4}}$    & the content code \\
          $c_{x}$, $c_{y}$                                                    & the content feature map for $x$, $y$ \\
          $K_{m}$, $K_{c}$                                                    & the dimensionality of $m$ and $c$ \\  

          \hdashline
       
          $E_{x}^{m}(\cdot)$                                                   & the color encoder for $x$ \\
          $P_{x}^{m}(\cdot, \cdot)$                                            & the color enhancement decoder for $x$ \\
          $E_{x}^{c}(\cdot)$, $E_{y}^{c}(\cdot)$                               & the content encoder for $x$, $y$ \\
          $E_{x}^{s}(\cdot)$, $E_{y}^{s}(\cdot)$                               & the style encoder for $x$, $y$ \\
          $P_{x}^{r}(\cdot, \cdot)$, $P_{y}^{r}(\cdot, \cdot)$                 & the reconstruction decoder for $x$, $y$ \\
          $D_{x}(\cdot)$, $D_{y}(\cdot)$                                       & the domain discriminator for $x$, $y$ \\

          \hdashline
          $\alpha \in \mathbb{R}$                                             & the weight for color fusion process \\
          $\hat{y}_{x,g}^{m} \in \mathbb{R}^{H \times W \times 3}$            & the color adapted $\hat{y}_{x}^{m}$ obtained by $g$ \\
          $\mu \in \mathbb{R}, \sigma^{2} \in \mathbb{R}$                     & the mean, variance of Gaussian distribution \\
          $N(\mu, \sigma^{2})$                                                & Gaussian distribution \\
          $z \in \mathbb{R}^{1 \times K_{m}}$                                 & a vector sample from $N(0, I)$ \\
          $\hat{y}_{x,z}^{m} \in \mathbb{R}^{H \times W \times 3}$            & the color interpolated $\hat{y}_{x}^{m}$ obtained by $z$ \\

          \hdashline
          $\mathcal{F}_{ce}(\cdot)$                                                     & the color enhancement function to obtain $\hat{y}_{x}^{m}$ \\
          $\mathcal{F}_{ca}(\cdot, \cdot, \cdot)$                                       & the color adaptation function to obtain $\hat{y}_{x,g}^{m}$ \\
          $\mathcal{F}_{ci}(\cdot, \cdot)$                                              & the color interpolation function to obtain $\hat{y}_{x,z}^{m}$ \\

          \hline
        \end{tabular}
      
\end{table}

\subsection{Enhancement Process}
\label{subsec:enhancement_process}
The basic purpose of UIE algorithms is to obtain enhancement results that are close to corresponding reference images. In line with this general purpose, the ColorCode we propose can generate a fixed enhanced result colse to the reference. Since we aim to manipulate of the color of underwater organisms, the concept of color code $m$ is proposed. Specifically, we consider the encoding process as the encoding of color code $m$ and content code $c$, where the concept of $c$ is consistent with \cite{huang2018multimodal}. The acquisition of the color code and content code for a distorted underwater image $x$ is
\begin{equation}
        c_{x} = E_{x}^{c}(x),
\end{equation}
\begin{equation}
        \label{eq:m_x}
        m_{x} = E_{x}^{m}(x),
\end{equation}
where $E_{x}^{m}(\cdot)$ and $E_{x}^{c}(\cdot)$ denote the encoders for $m$ and $c$, respectively. After obtaining $m_{x}$ and $c_{x}$, the enhanced image $\hat{y}_{x}^{m}$ is obtained by the color enhancement decoder $P_{x}^{m}(\cdot, \cdot)$ as

\begin{equation}
        \hat{y}_{x}^{m} = P_{x}^{m}(c_{x}, m_{x}).
\end{equation}

To reducing the distance between $\hat{y}_{x}^{m}$ and the reference $y$, the loss function $\mathcal{L}^{m}$ used in the enhancement process is
\begin{equation}
        \label{eq:enhancement_loss}
    \mathcal{L}^{m} = \mathbb{E}_{x, y \thicksim p(x, y)}[||\hat{y}_{x}^{m} - y||_{2} - \phi(\hat{y}_{x}^{m}, y)],
\end{equation}
where $p(x, y)$ represents the joint data distribution of $x$ and $y$. The $\phi(\cdot, \cdot)$ denote the structure loss \cite{zhao2016loss}. We define $P_{x}^{m}(E_{x}^{c}(\cdot), E_{x}^{m}(\cdot))$ as the color enhancement function $\mathcal{F}_{ce}(\cdot)$, which including all networks that need to be used during the inference phase. The enhancement process of the distorted underwater image is shown in Fig. \ref{fig:network_structure}, which is marked as $P1$. In the above analysis, we assume that $c_{x}$ and $m_{x}$ have the semantics we want. We will discuss how to obtain such semantics in Section \ref{subsec:extraction_of_the_content_code} and Section \ref{subsec:explicit_constraint_on_the_color_code}.


\subsection{Decomposition of the Color Code and Content Code}
\label{subsec:extraction_of_the_content_code}
In the above discussion, we assume that the color code and content code have the semantic information we want. However, it is almost impossible to obtain effective semantic representation by directly optimizing Eq. \ref{eq:enhancement_loss}. Therefore, we use an implicit way to constrain the encoders $E_{x}^{m}(\cdot)$ and $E_{x}^{c}(\cdot)$.

When the semantics of one encoder have been acquired, the semantics of the other encoder can be acquired implicitly. Assuming that we have obtained the content code, which can represent the structure and texture information of the image. By optimizing the Eq. \ref{eq:enhancement_loss}, when the encoder $E_{x}^{c}(\cdot)$ can represent the content information, the other encoder $E_{x}^{m}(\cdot)$ can represent the color information in the enhancement process. Here, we adopt the self-reconstruction and cross-reconstruction process provided by MUNIT \cite{huang2018multimodal} to obtain the content representations.

\subsubsection{Self-reconstruction}
For a given image $x$ and $y$, after going through the respective encoding and decoding paths, the corresponding reconstructed image can be obtained. The loss of self-reconstruction is

\begin{equation}
        \begin{split}
            \mathcal{L}_{x,y}^{r} = \hspace{0.1cm} & \mathbb{E}_{x, y \thicksim p(x, y)}[||P_{x}^{r}(E_{x}^{c}(x), E_{x}^{s}(x)) - x||_{1} + \\
                                         & ||P_{y}^{r}(E_{y}^{c}(y), E_{y}^{s}(y)) - y||_{1}].
        \end{split}
\end{equation}

\subsubsection{Cross-reconstruction}
In order to decompose the content from the image, the content code and style code need to be reconstructed after passing through the decoding and encoding paths \cite{huang2018multimodal}. The reconstruction loss of the content code is
\begin{equation}
        \small
        \begin{split}
        \mathcal{L}_{c_{x}, c_{y}}^{r} = \hspace{0.1cm} & \mathbb{E}_{c_{x} \thicksim p(c_{x}), \dot{s}_{y} \thicksim q(\dot{s}_{y})}[||E_{y}^{c}(P_{y}^{r}(c_{x}, \dot{s}_{y})) - c_{x}||_{1}] + \\
                                              & \mathbb{E}_{c_{y} \thicksim p(c_{y}), \dot{s}_{x} \thicksim q(\dot{s}_{x})}[||E_{x}^{c}(P_{x}^{r}(c_{y}, \dot{s}_{x})) - c_{y}||_{1}],
        \end{split}
\end{equation}
where $q(\dot{s}_{x})$ and $q(\dot{s}_{y})$ represent the standard normal distribution $N(0, I)$. The $c_{x}$ and $c_{y}$ stand for the encoding result $E_{x}^{c}(x)$ and $E_{y}^{c}(y)$, respectively. Similarly, the reconstruction loss of the style code is

\begin{equation}
        \small
        \begin{split}
        \mathcal{L}_{\dot{s}_{x}, \dot{s}_{y}}^{r} = \hspace{0.1cm} & \mathbb{E}_{c_{x} \thicksim p(c_{x}), \dot{s}_{y} \thicksim q(\dot{s}_{y})}[||E_{y}^{s}(P_{y}^{r}(c_{x}, \dot{s}_{y})) - \dot{s}_{y}||_{1}] + \\
                                              & \mathbb{E}_{c_{y} \thicksim p(c_{y}), \dot{s}_{x} \thicksim q(\dot{s}_{x})}[||E_{x}^{s}(P_{x}^{r}(c_{y}, \dot{s}_{x})) - \dot{s}_{x}||_{1}].
        \end{split}
\end{equation}

For the generated images decoded from each image domain, adversarial training can be used to match the distribution. The adversarial losses for the two domains are
\begin{equation}
        \small
        \begin{split}
        \mathcal{L}_{x}^{D} = \hspace{0.1cm} & \mathbb{E}_{c_{y} \thicksim p(c_{y}), \dot{s}_{x} \thicksim q(\dot{s}_{x})}[\log (1 - D_{x}(P_{x}^{r}(c_{y}, \dot{s}_{x})))] + \\
                                     & \mathbb{E}_{x \thicksim p(x)}[\log D_{x}(x)],
        \end{split}
\end{equation}
    
\begin{equation}
        \small
        \begin{split}
        \mathcal{L}_{y}^{D} = \hspace{0.1cm} & \mathbb{E}_{c_{x} \thicksim p(c_{x}), \dot{s}_{y} \thicksim q(\dot{s}_{y})}[\log (1 - D_{y}(P_{y}^{r}(c_{x}, \dot{s}_{y})))] + \\
                                     & \mathbb{E}_{y \thicksim p(y)}[\log D_{y}(y)],
        \end{split}
\end{equation}

The self- and cross-reconstruction processes of the content code and style code are shown in Fig. \ref{fig:network_structure}, which is marked as $P2$. The $\mathcal{L}_{c_{x}, c_{y}}^{r}$, $\mathcal{L}_{c_{x}, c_{y}}^{r}$ and $\mathcal{L}_{\dot{s}_{x}, \dot{s}_{y}}^{r}$ are regarded as the sum of the losses of the two domains, which are
\begin{equation}
        \begin{cases}
                \mathcal{L}_{x,y}^{r} = \mathcal{L}_{x}^{r} + \mathcal{L}_{y}^{r}, \\
                \mathcal{L}_{c_{x}, c_{y}}^{r} = \mathcal{L}_{c_{x}}^{r} + \mathcal{L}_{c_{y}}^{r}, \\
                \mathcal{L}_{\dot{s}_{x}, \dot{s}_{y}}^{r} = \mathcal{L}_{\dot{s}_{x}}^{r} + \mathcal{L}_{\dot{s}_{y}}^{r}. \\
        \end{cases}
\end{equation}

It is worth emphasizing that the purpose of using losses ($\mathcal{L}_{x,y}^{r}$, $\mathcal{L}_{c_{x}, c_{y}}^{r}$, $\mathcal{L}_{\dot{s}_{x}, \dot{s}_{y}}^{r}$, $\mathcal{L}_{x}^{D}$, $\mathcal{L}_{y}^{D}$) is to learn the content code $c_{x}$ by $E_{x}^{c}(\cdot)$. The style code is merely a concept used to assist in the training process.

\subsection{Explicit Constraints on the Color Code}
\label{subsec:explicit_constraint_on_the_color_code}
When no constraint is performed, the color code may not conform to a known distribution. As shown in Fig. \ref{fig:color_code_without_any_constraints}(a), different dimensions of color code do not seem to follow the same distribution. Obviously, the means of color codes of different dimensions are different. A detailed discussion is placed in Section \ref{subsubsec:analysis_of_the_constraint_on_color_code}. A series of color codes with continuous color properties may not be generated if we can not sample color codes from a known distribution. 

However, in order to obtain diverse colors by interpolation, we must ensure that the color code conforms to a certain distribution $\mathcal{P}$. Specifically, we hope that the color encoder $E_{x}^{m}(\cdot)$ can be used as a mapping function so that $E_{x}^{m}(x) \sim \mathcal{P}$ holds. The optimization purpose \cite{dziugaite2015training} for color code is
\begin{equation}
        \arg \min_{\theta} \Gamma(\mathcal{P}, E_{x}^{m}{(\mathcal{X})}),
\end{equation}
where $\theta$ denotes the parameters for $E_{x}^{m}(\cdot)$ and $x \sim \mathcal{X}$. $\Gamma$ represents a measure of discrepancy.
Existing work \cite{li2019joint,tolstikhin2017wasserstein} has shown that Gaussian distribution is suitable for latent space representation of generative models. We choose Gaussian distribution $N(\mu, \sigma^{2})$ as $\mathcal{P}$. In each iteration of the training process of ColorCode, we can sample the independent and identically distributed noise $z \sim N(\mu, \sigma^{2})$. The maximum mean discrepancy \cite{dziugaite2015training} is used as the constraint
\begin{equation}
        \label{eq:mmd_loss}
        \begin{split}
        \mathcal{L}_{x}^{cc} = & \frac{1}{n^{2} - n}[\sum_{l \neq j}^{n}{k(m_{x}^{l}, m_{x}^{j})} + \sum_{l \neq j}^{n}{k(z^{l}, z^{j})}] \\
                               & - \frac{2}{n^{2}}{\sum_{l}^{n} \sum_{j}^{n} {k(m_{x}^{l}, z^{j})}},
        \end{split}
\end{equation}
where $n$ and $k(\cdot, \cdot)$ denote the number of training examples and kernel function \cite{tolstikhin2017wasserstein}, respectively. With the above explicit constraints, the color code is approximately following $N(\mu, \sigma^{2})$ as shown in Fig. \ref{fig:color_code_without_any_constraints}. Therefore, we can sample color codes from a known distribution in the inference phase.


\begin{table*}
    
        \centering
        \scriptsize
        \setlength{\tabcolsep}{2.9mm}
        \renewcommand{\arraystretch}{1.1}
          \flushleft
          \caption{Quantitative evaluations on benchmark datasets. The first, second, and third best performances are in \textcolor{red}{red}, \textcolor{green}{green}, and \textcolor{blue}{blue}, respectively.}
          \label{tab:full_reference_results}
            
            \begin{tabular}{@{}c|ccc|ccc|ccc|ccc@{}}
              \toprule
              \multirow{2}{*}{Method} & \multicolumn{3}{c|}{UIEB} & \multicolumn{3}{c|}{LSUI} & \multicolumn{3}{c}{EUVP} & \multicolumn{3}{c}{UFO-120}\\
              
              & PSNR$\uparrow$ & SSIM$\uparrow$ & UIQM$\uparrow$ & PSNR$\uparrow$ & SSIM$\uparrow$ & UIQM$\uparrow$ & PSNR$\uparrow$ & SSIM$\uparrow$ & UIQM$\uparrow$ & PSNR$\uparrow$ & SSIM$\uparrow$ & UIQM$\uparrow$\\
              \midrule

              {IBLA}        & 15.47 & 0.6786 & 4.153 & 18.11 & 0.7278 & 3.464 & 15.73 & 0.6404 & 2.772 & 17.65 & 0.6715 & 3.200 \\
              {MIP}         & 13.84 & 0.6055 & 4.231 & 15.78 & 0.6526 & 3.629 & 15.61 & 0.6316 & 3.077 & 16.27 & 0.6095 & 3.643 \\
              {UDCP}        & 11.59 & 0.5263 & \textcolor{green}{4.510} & 13.78 & 0.5829 & 4.085 & 12.17 & 0.5149 & 3.290 & 14.77 & 0.5818 & 3.840 \\
              {ULAP}        & 14.64 & 0.6942 & \textcolor{blue}{4.501} & 18.03 & 0.7459 & 3.882 & 17.21 & 0.6930 & 3.325 & 19.40 & 0.7420 & 4.133 \\
              {SMBLOTMOP}   & 16.65 & 0.7648 & 4.042 & 16.64 & 0.7360 & 4.033 & 16.16 & 0.6630 & 3.483 & 15.99 & 0.6968 & 3.828 \\
              {MLLE}        & 17.21 & 0.7360 & 4.269 & 16.83 & 0.6726 & 4.230 & 16.25 & 0.6025 & 4.308 & 15.01 & 0.6070 & 4.203 \\
              {HLRP}        & 15.40 & 0.6343 & 4.457 & 15.96 & 0.6255 & 4.542 & 15.24 & 0.5612 & 4.642 & 14.34 & 0.5402 & \textcolor{red}{4.469} \\

              {UWNet}       & 17.36 & 0.7948 & 4.188 & 22.01 & 0.8489 & 4.284 & 22.54 & 0.8115 & 4.293 & 25.02 & 0.8505 & 3.948 \\
              {PhyicalNN}   & 18.08 & 0.7693 & 4.112 & 22.52 & 0.8468 & 4.382 & 23.68 & 0.8108 & 4.550 & 25.68 & 0.8414 & 4.334 \\
              {FUnIEGAN}    & 19.12 & 0.8321 & 4.480 & 24.23 & 0.8579 & 4.487 & 23.06 & 0.7731 & \textcolor{red}{4.910} & 26.79 & 0.8378 & \textcolor{green}{4.383} \\
              {WaterNet}    & 21.04 & 0.8601 & 4.220 & 22.74 & 0.8560 & 4.361 & 23.99 & 0.8325 & 4.613 & 23.84 & 0.8188 & \textcolor{blue}{4.373} \\
              {ADMNNet}     & 20.77 & 0.8709 & 4.386 & 26.11 & 0.9051 & 4.510 & 24.65 & \textcolor{blue}{0.8424} & \textcolor{green}{4.712} & 26.67 & 0.8757 & 4.129 \\
              {UColor}      & 20.13 & \textcolor{blue}{0.8769} & 4.322 & 24.03 & 0.8855 & 4.401 & 24.43 & \textcolor{green}{0.8462} & 4.602 & 26.95 & \textcolor{red}{0.8877} & 4.203 \\
              {UGAN}        & \textcolor{green}{21.37} & 0.8534 & \textcolor{red}{4.556} & 26.93 & 0.8983 & \textcolor{red}{4.636} & \textcolor{green}{24.71} & 0.8276 & 4.649 & \textcolor{green}{27.61} & 0.8672 & 4.295 \\
              {U-Trans}     & 20.39 & 0.8034 & 4.034 & \textcolor{green}{27.48} & 0.9013 & 4.452 & 24.56 & 0.8309 & 4.689 & \textcolor{blue}{27.60} & 0.8525 & 4.334 \\
              {UIE-WD}      & 20.10 & 0.8617 & 4.263 & 26.81 & \textcolor{blue}{0.9088} & 4.560 & 24.55 & 0.8365 & 4.694 & 27.48 & 0.8756 & 4.221 \\
              {SGUIE}       & \textcolor{blue}{21.12} & \textcolor{green}{0.8882} & 4.392 & \textcolor{red}{27.77} & \textcolor{red}{0.9129} & \textcolor{green}{4.610} & \textcolor{blue}{24.68} & 0.8356 & 4.675 & \textcolor{red}{28.30} & \textcolor{blue}{0.8784} & 4.263 \\
              \midrule
              {ColorCode}      & \textcolor{red}{21.65} & \textcolor{red}{0.8910} & 4.225 & \textcolor{blue}{27.01} & \textcolor{green}{0.9104} & \textcolor{blue}{4.561} & \textcolor{red}{25.10} & \textcolor{red}{0.8571} & \textcolor{blue}{4.698} & 26.79 & \textcolor{green}{0.8786} & 4.291 \\
              \bottomrule
            
            \end{tabular}
          
\end{table*}

\subsection{Color Adaptation by Image with Long-wavelength Colors}
\label{subsec:color_guidance_by_the_image_with_long_wavelength_color}
As the water depth increases, light components with longer wavelengths disappear earlier. According to \cite{zhang2019survey}, the order of decay is red, orange, yellow, green and blue. When referring to distorted underwater images, we usually mean images with obvious loss of red and yellow components \cite{cong2024comprehensive}. Here, we divide the distortion into global distortion and non-global distortion according to the degree of distortion. 
\begin{itemize}
        \item Global distortion refers to the lack of red or yellow tones almost over the entire image.
        \item Non-global distortion refers to the red or yellow tones that may be retained in local areas of the image.
\end{itemize}

As we can observe in Fig. \ref{fig:illustration_of_the_invariant_of_the_hue_of_long_wavelength_colors}(a), when the entire image has a noticeable blue-green or blue distortion, the colors with long wavelength are absent from the distorted images. When there is a slight local distortion inside the distorted image as shown in Fig. \ref{fig:illustration_of_the_invariant_of_the_hue_of_long_wavelength_colors}(b), the hue of the colors with long-wavelength remains approximately preserved. It is worth pointing out that the hue here is approximately preserved, not strictly preserved.

For a well-trained UIE model, when the input distorted image contains long-wavelength colors, these colors are usually not mapped to blue-green, but are retained approximately. This means that the hue of long-wavelength colors should be preserved by the UIE model. We call this observation the hue invariance of long-wavelength colors \cite{zhang2019survey,cong2024underwater}.

Based on the observation of the hue invariance of long-wavelength colors, we propose the concept of guidance $g$. For an image with global long-wavelength colors, assuming that we can obtain a color code with color semantics, then this color code should preserve the hue of the long-wavelength colors. We define the color adaptation function as $\mathcal{F}_{ca}(\cdot, \cdot, \cdot)$, where the three parameters represent the distorted underwater image, color guidance and weight factor, respectively. The color adaptation process of the enhanced underwater image is shown in Fig. \ref{fig:network_structure}, which is marked as $P3$. For the original distorted image $x$, the color guidance function can provide a fixed color enhancement result as
\begin{equation}
        \hat{y}_{x}^{m} = \mathcal{F}_{ca}(x, x, 0).
\end{equation}

Beyond this general purpose, when a guidance $g$ with a long-wavelength color is given, $\mathcal{F}_{ca}(\cdot, \cdot, \cdot)$ can generate $\hat{y}_{x,g}^{m}$ with the color properties of $g$. During this color adaptation process, the content of the image should remain unchanged. The purpose of color adaptation can be expressed by
\begin{equation}
        \hat{y}_{x,g}^{m} = \mathcal{F}_{ca}(x, g, \alpha).
\end{equation}

Color adaptation is achieved by the fusion of color codes. Specifically, the color code $m_{x}$ of $x$ and the color code $m_{g}$ of the guidance $g$ are fused under a given weight $\alpha$. The transformation function \cite{mi2021revisiting} in $N$-dimensional space is
\begin{equation}
        t : (m_{x}, m_{g}, \alpha) \in \mathbb{R}^{N} \times \mathbb{R}^{N} \times [0, 1] \mapsto m_{x}, m_{g} \in \mathbb{R}^{N},
\end{equation}
where the color code $m_{g}$ is obtained by
\begin{equation}
        m_{g} = E_{x}^{m}(g).
\end{equation}

The function $\mathcal{F}_{ca}(\cdot, \cdot, \cdot)$ fuses $m_{x}$ and $m_{g}$ in the form of
\begin{equation}
        \label{eq:fusion_of_mx_and_mg}
        m_{x, g} = \psi(m_{x}, m_{g}) = \frac{(1 - \alpha) \cdot m_{x} + \alpha \cdot \kappa(m_{g})}{\sqrt{(1 - \alpha)^{2} + \alpha^{2}}},
\end{equation}
where $\kappa(\cdot)$ denotes the truncation operation. The color adaptation result $\hat{y}_{x,g}^{m}$ is
\begin{equation}
        \hat{y}_{x,g}^{m} = P_{x}^{m}(c_{x}, m_{x, g}).
\end{equation}

It is worth pointing out that what we ultimately want to obtain is not $\hat{y}_{x,g}^{m}$, but the masked $\hat{y}_{x,g}^{m} \odot mask$, where the foreground $mask$ is obtained by semantic segmentation \cite{islam2020semantic}.

\subsection{Color Interpolation by Sampling Color Codes}
\label{subsec:color_interpolation_by_sampling_color_codes}
By the constraints provided by Eq. \ref{eq:mmd_loss}, the color code will conform to $N(\mu, \sigma^{2})$. Therefore, we can sample new color codes $m_{z}$ from $N(\mu, \sigma^{2})$. Then, follow the same process as Eq. \ref{eq:fusion_of_mx_and_mg}, that is, replace $m_{g}$ with $m_{z}$, to obtain the fused color code $m_{x, z}$. The color interpolated-image is obtained by
\begin{equation}
        \hat{y}_{x,z}^{m} = P_{x}^{m}(c_{x}, m_{x,z}),
\end{equation}
the above process is defined as the color interpolation function $\mathcal{F}_{ci}(\cdot, \cdot)$. During the inference process, values at equal intervals can be sampled to achieve the purpose of color interpolation.

\subsection{The Overall Loss Function}
The overall loss function of ColorCode includes adversarial loss $\mathcal{L}_{x}^{D}$ and $\mathcal{L}_{y}^{D}$ for domain discrimination, restoration loss $\mathcal{L}^{m}$ for enhancement, pixel loss $\mathcal{L}_{x,y}^{r}$ for self-reconstruction, $\mathcal{L}_1$ loss $\mathcal{L}_{c_x, c_y}^{r}$ and $\mathcal{L}_{\dot{s}_x, \dot{s}_y}^{r}$ for the decomposition, and distribution constraint loss $\mathcal{L}_{x}^{cc}$ for controlling the distribution of color codes. All networks are trained synchronously. The overall loss function is
\begin{equation}
        \label{eq:total_loss}
        \begin{split}
        \min_{\mathbb{M}} \max_{\mathbb{N}} \mathcal{L} (\mathbb{M}, \mathbb{N}) = & \mathcal{L}_{x}^{D} + \mathcal{L}_{y}^{D} + \lambda_{1} \cdot  (\mathcal{L}^{m}  + \mathcal{L}_{x,y}^{r}) + \\
                                                                        & \lambda_{2} \cdot (\mathcal{L}_{c_x, c_y}^{r} + \mathcal{L}_{\dot{s}_x, \dot{s}_y}^{r}) + \lambda_{3} \cdot \mathcal{L}_{x}^{cc}, \\
        \end{split}
    \end{equation}
where $\lambda_1$, $\lambda_2$ and $\lambda_3$ represent weight factors for different terms. For simplicity, different networks are represented by sets as $\mathbb{M}$ and $\mathbb{N}$, which are
\begin{equation}
    \mathbb{M} = \{E_{x}^{m}, E_{x}^{c}, E_{x}^{s}, E_{y}^{c}, E_{y}^{s}, P_{x}^{r}, P_{x}^{m}, P_{y}^{r}\},
\end{equation}
\begin{equation}
    \mathbb{N} = \{D_{x}, D_{y}\}.
\end{equation}

\section{Experiments}
Section \ref{subsec:experimental_settings} introduces the experimental settings. Quantitative and qualitative experimental results of CE, CA and CI capabilities are given in Sections \ref{subsec:experiment_color_enhancement}, \ref{subsec:experiment_color_fine_tuning} and \ref{subsec:experiment_color_interpolation}, respectively. Section \ref{subsec:ablation_study} provides analyses of the components and capabilities of ColorCode. Extensive visual results are provided in the Supplementary Material.

\subsection{Experimental Settings}
\label{subsec:experimental_settings}

\subsubsection{Datasets}
UIEB \cite{li2019underwater}, LSUI \cite{peng2023u}, EUVP (image-net) from \cite{islam2020fast} and UFO-120 \cite{islam2020simultaneous} are used for evaluations.

\subsubsection{Evaluation Metrics}
Common evaluation metrics for the UIE task are used, which include the Peak Signal-to-Noise Ratio (PSNR) \cite{kim2021pixel}, Structural Similarity (SSIM) \cite{wang2004image} and Underwater Image Quality Measures (UIQM) \cite{panetta2015human}. UIQM is calculated by the MATLAB code provided by \cite{huang2023contrastive}.


\begin{figure*}
        \centering
        \includegraphics[width=1.9cm,height=2.0cm]{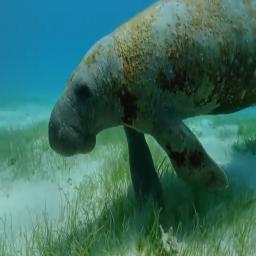}
        \includegraphics[width=1.9cm,height=2.0cm]{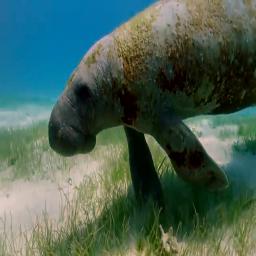}
        \includegraphics[width=1.9cm,height=2.0cm]{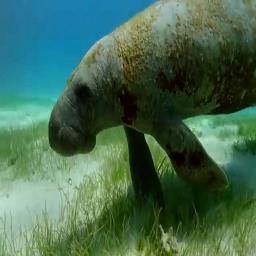}
        \includegraphics[width=1.9cm,height=2.0cm]{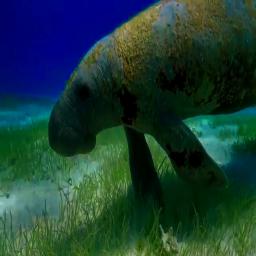}
        \includegraphics[width=1.9cm,height=2.0cm]{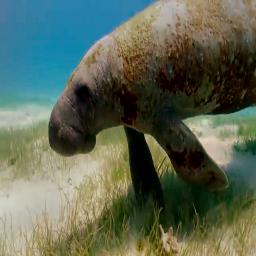}
        \includegraphics[width=1.9cm,height=2.0cm]{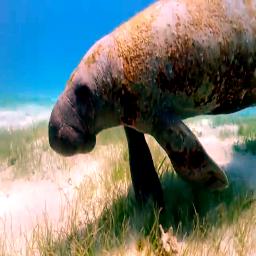}
        \includegraphics[width=1.9cm,height=2.0cm]{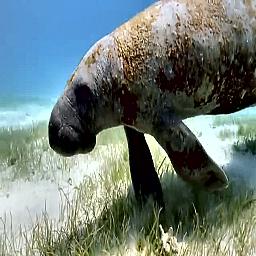}
        \includegraphics[width=1.9cm,height=2.0cm]{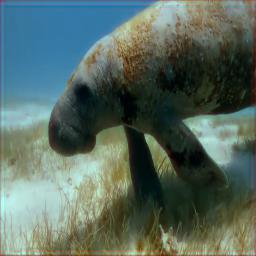}
        \includegraphics[width=1.9cm,height=2.0cm]{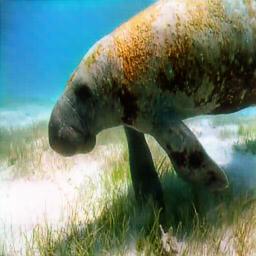}

        \leftline{\hspace{0.1cm} Distortion \hspace{0.7cm} IBLA \hspace{0.9cm} UDCP \hspace{0.9cm}  MIP \hspace{1cm} ULAP \hspace{0.3cm} SMBLOTMOP \hspace{0.3cm} MLLE \hspace{0.45cm} PhysicalNN \hspace{0.05cm} FUnIEGAN}
    
        \vspace{0.1cm}
    
        \includegraphics[width=1.9cm,height=2.0cm]{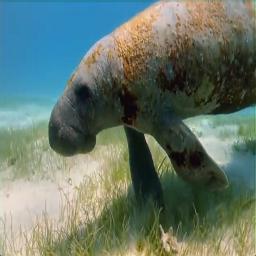}
        \includegraphics[width=1.9cm,height=2.0cm]{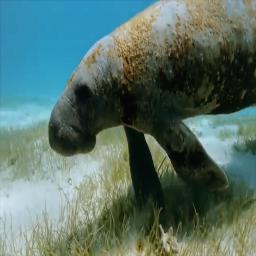}
        \includegraphics[width=1.9cm,height=2.0cm]{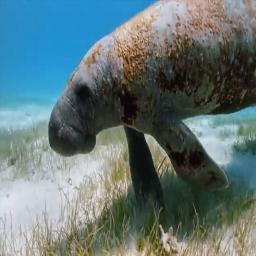}
        \includegraphics[width=1.9cm,height=2.0cm]{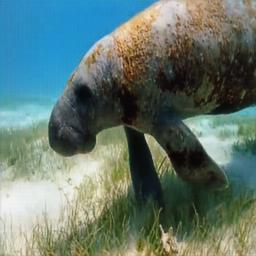}
        \includegraphics[width=1.9cm,height=2.0cm]{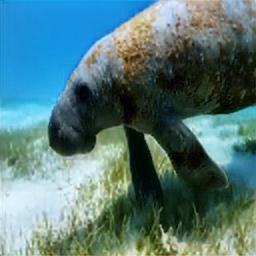}
        \includegraphics[width=1.9cm,height=2.0cm]{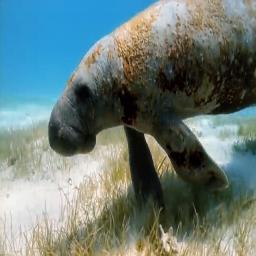}
        \includegraphics[width=1.9cm,height=2.0cm]{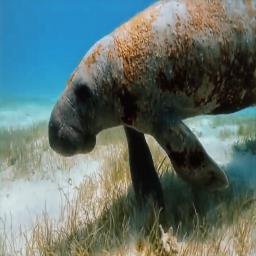}
        \includegraphics[width=1.9cm,height=2.0cm]{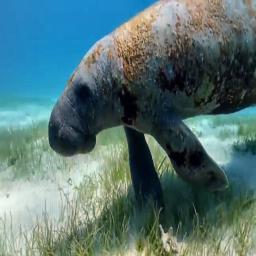}
        \includegraphics[width=1.9cm,height=2.0cm]{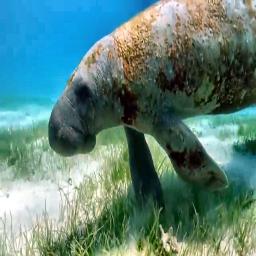}

        \leftline{\hspace{0.2cm} WaterNet \hspace{0.3cm} ADMNNet \hspace{0.5cm} UColor \hspace{0.7cm} UGAN \hspace{0.7cm} U-Trans \hspace{0.5cm} UIE-WD \hspace{0.6cm} SGUIE \hspace{0.6cm} ColorCode \hspace{0.3cm} Reference}
    
        \caption{Visual results obtained by various UIE algorithms on UIEB dataset.}
        \label{fig:visual_results_uieb_2774}
      \end{figure*}

\begin{figure}[!h]
        \small
        \centering
        \includegraphics[width=8.8cm,height=3cm]{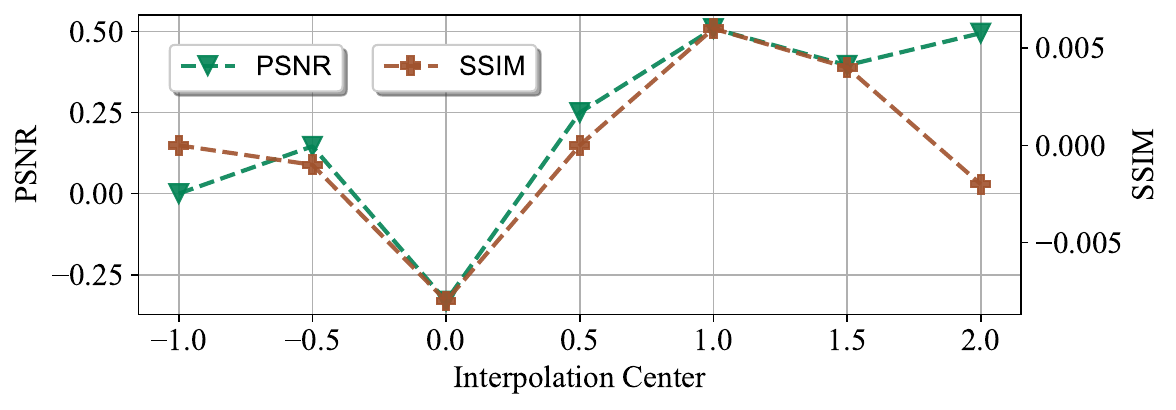}
        \caption{Evaluation curves for the effect of choosing different interpolation centers on the performance. The values of PSNR and SSIM are changes relative to the interpolation center at -1.}
        \label{fig:differ_gaussian_mean}
\end{figure}

\begin{figure}
        \small
        \centering
        \includegraphics[width=8.5cm,height=3cm]{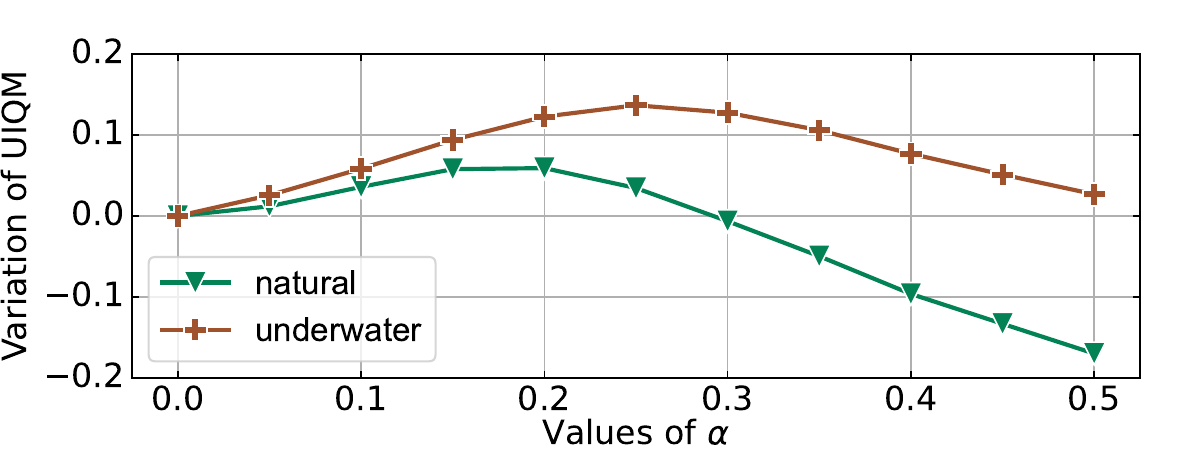}
        \caption{Quantitative evaluation curves obtained by using natural images and underwater images as guidance under different $\alpha$ settings.}
        \label{fig:eval_NR_differ_guidence_images_UIEB}
\end{figure}

\subsubsection{Comparison Algorithms}
The algorithms used for comparison include IBLA \cite{peng2017underwater}, MIP \cite{carlevaris2010initial}, UDCP \cite{drews2016underwater}, ULAP \cite{song2018rapid}, SMBLOTMOP \cite{song2020enhancement}, MLLE \cite{zhang2022underwater}, HLRP \cite{zhuang2022underwater}, UWNet \cite{naik2021shallow}, PhysicalNN \cite{chen2021underwater}, FUnIEGAN \cite{islam2020fast}, WaterNet \cite{li2019underwater}, ADMNNet \cite{yan2022attention}, UColor \cite{li2021underwater}, UGAN \cite{fabbri2018enhancing}, U-Trans \cite{peng2023u}, UIE-WD \cite{ma2022wavelet} and SGUIE \cite{qi2022sguie}, respectively.

\subsubsection{Settings}
The learning rate of all networks in ColorCode is set to 0.0001. The values of $\lambda_1$, $\lambda_2$ and $\lambda_3$ are set to 10, 1 and 10, respectively. Adam optimizer is used, where $\beta_{1} = 0.5$ and $\beta_{2} = 0.999$. All content encoders, style encoders, decoders, and discriminators remain consistent with \cite{huang2018multimodal}. Compared to \cite{huang2018multimodal}, ColorCode adds a color encoder $E_{x}^{m}(\cdot)$ and enhancement decoder $P_{x}^{m}(\cdot)$ for supervised training, where $E_{x}^{m}(\cdot)$ and $P_{x}^{m}(\cdot)$ adopt the architecture of style encoder and decoder that proposed in \cite{huang2018multimodal}, respectively. The ColorCode proposed in this paper is an extension of CECF, that is, the CI capability obtained by the loss $\mathcal{L}_{x}^{cc}$ is newly added. We follow the implicit setting of CECF and use $\mathcal{L}_{x}^{cc}$ on the dataset that obtains color manipulation capabilities, namely the UFO-120 dataset. For other datasets, there is no need to use $\mathcal{L}_{x}^{cc}$ as will be discussed in Section \ref{subsec:what_kind_of_datasets_meet_the_needs_of_CECFCI}.

\subsection{Evaluation on Color Enhancement}
\label{subsec:experiment_color_enhancement}
We evaluate the proposed ColorCode and existing algorithms quantitatively and visually. It is worth pointing out that UIE datasets are synthesized, generated or selected in different ways, that is, perfect labels may not exist \cite{cong2024comprehensive}. 

\subsubsection{Quantitative Evaluation}
Quantitative evaluation results in Table \ref{tab:full_reference_results} show that the PSNR, SSIM and UIQM values obtained by the proposed ColorCode are impressive. On most of the datasets, ColorCode can rank in the top three.

\subsubsection{Visual Evaluation}
The enhancement results shown in Fig. \ref{fig:visual_results_uieb_2774} illustrate that the visual effects obtained by most UIE algorithms are acceptable. The ColorCode can generate visually beautiful enhancement results that contrast, color, textures and details are effectively enhanced.

\subsubsection{Conclusions} 
Quantitative and qualitative results show that ColorCode can achieve competitive color enhancement performance compared with existing state-of-the-art methods.

\subsection{Evaluation on Color Adaptation}
\label{subsec:experiment_color_fine_tuning}
\begin{figure*}[!t]
        \centering
        \footnotesize
        \textcolor{black}{\leftline{\hspace{0.4cm} $x \rightarrow \hat{y}_{x}^{m}$ \hspace{0.9cm}  $\alpha = 0.1$ \hspace{0.5cm}  $\alpha = 0.2$  \hspace{0.5cm}   $\alpha = 0.25$ \hspace{0.5cm}  $\alpha = 0.3$  \hspace{0.5cm}  $\alpha = 0.35$  \hspace{0.45cm} $\alpha = 0.4$  \hspace{0.5cm}   $\alpha = 0.45$  \hspace{0.45cm}  $\alpha = 0.5$ \hspace{1.3cm}  $g$}}
        \\

        \vspace{0.02cm}

        \includegraphics[width=1.65cm,height=1.4cm]{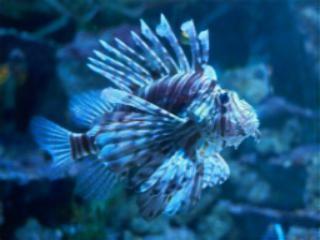}
        \includegraphics[width=0.2cm,height=1.4cm]{00_figures/materials/black_vertical_line.PNG}
        \includegraphics[width=1.65cm,height=1.4cm]{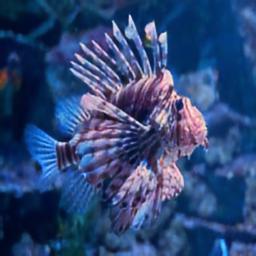}
        \includegraphics[width=1.65cm,height=1.4cm]{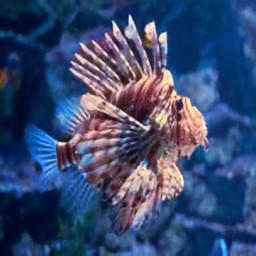}
        \includegraphics[width=1.65cm,height=1.4cm]{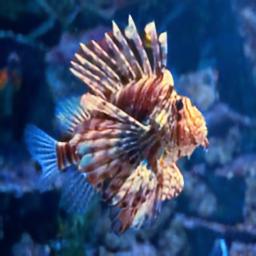}
        \includegraphics[width=1.65cm,height=1.4cm]{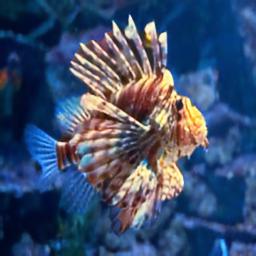}
        \includegraphics[width=1.65cm,height=1.4cm]{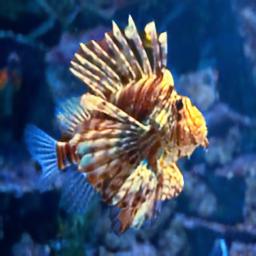}
        \includegraphics[width=1.65cm,height=1.4cm]{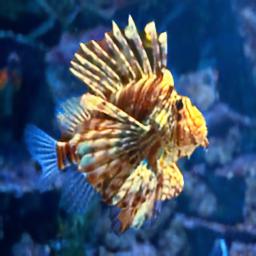}
        \includegraphics[width=1.65cm,height=1.4cm]{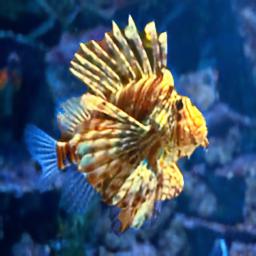}
        \includegraphics[width=1.65cm,height=1.4cm]{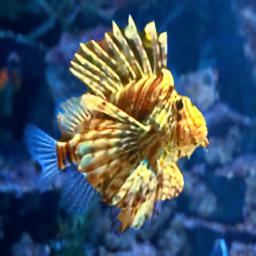}
        \includegraphics[width=0.2cm,height=1.4cm]{00_figures/materials/black_vertical_line.PNG}
        \includegraphics[width=1.65cm,height=1.4cm]{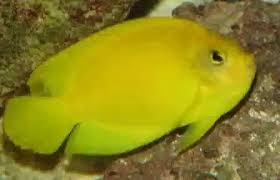}

        \vspace{0.1cm}

        \includegraphics[width=1.65cm,height=1.4cm]{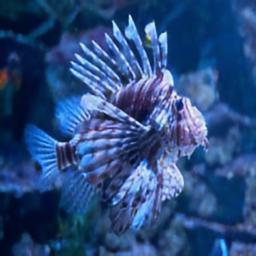}
        \includegraphics[width=0.2cm,height=1.4cm]{00_figures/materials/black_vertical_line.PNG}
        \includegraphics[width=1.65cm,height=1.4cm]{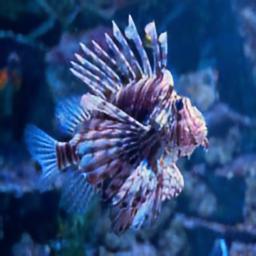}
        \includegraphics[width=1.65cm,height=1.4cm]{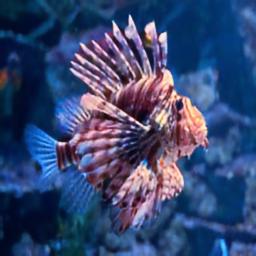}
        \includegraphics[width=1.65cm,height=1.4cm]{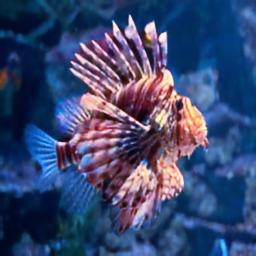}
        \includegraphics[width=1.65cm,height=1.4cm]{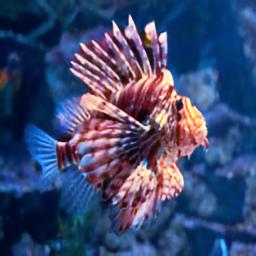}
        \includegraphics[width=1.65cm,height=1.4cm]{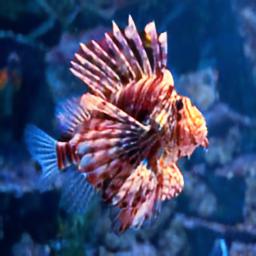}
        \includegraphics[width=1.65cm,height=1.4cm]{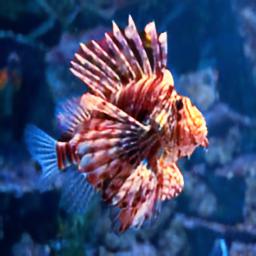}
        \includegraphics[width=1.65cm,height=1.4cm]{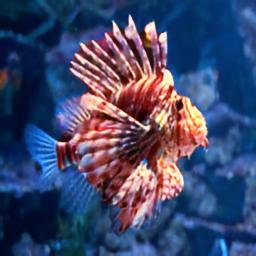}
        \includegraphics[width=1.65cm,height=1.4cm]{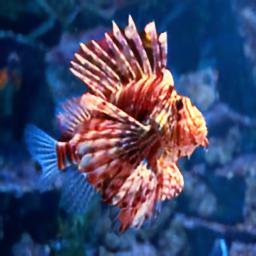}
        \includegraphics[width=0.2cm,height=1.4cm]{00_figures/materials/black_vertical_line.PNG}
        \includegraphics[width=1.65cm,height=1.4cm]{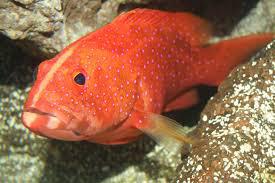}

        \textcolor{black}{(a) \textbf{Underwater} images are used as guidance for the color adaptation process.}
        \vspace{0.1cm}

        \includegraphics[width=1.65cm,height=1.4cm]{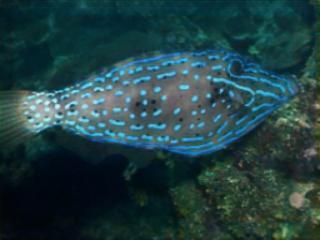}
        \includegraphics[width=0.2cm,height=1.4cm]{00_figures/materials/black_vertical_line.PNG}
        \includegraphics[width=1.65cm,height=1.4cm]{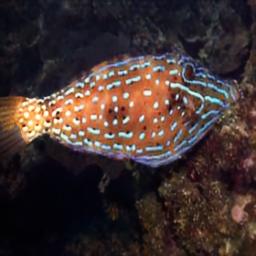}
        \includegraphics[width=1.65cm,height=1.4cm]{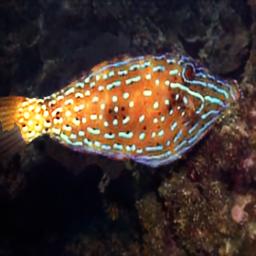}
        \includegraphics[width=1.65cm,height=1.4cm]{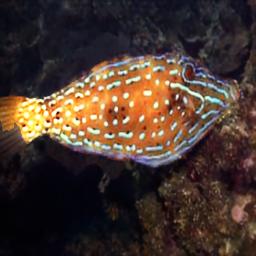}
        \includegraphics[width=1.65cm,height=1.4cm]{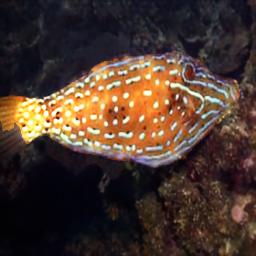}
        \includegraphics[width=1.65cm,height=1.4cm]{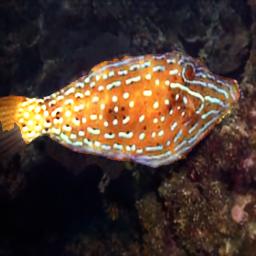}
        \includegraphics[width=1.65cm,height=1.4cm]{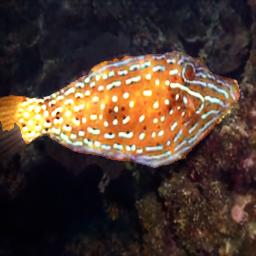}
        \includegraphics[width=1.65cm,height=1.4cm]{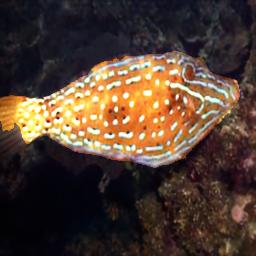}
        \includegraphics[width=1.65cm,height=1.4cm]{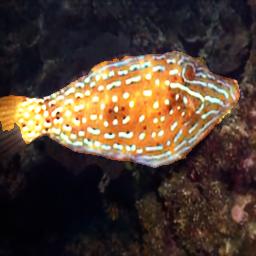}
        \includegraphics[width=0.2cm,height=1.4cm]{00_figures/materials/black_vertical_line.PNG}
        \includegraphics[width=1.65cm,height=1.4cm]{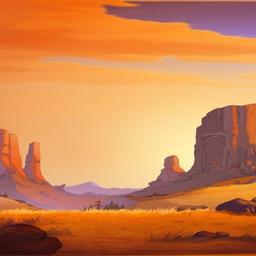}

        \vspace{0.1cm}

        \includegraphics[width=1.65cm,height=1.4cm]{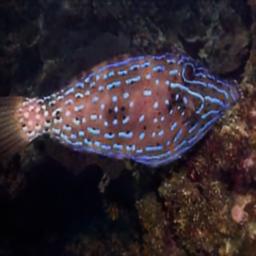}
        \includegraphics[width=0.2cm,height=1.4cm]{00_figures/materials/black_vertical_line.PNG}
        \includegraphics[width=1.65cm,height=1.4cm]{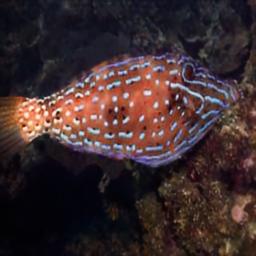}
        \includegraphics[width=1.65cm,height=1.4cm]{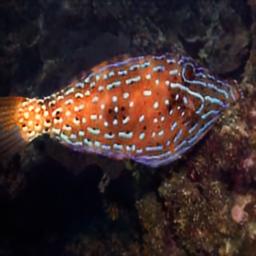}
        \includegraphics[width=1.65cm,height=1.4cm]{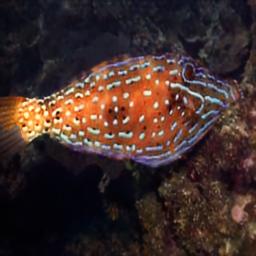}
        \includegraphics[width=1.65cm,height=1.4cm]{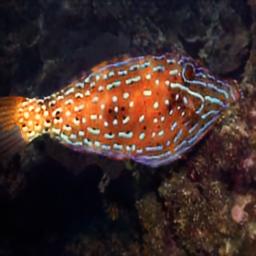}
        \includegraphics[width=1.65cm,height=1.4cm]{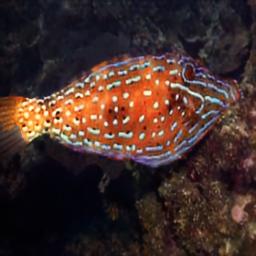}
        \includegraphics[width=1.65cm,height=1.4cm]{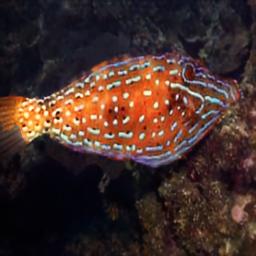}
        \includegraphics[width=1.65cm,height=1.4cm]{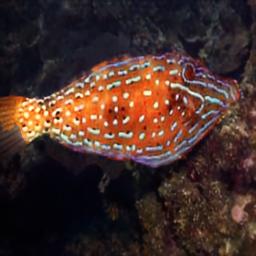}
        \includegraphics[width=1.65cm,height=1.4cm]{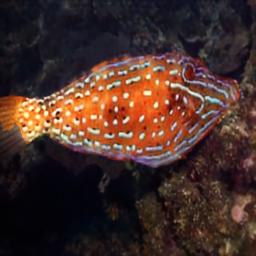}
        \includegraphics[width=0.2cm,height=1.4cm]{00_figures/materials/black_vertical_line.PNG}
        \includegraphics[width=1.65cm,height=1.4cm]{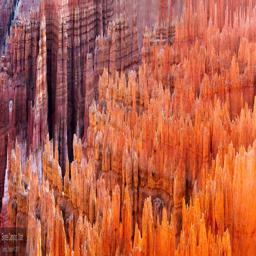}
        
        \textcolor{black}{(b) \textbf{Natural} images are used as guidance for the color adaptation process.}

        \caption{Visualization of ColorCode's color adaptation capability. The $\alpha$ represents the weight parameter of the color guidance function $\mathcal{F}_{ca}(x, y, \alpha)$. For the first image of every two rows, the original distorted underwater image $x$ and original enhanced result $\hat{y}_{x}^{m}$ are images at the top and bottom, respectively.}
        \label{fig:guided_by_two_kinds_images}
\end{figure*}

Beyond obtaining enhanced underwater organisms with deterministic colors, we hope that the colors of underwater organisms can be adapted by guidance. Based on a simple and intuitive observation that the hue of colors with long wavelengths is approximately unchanged during enhancement, we can use underwater images or natural images as guidance to diversify the colors of an enhanced underwater organism. Fig. \ref{fig:guided_by_two_kinds_images}(a) and \ref{fig:guided_by_two_kinds_images}(b) show the visual effects obtained by using underwater images and natural images as guidance in multiple scenes. As we can observe, the color of underwater organism $x$ shifts toward the tone of the guidance $g$ under the color control function $\mathcal{F}_{ca}(x, y, \alpha)$. As $\alpha$ increases, the degree of color shift increases. Not only that, under different guidance with long-wavelength colors, the obtained color shift effects are also different. In conclusion, the visualization results of Fig. \ref{fig:guided_by_two_kinds_images} show that the proposed ColorCode has the ability to make the colors of underwater organisms adapted to the guidance.


\subsection{Evaluation on Color Interpolation}
\label{subsec:experiment_color_interpolation}
Beyond adopting guidance to adjust the colors, here we try to expand the adjustment range of colors to a larger color space. Since the color code has been constrained to conform to the Gaussian distribution, we can sample color codes from $N(0, 1)$ distribution at equal intervals. The sampled value range is $[-5,5]$. When the dimension of the color code exceeds 2, the relationship between numerical information and visual effects cannot be intuitively displayed in a 2D plane. Therefore, we trained a model with the dimension of the color code equal to 2. The sampled color code $m_{z}$ is fused with the original color code $m_{x}$ to obtain $\hat{y}^{x,z}_{m}$, where the fusion weight $\alpha$ is set to 0.5. The visual results are shown in Fig. \ref{fig:color_code_interpolation}. We can observe that the changes in colors are approximately continuous. As the value of the sampled $m_{z}$ changes, the color of the organism moves in different directions. In conclusion, the visual effect shown in Fig. \ref{fig:color_code_interpolation} proves that our proposed ColorCode can perform color interpolating.

\begin{figure}
        \small
        \centering
        \includegraphics[width=8.5cm,height=3cm]{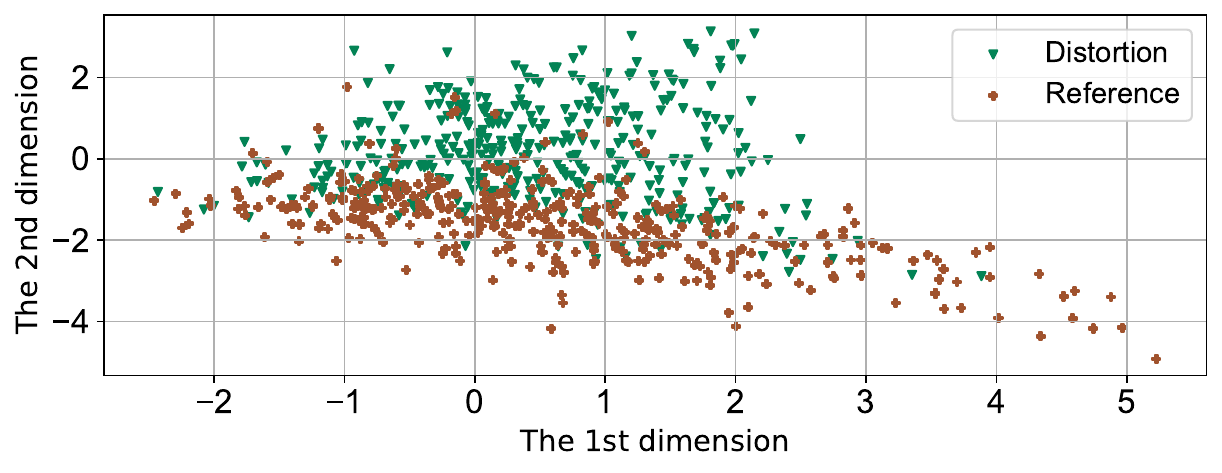}
        \caption{The distribution of distorted images and reference images in the 2D encoding space. The dimension of the color code is set to 2 during training.}
        \label{fig:mapping_code_distortion_and_reference_EUVP}
\end{figure}

\subsection{Ablation Study and Discussions}
\label{subsec:ablation_study}
\subsubsection{The Length of the Color Code}
Color code of distorted underwater image $x$ is one of the main concepts in the proposed algorithm. Therefore, we train models with color codes of different lengths. We evaluate the corresponding enhancement performance as shown in Table \ref{tab:results_differ_length_enhancement_code}. Within the value range we set, the length of the color code $m_{x}$ does not have a significant impact on the performance of the model. Therefore, we empirically set the length of $m_{x}$ to 8.

\begin{table}[t]
        \small
        \centering
        \setlength{\tabcolsep}{1.0mm}
        \renewcommand{\arraystretch}{1.2}
        \caption{Quantitative evaluation results obtained when the length (denoted by $l$) of color code $m_{x}$ is set to different values. The values in the table represent increases compared to $l=8$. The best results are in bold.}
        \label{tab:results_differ_length_enhancement_code}
        \begin{tabular}{ccccccc}
          \hline
          \multirow{2}{*}{Config.} & \multicolumn{3}{c}{UIEB} & \multicolumn{3}{c}{EUVP} \\
          \cline{2-7}
          & PSNR$\uparrow$ & SSIM$\uparrow$ & UIQM$\uparrow$ & PSNR$\uparrow$ & SSIM$\uparrow$ & UIQM$\uparrow$ \\
          \hline
          $l = 16$  & -0.1029 & 0.0023  & -0.0473 & -0.0341 & 0.0012  & 0.0491 \\
          $l = 32$  & -0.1103 & 0.0017  & -0.0396 &  \textbf{0.0500} & \textbf{0.0041}  & -0.0573 \\
          $l = 48$  & -0.1431 & \textbf{0.0026}  &  \textbf{0.0306} & -0.0405 & 0.0036  & \textbf{0.0762} \\
          $l = 64$  &  \textbf{0.0310} & -0.0039 & -0.0491 & -0.0724 & -0.0021 & 0.0617 \\
          \hline
        \end{tabular}
      \end{table}


\subsubsection{The Evaluation of the Interpolation Center}
As we point out in Section \ref{subsec:explicit_constraint_on_the_color_code}, constraining the color code can obtain color interpolation capabilities. One factor worth verifying is whether the interpolation center affects the enhancement performance in terms of quantitative evaluation. As shown in Fig. \ref{fig:differ_gaussian_mean}, we set the interpolation center to $\{-1, -0.5, 0, 0.5, 1, 1.5, 2\}$. The quantitative evaluation results obtained at different interpolation positions vary to a certain extent. The results show that the best interpolation center is 1.

\subsubsection{The Evaluation of Structure Loss}
We conducted an experiment on the structure loss (SL) in Eq. \ref{eq:enhancement_loss}, which is to compare the changes in quantitative evaluation metrics before and after the use of the SL. As shown in Table \ref{tab:results_no_ssim}, after removing the SL, the results of the quantitative evaluation are generally reduced. Therefore, the SL is used in our final setting.

\begin{figure*}
        \small
        \centering
        \includegraphics[width=18cm,height=10cm]{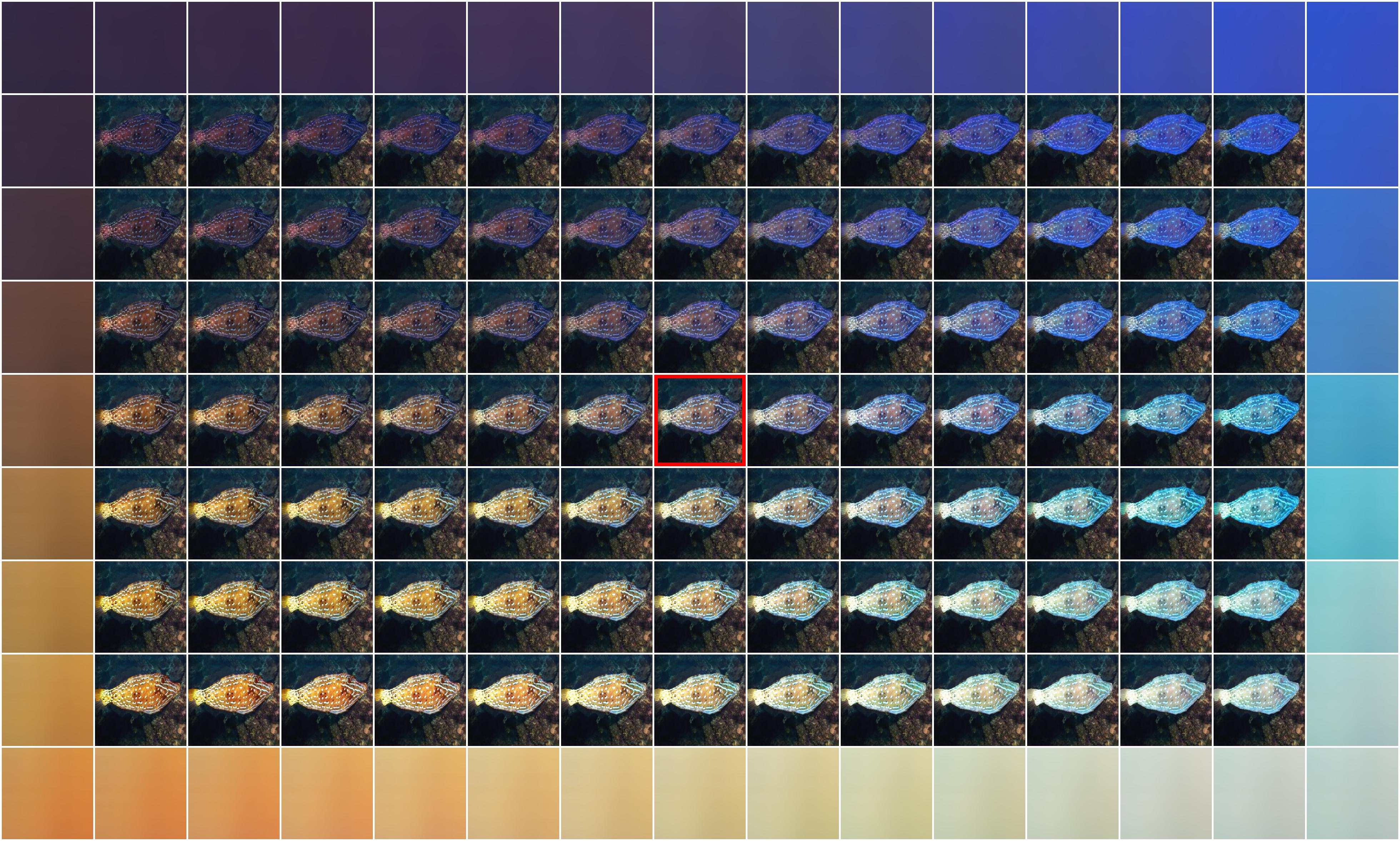}
        \caption{Diversification results obtained by using the color interpolation function $\mathcal{F}_{ci}(x,z)$. The image with a red border is the originally enhanced result with fixed color output by the enhancement network. The color blocks on the periphery represent the main colors of the organisms in the outermost image. The acquisition of color block is achieved by calculating the center of mass of the organism and the surrounding main color.}
        \label{fig:color_code_interpolation}
      \end{figure*}

\begin{table}[t]
        \small
        \centering
        \setlength{\tabcolsep}{1.0mm}
        \renewcommand{\arraystretch}{1.2}
        \caption{The change of metrics after removing the structure loss (SL). Bold values represent values that are boosted.}
        \label{tab:results_no_ssim}
        \begin{tabular}{ccccccc}
          \hline
          \multirow{2}{*}{Config.} & \multicolumn{3}{c}{UIEB} & \multicolumn{3}{c}{EUVP} \\
          \cline{2-7}
          & PSNR$\uparrow$ & SSIM$\uparrow$ & UIQM$\uparrow$ & PSNR$\uparrow$ & SSIM$\uparrow$ & UIQM$\uparrow$ \\
          \hline
          No-SL        & -0.1064 & -0.0139 & -0.0210 & -0.1103 & -0.0029 & \textbf{0.2147}  \\
          \hline
        \end{tabular}
\end{table}


\subsubsection{Analysis of the Constraint on the Color Code}
\label{subsubsec:analysis_of_the_constraint_on_color_code}
We hope the color code conforms to a specific distribution. During the training process, the color code is constrained by the loss $\mathcal{L}_{x}^{cc}$. To prove that $\mathcal{L}_{x}^{cc}$ is effective, we visualize the color code distribution after 80k iterations under three settings, which are
\begin{itemize}
        \item No constraint on the color code.
        \item Constrain the color code by $N(0, 1)$.
        \item Constrain the color code by different $\mu$ in $N(\mu, 1)$. 
\end{itemize}

The purpose of the above three settings is to prove
\begin{itemize}
        \item Without any constraints, each dimension of the color code will not conform to the same statistical distribution.
        \item Under a certain constraint, the color code can conform to the desired Gaussian distribution.
\end{itemize}

\begin{figure}
        \small
        \centering
        \includegraphics[width=2.1cm,height=1.5cm]{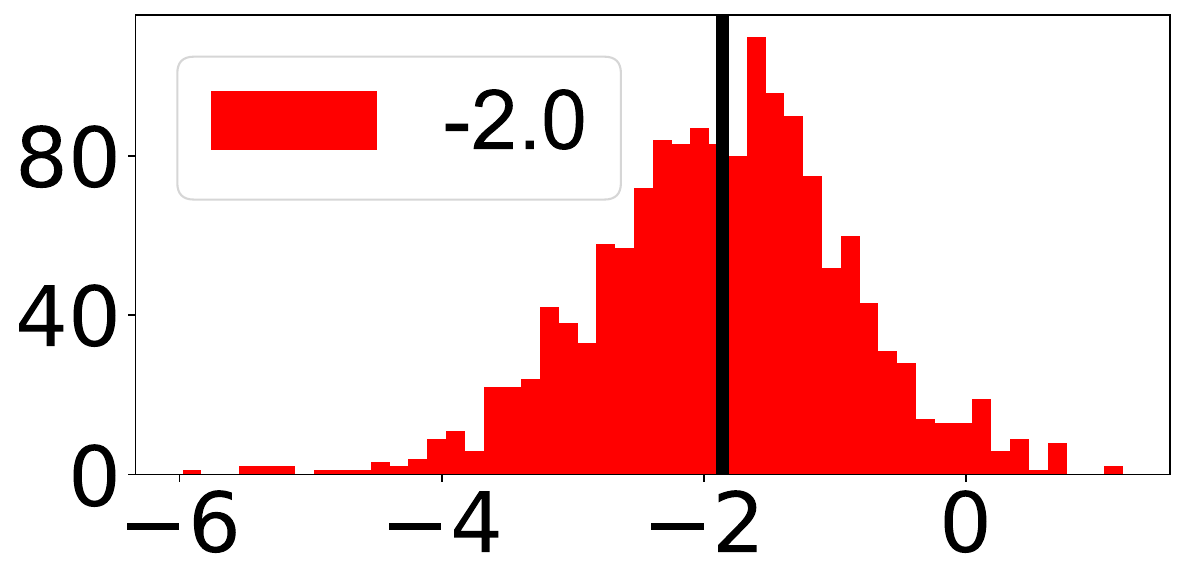}
        \includegraphics[width=2.1cm,height=1.5cm]{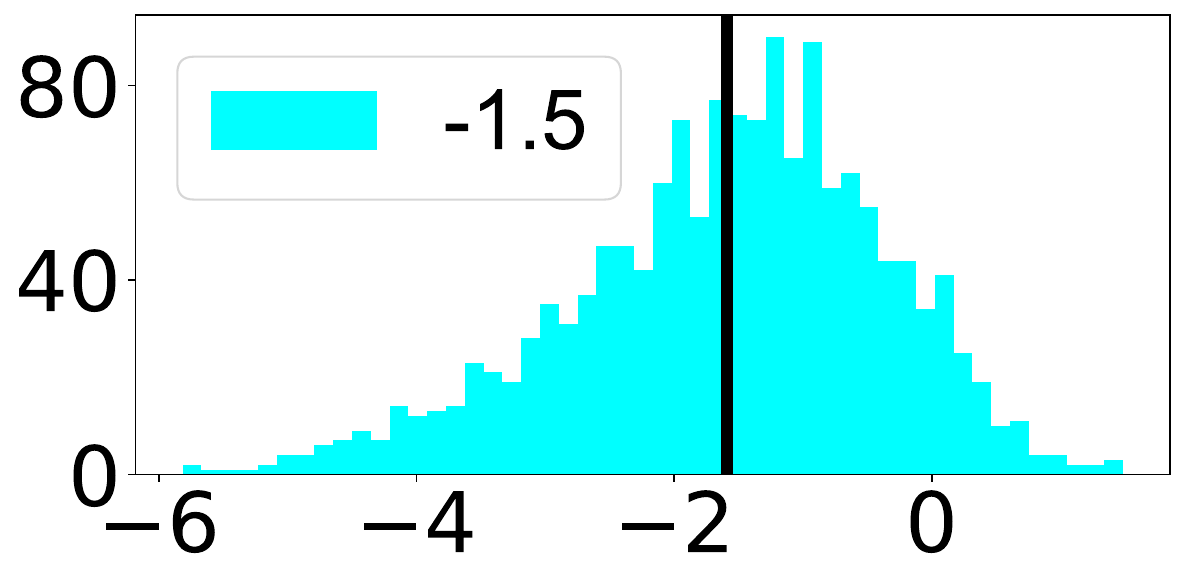}
        \includegraphics[width=2.1cm,height=1.5cm]{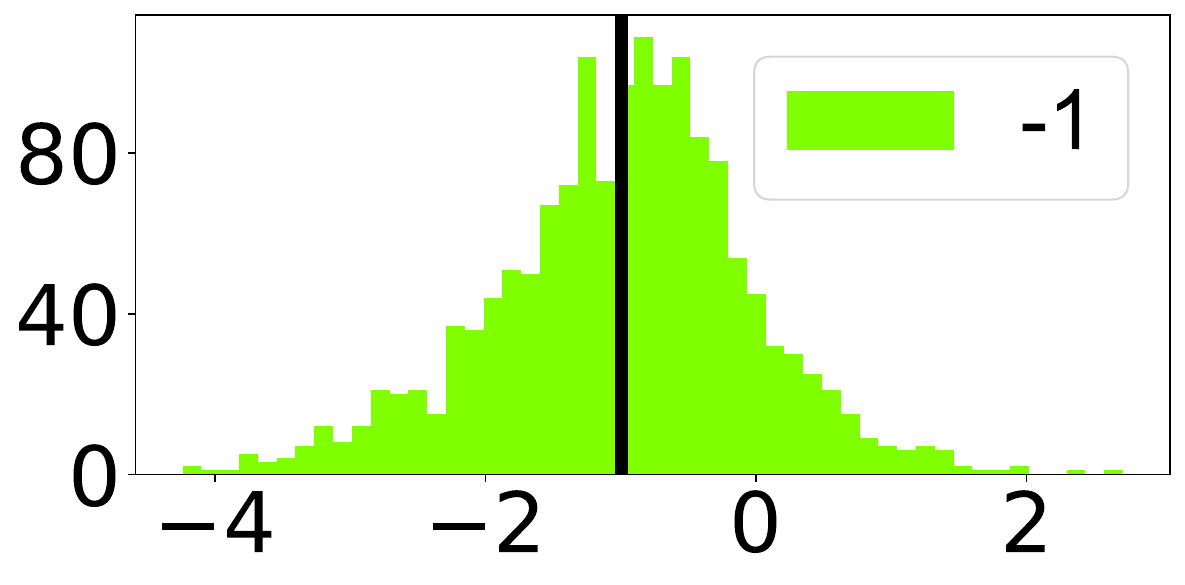}
        \includegraphics[width=2.1cm,height=1.5cm]{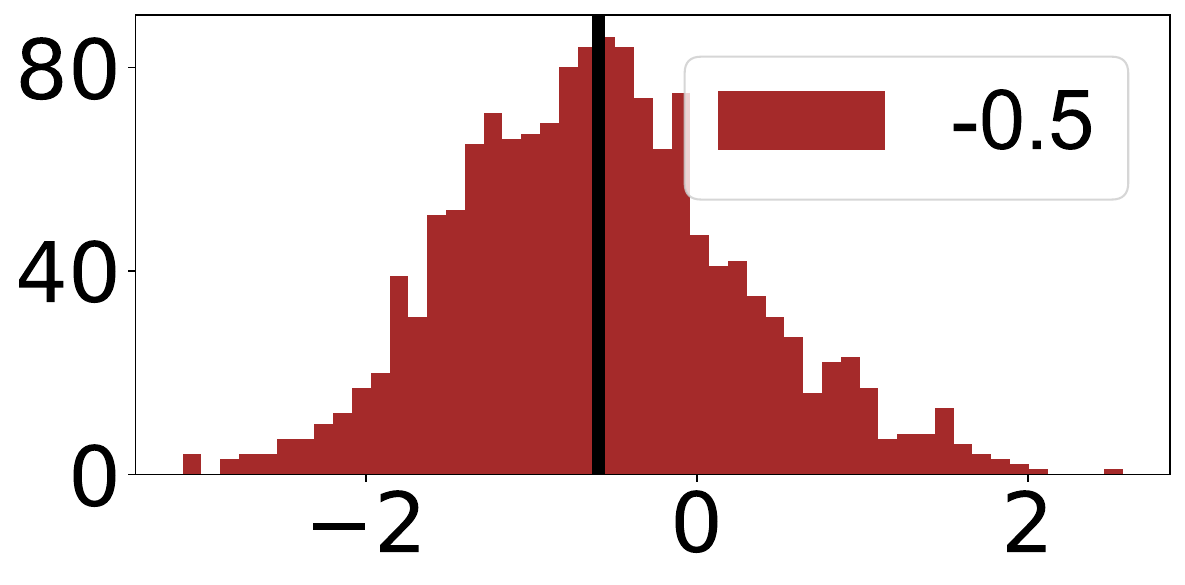}

        \includegraphics[width=2.1cm,height=1.5cm]{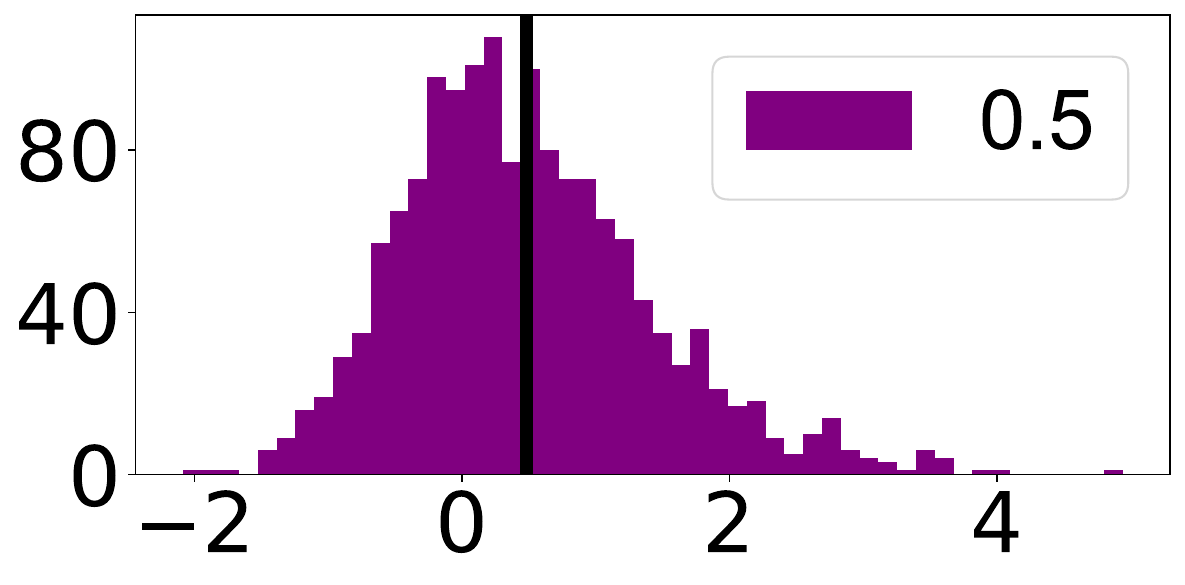}
        \includegraphics[width=2.1cm,height=1.5cm]{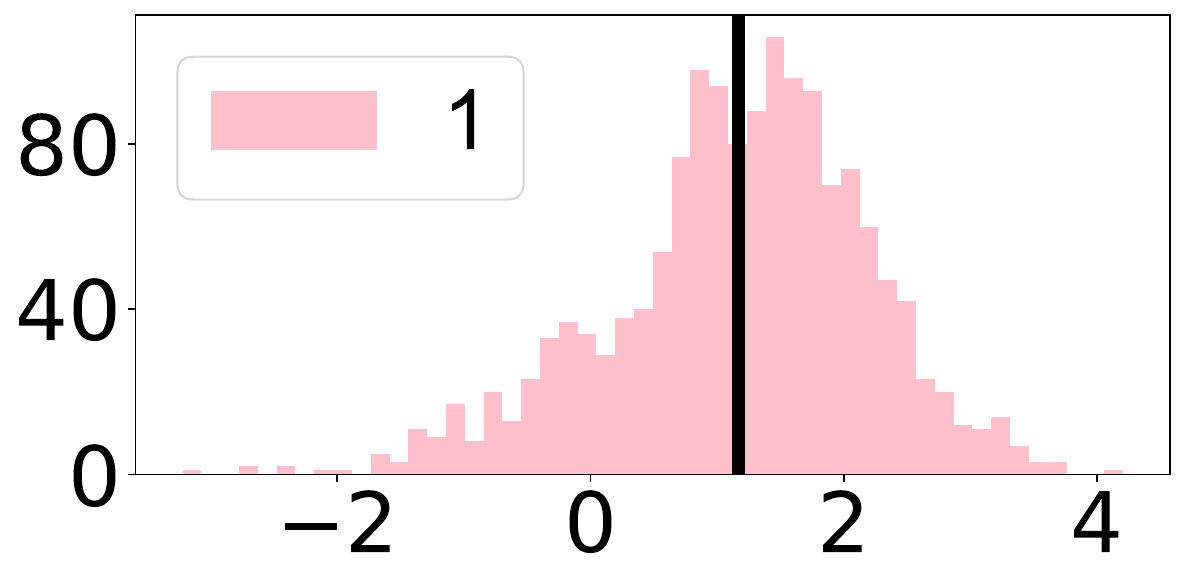}
        \includegraphics[width=2.1cm,height=1.5cm]{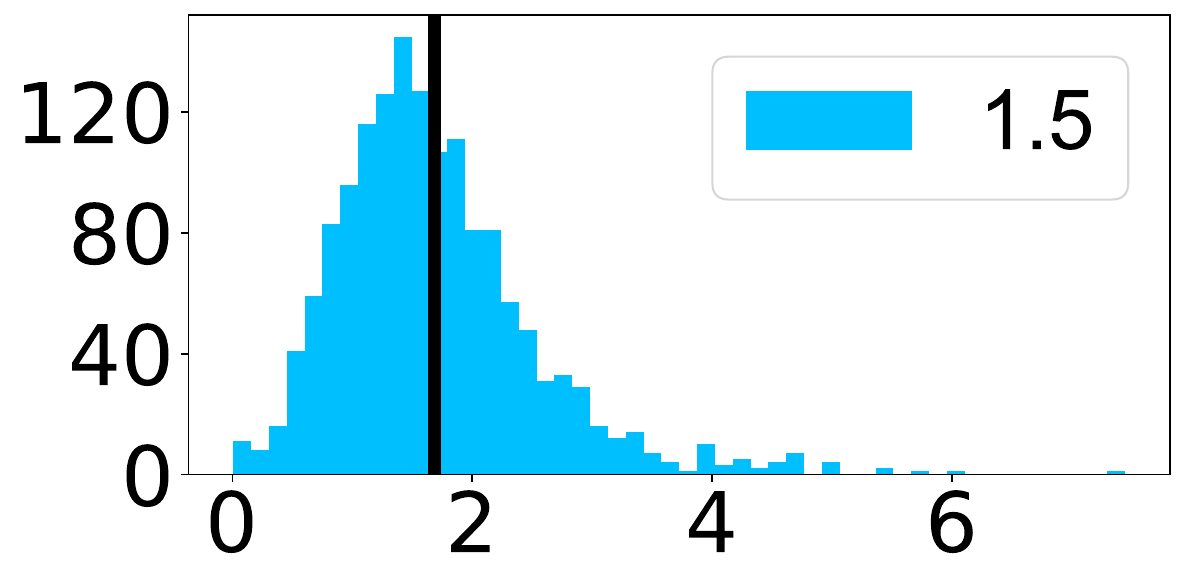}
        \includegraphics[width=2.1cm,height=1.5cm]{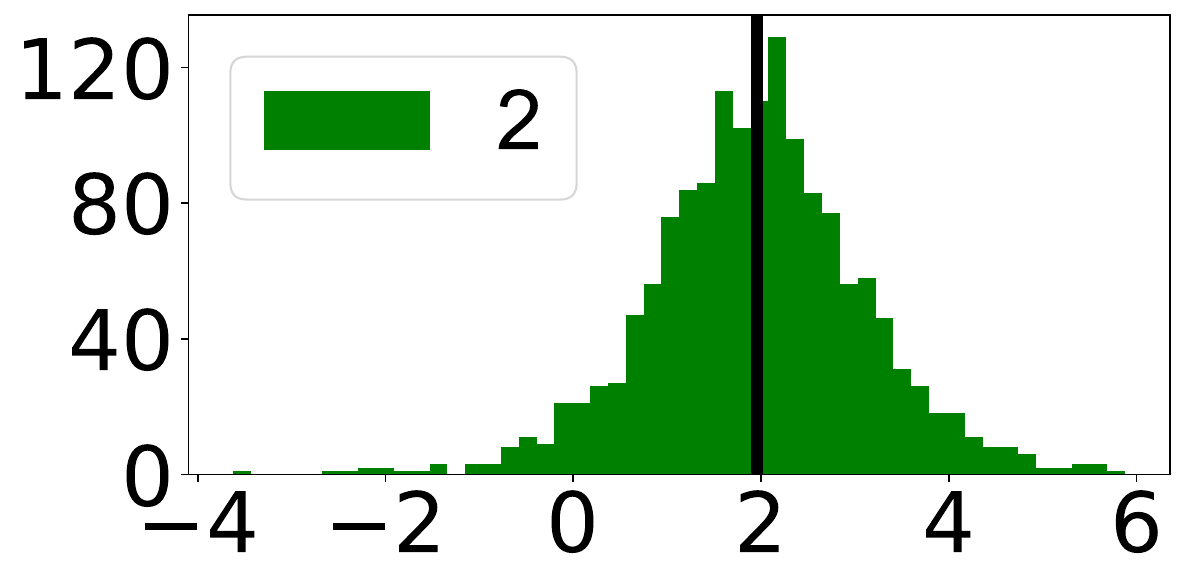}
        \caption{The distribution histogram obtained by constraining $m_{x}$ by different $\mu$ in $N(\mu, \sigma^{2})$, where the $x$ and $y$ axis represent the value range and amount, respectively. The value of $\mu$ is given in the legend. The black vertical line represents the mean of the data.}
        \label{fig:compare_of_color_code_after_constraint}
\end{figure}

Fig. \ref{fig:color_code_without_any_constraints} and Fig. \ref{fig:compare_of_color_code_after_constraint} show the distribution histogram of the color codes under three settings. As shown in Fig. \ref{fig:color_code_without_any_constraints}(a), without any restrictions, the distribution intervals of different dimensions of the color code are inconsistent. This means that obtaining diversified color codes with continuous semantics by sampling cannot be directly achieved. When the $N(0, 1)$ constraint is added, the distribution histogram shown in Fig. \ref{fig:color_code_without_any_constraints}(b) proves that the distribution of different dimensions of the color code is approximately consistent. The mean of the color code in each dimension is around 0. Fig. \ref{fig:compare_of_color_code_after_constraint} shows that when different constraints are used, the mean of the color code tends to be changed accordingly. These results prove the effectiveness of the $\mathcal{L}_{x}^{cc}$.

\begin{figure}[!t]
        \centering
        \footnotesize
        \textcolor{black}{\leftline{\hspace{0.4cm} $\alpha = 0$ \hspace{0.2cm}  $\alpha = 0.05$  \hspace{0.2cm}   $\alpha = 0.1$ \hspace{0.2cm}  $\alpha = 0.2$  \hspace{0.2cm}  $\alpha = 0.3$  \hspace{0.9cm}  $g$}}
        \\
        \includegraphics[width=1.3cm,height=1.3cm]{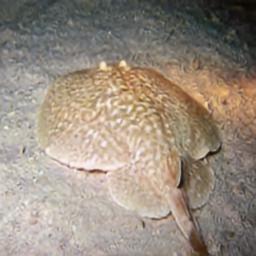}
        \includegraphics[width=1.3cm,height=1.3cm]{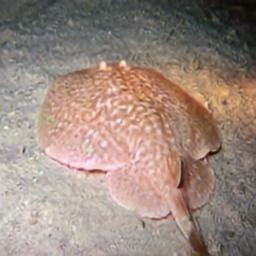}
        \includegraphics[width=1.3cm,height=1.3cm]{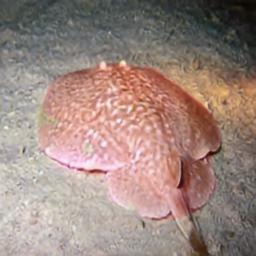}
        \includegraphics[width=1.3cm,height=1.3cm]{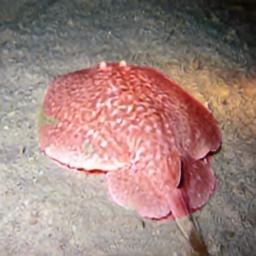}
        \includegraphics[width=1.3cm,height=1.3cm]{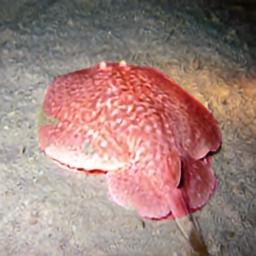}
        \includegraphics[width=0.2cm,height=1.3cm]{00_figures/materials/black_vertical_line.PNG}
        \includegraphics[width=1.3cm,height=1.3cm]{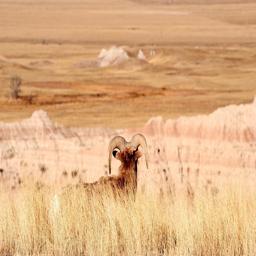}

        {(a) Color Adaptation results under a dataset \textbf{without} slight distortions.}

        \vspace{0.1cm}

        \includegraphics[width=1.3cm,height=1.3cm]{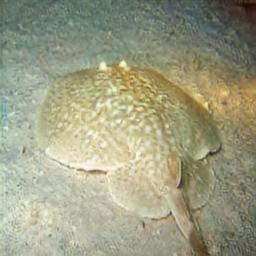}
        \includegraphics[width=1.3cm,height=1.3cm]{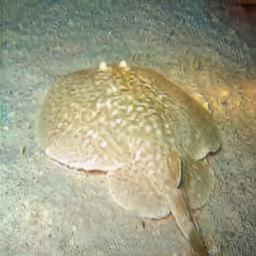}
        \includegraphics[width=1.3cm,height=1.3cm]{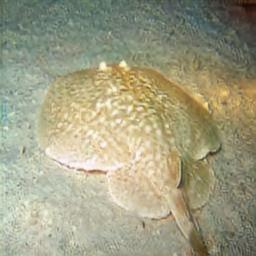}
        \includegraphics[width=1.3cm,height=1.3cm]{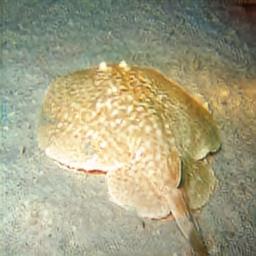}
        \includegraphics[width=1.3cm,height=1.3cm]{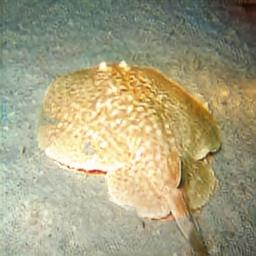}
        \includegraphics[width=0.2cm,height=1.3cm]{00_figures/materials/black_vertical_line.PNG}
        \includegraphics[width=1.3cm,height=1.3cm]{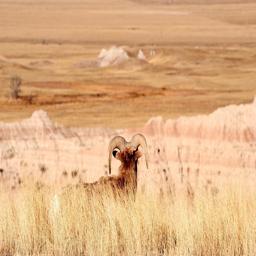}

        \vspace{0.1cm}

        {(b) Color Adaptation results under a dataset \textbf{with} slight distortions.}

        \caption{Comparison of color adaptation results obtained with and without slight distortion in the dataset. (a) represents unsuccessful color adaptation results, while (b) denotes successful color adaptation results.}
        \label{fig:what_kinds_of_dataset_meets_our_need}
\end{figure}

\begin{figure*}
        \small
        \centering

        \leftline{\hspace{1cm}  $g_{1}$  \hspace{1.6cm}  $\hat{g}_{1}$  \hspace{1.8cm}  $g_{2}$  \hspace{1.6cm}  $\hat{g}_{2}$ \hspace{2cm} $g_{3}$  \hspace{1.6cm}  $\hat{g}_{3}$  \hspace{1.7cm}  $g_{4}$  \hspace{1.7cm}  $\hat{g}_{4}$}
        \vspace{0.06cm}
        \includegraphics[width=2.1cm,height=1.7cm]{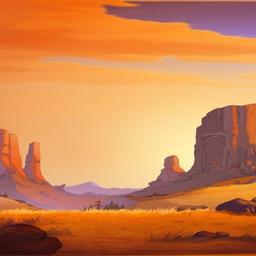}
        \includegraphics[width=2.1cm,height=1.7cm]{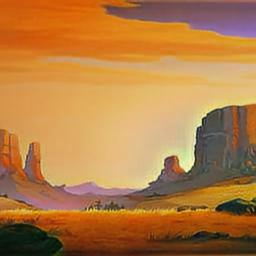}
        \includegraphics[width=2.1cm,height=1.7cm]{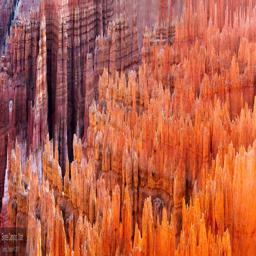}
        \includegraphics[width=2.1cm,height=1.7cm]{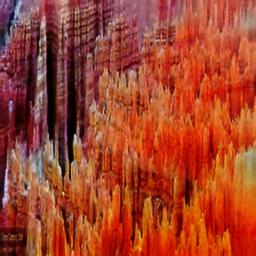}
        \includegraphics[width=0.2cm,height=1.7cm]{00_figures/materials/black_vertical_line.PNG}
        \includegraphics[width=2.1cm,height=1.7cm]{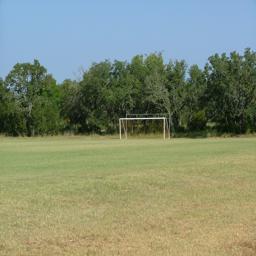}
        \includegraphics[width=2.1cm,height=1.7cm]{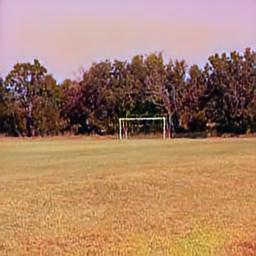}
        \includegraphics[width=2.1cm,height=1.7cm]{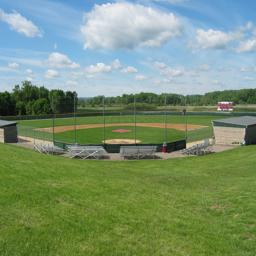}
        \includegraphics[width=2.1cm,height=1.7cm]{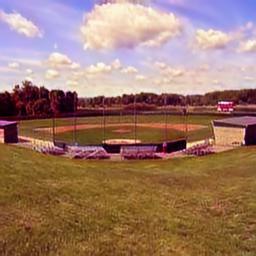}

        \vspace{0.1cm}

        \includegraphics[width=2.1cm,height=1.7cm]{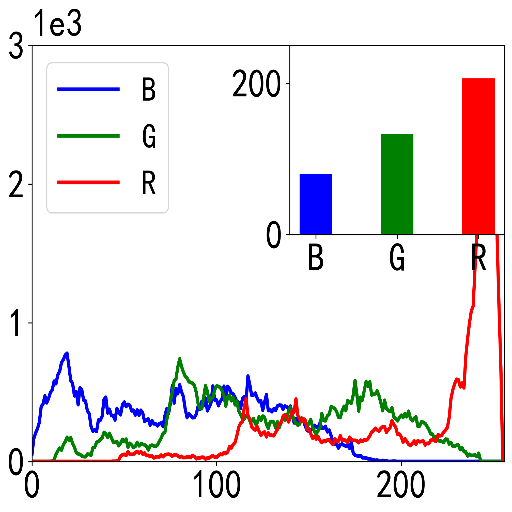}
        \includegraphics[width=2.1cm,height=1.7cm]{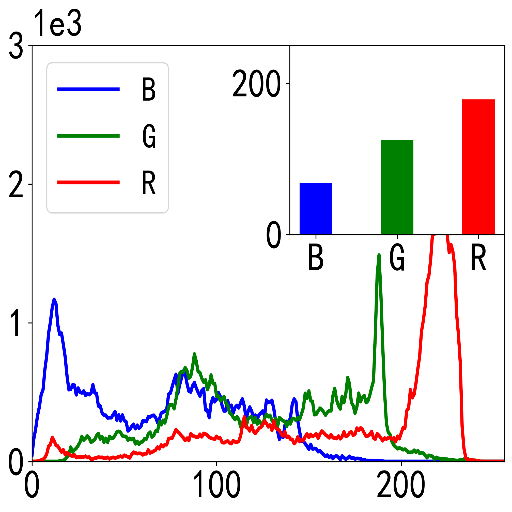}
        \includegraphics[width=2.1cm,height=1.7cm]{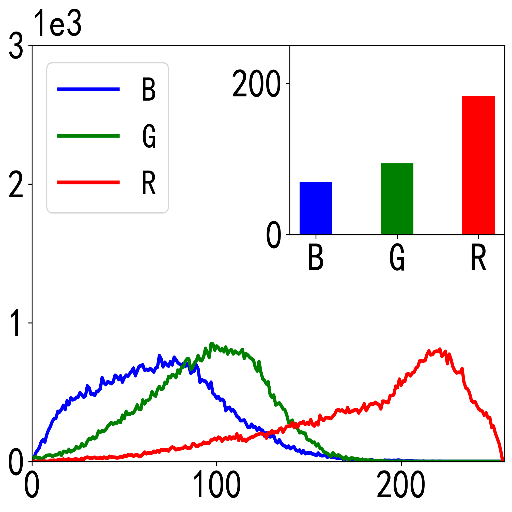}
        \includegraphics[width=2.1cm,height=1.7cm]{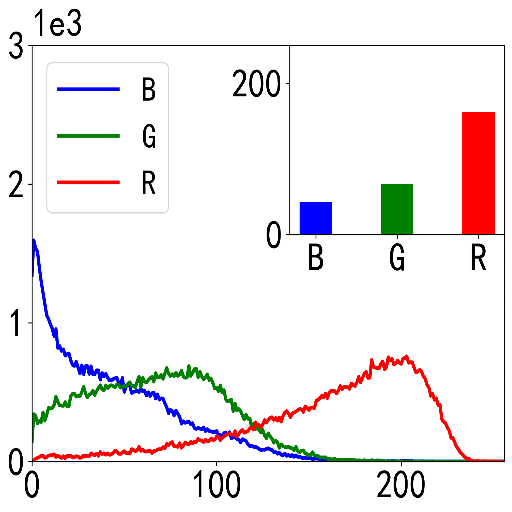}
        \includegraphics[width=0.2cm,height=1.7cm]{00_figures/materials/black_vertical_line.PNG}
        \includegraphics[width=2.1cm,height=1.7cm]{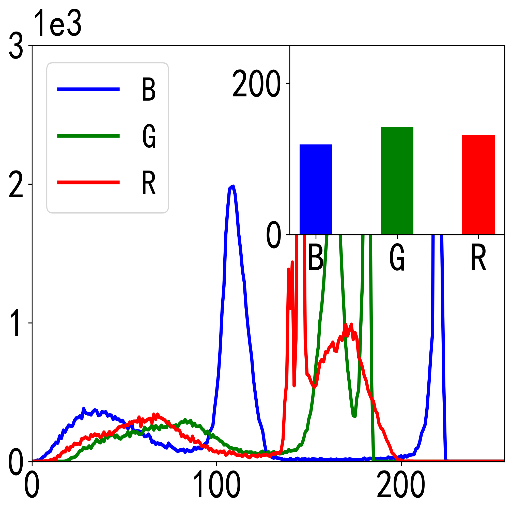}
        \includegraphics[width=2.1cm,height=1.7cm]{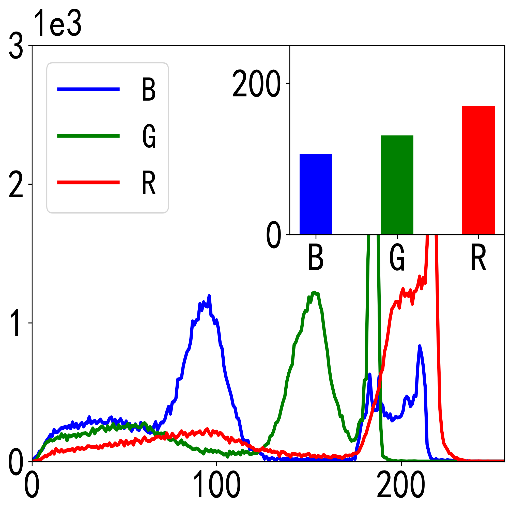}
        \includegraphics[width=2.1cm,height=1.7cm]{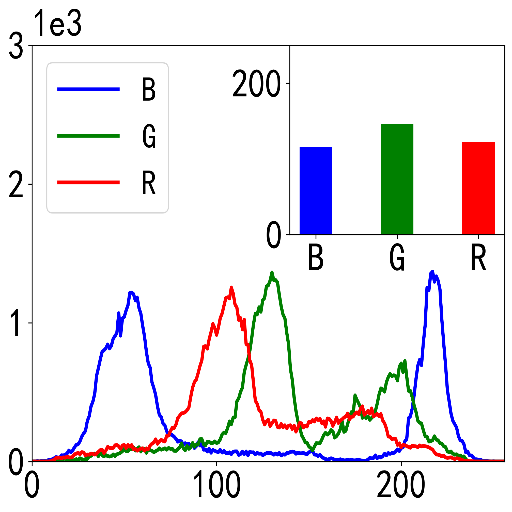}
        \includegraphics[width=2.1cm,height=1.7cm]{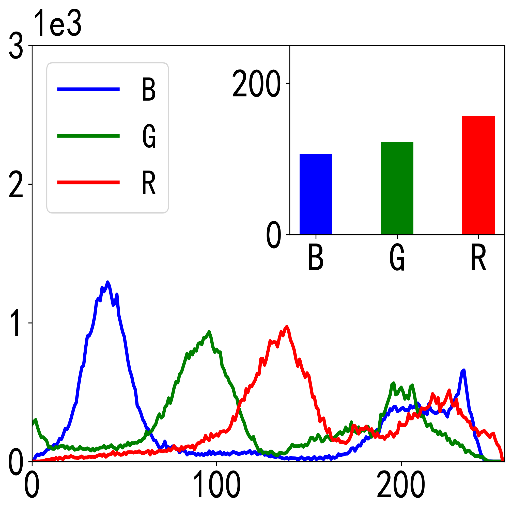}
        
        (a) Images with \textbf{long} wavelength colors. \hspace{3cm} (b) Images with \textbf{short} wavelength colors.

        \caption{The change of hue of the guidance. The $\{g_{1}$, $g_{2}$, $g_{3}, g_{4}\}$ and $\{\hat{g}_{1}$, $\hat{g}_{2}$, $\hat{g}_{3}, \hat{g}_{4}\}$ represent images before and after enhancement, respectively. There are no natural images involved in the training process, that is, $g_{1}$, $g_{2}$, $g_{3}$ and $g_{4}$ do not come from the distribution of training data.}
        \label{fig:plot_the_invariance_of_hue_of_long_wavelength_color}
\end{figure*}

\subsubsection{Retention of Content Information} 
As pointed out in Section \ref{sec:introduction} and Section \ref{sec:methods}, we hope that the content of the image remains unchanged during the process of color adaptation and color interpolation. Fig. \ref{fig:guided_by_two_kinds_images} and Fig. \ref{fig:color_code_interpolation} provide the evaluation of color manipulation for multiple scenes. The results show that the contents of the images are unchanged. These results prove that the decomposition process of image content and color is reliable.

\subsubsection{Quantitative Evaluation of Color Adaptation}
During the color adaptation process, as $\alpha$ increases, the degree to which the color of the enhanced image shifts toward the color of the guidance increases. To quantitatively analyze whether the image quality changes during the color adaptation process, the value of $UIQM(\mathcal{F}_{ca}(x, g, \alpha)) - UIQM(\mathcal{F}_{ca}(x, x, 0))$ is calculated. As shown in Fig. \ref{fig:eval_NR_differ_guidence_images_UIEB}, for the case where natural and underwater images are used as guidance, the numerical trend of UIQM as $\alpha$ increases are consistent. Numerical evaluation results show that (i) color adaptation can change the quality of the enhanced image, (ii) different guidance changes the quality of the enhanced image to different degrees, and (iii) the quantitative evaluation of color adaptation performs best when $\alpha$ belongs to the interval $[0, 0.3]$.

\subsubsection{Comparison of the Distribution of the Color Code with and without Distortion}
As we analyzed in Section \ref{subsec:experiment_color_interpolation}, by interpolating and changing the value of the color code, we can obtain images with different colors. This phenomenon means that different values of the color code have different semantics during the decoding process. Therefore, a question worth analyzing is whether images from different domains can obtain different color codes during the encoding process. For visualization, we set the dimension of the color code to 2. Fig. \ref{fig:mapping_code_distortion_and_reference_EUVP} shows numerical distributions of distorted images and reference images in the encoding space. The results intuitively show that the color codes of distorted images and reference images are distributed in different positions in the encoding space. For the distorted images, their encoded values are around 0, since the color codes of the distorted images are constrained by $N(0, 1)$ during training. For the reference images with sufficient long-wavelength colors, their values are distributed below the horizontal axis.

\subsubsection{Analysis on Why the Style Code cannot be Directly Used for the UIE Task}
As we pointed out in Section \ref{sec:methods}, the style code cannot be directly used in our UIE task. To be specific, we refer to the style code studied in \cite{huang2018multimodal}. Fig. \ref{fig:visualization_of_main_color_map_of_color_code_and_style_code}(a) and Fig. \ref{fig:visualization_of_main_color_map_of_color_code_and_style_code}(b) show that both the style code and color code may obtain diverse color results. However, the images obtained by style code are not as natural as those obtained by color code. Another important reason is that the enhanced image obtained by the style code may not be close to the desired reference. During the training process for the style code, no explicit loss is used to obtain a deterministic enhancement result. As shown in Fig. \ref{fig:visualization_of_main_color_map_of_color_code_and_style_code}(c), the center of the diverse interpolation results obtained by the color code is close to the reference, that is, the enhancement result obtained directly through supervised training. However, the center of the enhanced result obtained by the style code is not sufficiently close to the reference.

\subsubsection{Hue Approximate Invariance of Colors with Long-wavelength}
As we discuss in Section \ref{subsec:color_guidance_by_the_image_with_long_wavelength_color}, if the proposed ColorCode has the capability to adjust the colors of underwater organisms, the colors with long wavelengths of the guidance should be approximately invariant during the enhancement process. By sending the guidance $g$ into $P_{x}^{m}(E_{x}^{c}(\cdot), E_{x}^{m}(\cdot))$, the enhanced image $\hat{g}$ can be obtained. On the one hand, as shown in Fig. \ref{fig:plot_the_invariance_of_hue_of_long_wavelength_color}(a), there is no obvious difference among each color component in $g_{1}$ and $g_{2}$ before and after enhancement, although the color of local areas may change to a certain extent. In general, the $P_{x}^{m}(E_{x}^{c}(\cdot), E_{x}^{m}(\cdot))$ in ColorCode can ensure that colors with long wavelengths approximately retained in the color code $m_{g}$. On the other hand, as shown in Fig. \ref{fig:plot_the_invariance_of_hue_of_long_wavelength_color}(b) shows scenes where the overall hue tends to be green or blue-green. The hue of the $g_{3}$ and $g_{4}$ has obvious changes. For these colors with short wavelengths, the $P_{x}^{m}(E_{x}^{c}(\cdot), E_{x}^{m}(\cdot))$ may not preserve them.

\subsubsection{How to Choose the Guidance}
The main color of the guidance should belong to the hue with a long wavelength. Apart from this, the selection of guidance is not subject to any strict conditions. Users can put a selected guidance $g$ into $P_{x}^{m}(E_{x}^{c}(\cdot), E_{x}^{m}(\cdot))$ to obtain the enhanced $\hat{g}$. If $\hat{g}$ approximately matches the observation of Fig. \ref{fig:plot_the_invariance_of_hue_of_long_wavelength_color}(a), then it can be used for color adaptation.

\subsubsection{What Kind of Datasets Meet the Needs of ColorCode?}
\label{subsec:what_kind_of_datasets_meet_the_needs_of_CECFCI}
As we mentioned in Section \ref{sec:methods}. In order to achieve the ability of color adaptation, ColorCode needs to learn the approximate invariance of the hue of long-wavelength colors. From a data-driven perspective, images with slight and local distortions as shown in Fig. \ref{fig:illustration_of_the_invariant_of_the_hue_of_long_wavelength_colors} are needed. Therefore, a question that needs to be discussed is how the model performs when the dataset does not contain slightly distorted images. We chose EUVP (image-net) as a dataset without slight distortions. As shown in Fig. \ref{fig:what_kinds_of_dataset_meets_our_need}, when there is no slight distortion in the dataset, the direction of color adaptation does not move towards the guidance $g$. Fig. \ref{fig:what_kinds_of_dataset_meets_our_need} demonstrate that the color adaptation capability requires the dataset to contain slightly distorted images.

\section{Conclusions}
In this paper, we introduced ColorCode, a novel method for enhancing underwater images that provides three distinct capabilities: color enhancement, color adaptation, and color interpolation. These features are made possible through the introduction of a fundamental concept, color code, which is learned via supervised training and constrained to follow a Gaussian distribution. This allows for efficient sampling and interpolation during inference. Color enhancement is achieved through a supervised training process, ensuring accurate recovery of underwater colors. Color adaptation leverages the stability of long-wavelength hues to adjust colors under guidance, maintaining the natural appearance of underwater organisms. Finally, color interpolation enables the generation of diverse color variations by smoothly sampling the color code. Our extensive quantitative and qualitative evaluations on benchmark datasets confirm that ColorCode not only outperforms existing methods but also offers greater flexibility and realism in underwater image enhancement.

\bibliography{ref.bib}

\begin{thebibliography}{10}

\bibitem{bing2023domain}
X.~Bing, W.~Ren, Y.~Tang, G.~G. Yen, and Q.~Sun.
\newblock Domain adaptation for in-air to underwater image enhancement via deep
  learning.
\newblock {\em IEEE Transactions on Emerging Topics in Computational
  Intelligence}, 2023.

\bibitem{carlevaris2010initial}
N.~Carlevaris-Bianco, A.~Mohan, and R.~M. Eustice.
\newblock Initial results in underwater single image dehazing.
\newblock In {\em Oceans 2010 Mts/IEEE Seattle}, pages 1--8, 2010.

\bibitem{chen2021underwater}
X.~Chen, P.~Zhang, L.~Quan, C.~Yi, and C.~Lu.
\newblock Underwater image enhancement based on deep learning and image
  formation model.
\newblock {\em arXiv preprint arXiv:2101.00991}, 2021.

\bibitem{chen2022domain}
Y.-W. Chen and S.-C. Pei.
\newblock Domain adaptation for underwater image enhancement via content and
  style separation.
\newblock {\em IEEE Access}, 10:90523--90534, 2022.

\bibitem{cong2024underwater}
X.~Cong, J.~Gui, and J.~Hou.
\newblock Underwater organism color fine-tuning via decomposition and guidance.
\newblock In {\em AAAI Conference on Artificial Intelligence}, volume~38, pages
  1389--1398, 2024.

\bibitem{cong2024comprehensive}
X.~Cong, Y.~Zhao, J.~Gui, J.~Hou, and D.~Tao.
\newblock A comprehensive survey on underwater image enhancement based on deep
  learning.
\newblock {\em arXiv preprint arXiv:2405.19684}, 2024.

\bibitem{drews2016underwater}
P.~L. Drews, E.~R. Nascimento, S.~S. Botelho, and M.~F.~M. Campos.
\newblock Underwater depth estimation and image restoration based on single
  images.
\newblock {\em IEEE computer graphics and applications}, 36(2):24--35, 2016.

\bibitem{dziugaite2015training}
G.~K. Dziugaite, D.~M. Roy, and Z.~Ghahramani.
\newblock Training generative neural networks via maximum mean discrepancy
  optimization.
\newblock In {\em Proceedings of the Thirty-First Conference on Uncertainty in
  Artificial Intelligence}, pages 258--267, 2015.

\bibitem{esmaeilzehi2024dmml}
A.~Esmaeilzehi, Y.~Ou, M.~O. Ahmad, and M.~Swamy.
\newblock Dmml: Deep multi-prior and multi-discriminator learning for
  underwater image enhancement.
\newblock {\em IEEE Transactions on Broadcasting}, 2024.

\bibitem{fabbri2018enhancing}
C.~Fabbri, M.~J. Islam, and J.~Sattar.
\newblock Enhancing underwater imagery using generative adversarial networks.
\newblock In {\em International Conference on Robotics and Automation}, pages
  7159--7165, 2018.

\bibitem{fu2022unsupervised}
Z.~Fu, H.~Lin, Y.~Yang, S.~Chai, L.~Sun, Y.~Huang, and X.~Ding.
\newblock Unsupervised underwater image restoration: From a homology
  perspective.
\newblock In {\em AAAI Conference on Artificial Intelligence}, pages 643--651,
  2022.

\bibitem{fu2022underwater}
Z.~Fu, X.~Lin, W.~Wang, Y.~Huang, and X.~Ding.
\newblock Underwater image enhancement via learning water type desensitized
  representations.
\newblock In {\em IEEE International Conference on Acoustics, Speech and Signal
  Processing}, pages 2764--2768, 2022.

\bibitem{gonzalez2024dgd}
S.~Gonzalez-Sabbagh, A.~Robles-Kelly, and S.~Gao.
\newblock Dgd-cgan: A dual generator for image dewatering and restoration.
\newblock {\em Pattern Recognition}, 148:110159, 2024.

\bibitem{guo2023underwater}
C.~Guo, R.~Wu, X.~Jin, L.~Han, W.~Zhang, Z.~Chai, and C.~Li.
\newblock Underwater ranker: Learn which is better and how to be better.
\newblock In {\em AAAI Conference on Artificial Intelligence}, pages 702--709,
  2023.

\bibitem{hou2023non}
G.~Hou, N.~Li, P.~Zhuang, K.~Li, H.~Sun, and C.~Li.
\newblock Non-uniform illumination underwater image restoration via
  illumination channel sparsity prior.
\newblock {\em IEEE Transactions on Circuits and Systems for Video Technology},
  34(2):799--814, 2024.

\bibitem{huang2023contrastive}
S.~Huang, K.~Wang, H.~Liu, J.~Chen, and Y.~Li.
\newblock Contrastive semi-supervised learning for underwater image restoration
  via reliable bank.
\newblock In {\em IEEE Conference on Computer Vision and Pattern Recognition},
  pages 18145--18155, 2023.

\bibitem{huang2018multimodal}
X.~Huang, M.-Y. Liu, S.~Belongie, and J.~Kautz.
\newblock Multimodal unsupervised image-to-image translation.
\newblock In {\em Proceedings of the European Conference on Computer Vision},
  pages 172--189, 2018.

\bibitem{islam2020semantic}
M.~J. Islam, C.~Edge, Y.~Xiao, P.~Luo, M.~Mehtaz, C.~Morse, S.~S. Enan, and
  J.~Sattar.
\newblock Semantic segmentation of underwater imagery: Dataset and benchmark.
\newblock In {\em IEEE/RSJ International Conference on Intelligent Robots and
  Systems}, pages 1769--1776, 2020.

\bibitem{islam2020simultaneous}
M.~J. Islam, P.~Luo, and J.~Sattar.
\newblock Simultaneous enhancement and super-resolution of underwater imagery
  for improved visual perception.
\newblock In {\em 16th Robotics: Science and Systems, RSS 2020}, 2020.

\bibitem{islam2020fast}
M.~J. Islam, Y.~Xia, and J.~Sattar.
\newblock Fast underwater image enhancement for improved visual perception.
\newblock {\em IEEE Robotics and Automation Letters}, 5(2):3227--3234, 2020.

\bibitem{jiang2023perception}
Q.~Jiang, Y.~Kang, Z.~Wang, W.~Ren, and C.~Li.
\newblock Perception-driven deep underwater image enhancement without paired
  supervision.
\newblock {\em IEEE Transactions on Multimedia}, 2023.

\bibitem{jiang2022two}
Q.~Jiang, Y.~Zhang, F.~Bao, X.~Zhao, C.~Zhang, and P.~Liu.
\newblock Two-step domain adaptation for underwater image enhancement.
\newblock {\em Pattern Recognition}, 122:108324, 2022.

\bibitem{khan2024spectroformer}
R.~Khan, P.~Mishra, N.~Mehta, S.~S. Phutke, S.~K. Vipparthi, S.~Nandi, and
  S.~Murala.
\newblock Spectroformer: Multi-domain query cascaded transformer network for
  underwater image enhancement.
\newblock In {\em IEEE Winter Conference on Applications of Computer Vision},
  pages 1454--1463, 2024.

\bibitem{kim2021pixel}
G.~Kim, S.~W. Park, and J.~Kwon.
\newblock Pixel-wise wasserstein autoencoder for highly generative dehazing.
\newblock {\em IEEE Transactions on Image Processing}, 30:5452--5462, 2021.

\bibitem{li2021underwater}
C.~Li, S.~Anwar, J.~Hou, R.~Cong, C.~Guo, and W.~Ren.
\newblock Underwater image enhancement via medium transmission-guided
  multi-color space embedding.
\newblock {\em IEEE Transactions on Image Processing}, 30:4985--5000, 2021.

\bibitem{li2019underwater}
C.~Li, C.~Guo, W.~Ren, R.~Cong, J.~Hou, S.~Kwong, and D.~Tao.
\newblock An underwater image enhancement benchmark dataset and beyond.
\newblock {\em IEEE Transactions on Image Processing}, 29:4376--4389, 2019.

\bibitem{li2016single}
C.~Li, J.~Quo, Y.~Pang, S.~Chen, and J.~Wang.
\newblock Single underwater image restoration by blue-green channels dehazing
  and red channel correction.
\newblock In {\em IEEE International Conference on Acoustics, Speech and Signal
  Processing}, pages 1731--1735, 2016.

\bibitem{li2023tctl}
K.~Li, H.~Fan, Q.~Qi, C.~Yan, K.~Sun, and Q.~J. Wu.
\newblock Tctl-net: Template-free color transfer learning for self-attention
  driven underwater image enhancement.
\newblock {\em IEEE Transactions on Circuits and Systems for Video Technology},
  2023.

\bibitem{li2023ruiesr}
Y.~Li, L.~Shen, M.~Li, Z.~Wang, and L.~Zhuang.
\newblock Ruiesr: Realistic underwater image enhancement and super resolution.
\newblock {\em IEEE Transactions on Circuits and Systems for Video Technology},
  2023.

\bibitem{li2019joint}
Z.~Li, C.~Zhang, G.~Meng, and Y.~Liu.
\newblock Joint haze image synthesis and dehazing with mmd-vae losses.
\newblock {\em arXiv preprint arXiv:1905.05947}, 2019.

\bibitem{liu2019underwater}
P.~Liu, G.~Wang, H.~Qi, C.~Zhang, H.~Zheng, and Z.~Yu.
\newblock Underwater image enhancement with a deep residual framework.
\newblock {\em IEEE Access}, 7:94614--94629, 2019.

\bibitem{liu2023wsds}
Q.~Liu, Q.~Zhang, W.~Liu, W.~Chen, X.~Liu, and X.~Wang.
\newblock Wsds-gan: A weak-strong dual supervised learning method for
  underwater image enhancement.
\newblock {\em Pattern Recognition}, page 109774, 2023.

\bibitem{liu2022twin}
R.~Liu, Z.~Jiang, S.~Yang, and X.~Fan.
\newblock Twin adversarial contrastive learning for underwater image
  enhancement and beyond.
\newblock {\em IEEE Transactions on Image Processing}, 31:4922--4936, 2022.

\bibitem{lu2023speed}
S.~Lu, F.~Guan, H.~Zhang, and H.~Lai.
\newblock Speed-up ddpm for real-time underwater image enhancement.
\newblock {\em IEEE Transactions on Circuits and Systems for Video Technology},
  2023.

\bibitem{ma2022wavelet}
Z.~Ma and C.~Oh.
\newblock A wavelet-based dual-stream network for underwater image enhancement.
\newblock In {\em IEEE International Conference on Acoustics, Speech and Signal
  Processing}, pages 2769--2773, 2022.

\bibitem{mi2021revisiting}
L.~Mi, T.~He, C.~F. Park, H.~Wang, Y.~Wang, and N.~Shavit.
\newblock Revisiting latent-space interpolation via a quantitative evaluation
  framework.
\newblock {\em arXiv preprint arXiv:2110.06421}, 2021.

\bibitem{mu2023generalized}
P.~Mu, H.~Xu, Z.~Liu, Z.~Wang, S.~Chan, and C.~Bai.
\newblock A generalized physical-knowledge-guided dynamic model for underwater
  image enhancement.
\newblock In {\em ACM International Conference on Multimedia}, pages
  7111--7120, 2023.

\bibitem{naik2021shallow}
A.~Naik, A.~Swarnakar, and K.~Mittal.
\newblock Shallow-uwnet: Compressed model for underwater image enhancement
  (student abstract).
\newblock In {\em AAAI Conference on Artificial Intelligence}, pages
  15853--15854, 2021.

\bibitem{panetta2015human}
K.~Panetta, C.~Gao, and S.~Agaian.
\newblock Human-visual-system-inspired underwater image quality measures.
\newblock {\em IEEE Journal of Oceanic Engineering}, 41(3):541--551, 2015.

\bibitem{peng2023u}
L.~Peng, C.~Zhu, and L.~Bian.
\newblock U-shape transformer for underwater image enhancement.
\newblock {\em IEEE Transactions on Image Processing}, 2023.

\bibitem{peng2017underwater}
Y.-T. Peng and P.~C. Cosman.
\newblock Underwater image restoration based on image blurriness and light
  absorption.
\newblock {\em IEEE transactions on image processing}, 26(4):1579--1594, 2017.

\bibitem{qi2023deep}
H.~Qi, H.~Zhou, J.~Dong, and X.~Dong.
\newblock Deep color-corrected multi-scale retinex network for underwater image
  enhancement.
\newblock {\em IEEE Transactions on Geoscience and Remote Sensing}, 2023.

\bibitem{qi2022sguie}
Q.~Qi, K.~Li, H.~Zheng, X.~Gao, G.~Hou, and K.~Sun.
\newblock Sguie-net: Semantic attention guided underwater image enhancement
  with multi-scale perception.
\newblock {\em IEEE Transactions on Image Processing}, 31:6816--6830, 2022.

\bibitem{rao2023deep}
Y.~Rao, W.~Liu, K.~Li, H.~Fan, S.~Wang, and J.~Dong.
\newblock Deep color compensation for generalized underwater image enhancement.
\newblock {\em IEEE Transactions on Circuits and Systems for Video Technology},
  2023.

\bibitem{song2024hierarchical}
W.~Song, Z.~Shen, M.~Zhang, Y.~Wang, and A.~Liotta.
\newblock A hierarchical probabilistic underwater image enhancement model with
  reinforcement tuning.
\newblock {\em Journal of Visual Communication and Image Representation}, page
  104052, 2024.

\bibitem{song2020enhancement}
W.~Song, Y.~Wang, D.~Huang, A.~Liotta, and C.~Perra.
\newblock Enhancement of underwater images with statistical model of background
  light and optimization of transmission map.
\newblock {\em IEEE Transactions on Broadcasting}, 66(1):153--169, 2020.

\bibitem{song2018rapid}
W.~Song, Y.~Wang, D.~Huang, and D.~Tjondronegoro.
\newblock A rapid scene depth estimation model based on underwater light
  attenuation prior for underwater image restoration.
\newblock In {\em Advances in Multimedia Information Processing Pacific-Rim
  Conference on Multimedia}, pages 678--688, 2018.

\bibitem{tang2022autoenhancer}
Y.~Tang, T.~Iwaguchi, H.~Kawasaki, R.~Sagawa, and R.~Furukawa.
\newblock Autoenhancer: Transformer on u-net architecture search for underwater
  image enhancement.
\newblock In {\em Proceedings of the Asian Conference on Computer Vision},
  pages 1403--1420, 2022.

\bibitem{tang2023underwater}
Y.~Tang, H.~Kawasaki, and T.~Iwaguchi.
\newblock Underwater image enhancement by transformer-based diffusion model
  with non-uniform sampling for skip strategy.
\newblock In {\em ACM International Conference on Multimedia}, pages
  5419--5427, 2023.

\bibitem{tolstikhin2017wasserstein}
I.~Tolstikhin, O.~Bousquet, S.~Gelly, and B.~Schoelkopf.
\newblock Wasserstein auto-encoders.
\newblock {\em arXiv preprint arXiv:1711.01558}, 2017.

\bibitem{uplavikar2019all}
P.~M. Uplavikar, Z.~Wu, and Z.~Wang.
\newblock All-in-one underwater image enhancement using domain-adversarial
  learning.
\newblock In {\em IEEE Conference on Computer Vision and Pattern Recognition
  Workshops}, pages 1--8, 2019.

\bibitem{wang2022semantic}
D.~Wang, L.~Ma, R.~Liu, and X.~Fan.
\newblock Semantic-aware texture-structure feature collaboration for underwater
  image enhancement.
\newblock In {\em International Conference on Robotics and Automation}, pages
  4592--4598, 2022.

\bibitem{wang2024metalantis}
H.~Wang, W.~Zhang, L.~Bai, and P.~Ren.
\newblock Metalantis: A comprehensive underwater image enhancement framework.
\newblock {\em IEEE Transactions on Geoscience and Remote Sensing}, 2024.

\bibitem{wang2021leveraging}
Y.~Wang, Y.~Cao, J.~Zhang, F.~Wu, and Z.-J. Zha.
\newblock Leveraging deep statistics for underwater image enhancement.
\newblock {\em ACM Transactions on Multimedia Computing, Communications, and
  Applications}, 17(3s):1--20, 2021.

\bibitem{wang2024multi}
Y.~Wang, S.~Hu, S.~Yin, Z.~Deng, and Y.-H. Yang.
\newblock A multi-level wavelet-based underwater image enhancement network with
  color compensation prior.
\newblock {\em Expert Systems with Applications}, 242:122710, 2024.

\bibitem{wang2017deep}
Y.~Wang, J.~Zhang, Y.~Cao, and Z.~Wang.
\newblock A deep cnn method for underwater image enhancement.
\newblock In {\em 2017 IEEE international conference on image processing},
  pages 1382--1386, 2017.

\bibitem{wang2004image}
Z.~Wang, A.~C. Bovik, H.~R. Sheikh, and E.~P. Simoncelli.
\newblock Image quality assessment: from error visibility to structural
  similarity.
\newblock {\em IEEE transactions on image processing}, 13(4):600--612, 2004.

\bibitem{wang2023domain}
Z.~Wang, L.~Shen, M.~Xu, M.~Yu, K.~Wang, and Y.~Lin.
\newblock Domain adaptation for underwater image enhancement.
\newblock {\em IEEE Transactions on Image Processing}, 32:1442--1457, 2023.

\bibitem{wei2022uhd}
Y.~Wei, Z.~Zheng, and X.~Jia.
\newblock Uhd underwater image enhancement via frequency-spatial domain aware
  network.
\newblock In {\em Proceedings of the Asian Conference on Computer Vision},
  pages 299--314, 2022.

\bibitem{wu2024self}
Z.~Wu, Z.~Wu, X.~Chen, Y.~Lu, and J.~Yu.
\newblock Self-supervised underwater image generation for underwater domain
  pre-training.
\newblock {\em IEEE Transactions on Instrumentation and Measurement}, 2024.

\bibitem{xie2024uveb}
Y.~Xie, L.~Kong, K.~Chen, Z.~Zheng, X.~Yu, Z.~Yu, and B.~Zheng.
\newblock Uveb: A large-scale benchmark and baseline towards real-world
  underwater video enhancement.
\newblock In {\em IEEE Conference on Computer Vision and Pattern Recognition},
  2024.

\bibitem{xue2023investigating}
X.~Xue, Z.~Li, L.~Ma, Q.~Jia, R.~Liu, and X.~Fan.
\newblock Investigating intrinsic degradation factors by multi-branch
  aggregation for real-world underwater image enhancement.
\newblock {\em Pattern Recognition}, 133:109041, 2023.

\bibitem{yan2023uw}
H.~Yan, Z.~Zhang, J.~Xu, T.~Wang, P.~An, A.~Wang, and Y.~Duan.
\newblock Uw-cyclegan: Model-driven cyclegan for underwater image restoration.
\newblock {\em IEEE Transactions on Geoscience and Remote Sensing}, 2023.

\bibitem{yan2022attention}
X.~Yan, W.~Qin, Y.~Wang, G.~Wang, and X.~Fu.
\newblock Attention-guided dynamic multi-branch neural network for underwater
  image enhancement.
\newblock {\em Knowledge-Based Systems}, 258:110041, 2022.

\bibitem{yang2023joint}
G.~Yang, G.~Kang, J.~Lee, and Y.~Cho.
\newblock Joint-id: Transformer-based joint image enhancement and depth
  estimation for underwater environments.
\newblock {\em IEEE Sensors Journal}, 2023.

\bibitem{yin2024unsupervised}
J.~Yin, Y.~Wang, B.~Guan, X.~Zeng, and L.~Guo.
\newblock Unsupervised underwater image enhancement based on disentangled
  representations via double-order contrastive loss.
\newblock {\em IEEE Transactions on Geoscience and Remote Sensing}, 2024.

\bibitem{yu2023task}
M.~Yu, L.~Shen, Z.~Wang, and X.~Hua.
\newblock Task-friendly underwater image enhancement for machine vision
  applications.
\newblock {\em IEEE Transactions on Geoscience and Remote Sensing}, 2023.

\bibitem{zhang2024robust}
D.~Zhang, C.~Wu, J.~Zhou, W.~Zhang, Z.~Lin, K.~Polat, and F.~Alenezi.
\newblock Robust underwater image enhancement with cascaded multi-level
  sub-networks and triple attention mechanism.
\newblock {\em Neural Networks}, 169:685--697, 2024.

\bibitem{zhang2024atlantis}
F.~Zhang, S.~You, Y.~Li, and Y.~Fu.
\newblock Atlantis: Enabling underwater depth estimation with stable diffusion.
\newblock In {\em IEEE Conference on Computer Vision and Pattern Recognition},
  2024.

\bibitem{zhang2019survey}
W.~Zhang, L.~Dong, X.~Pan, P.~Zou, L.~Qin, and W.~Xu.
\newblock A survey of restoration and enhancement for underwater images.
\newblock {\em IEEE Access}, 7:182259--182279, 2019.

\bibitem{zhang2022underwater}
W.~Zhang, P.~Zhuang, H.-H. Sun, G.~Li, S.~Kwong, and C.~Li.
\newblock Underwater image enhancement via minimal color loss and locally
  adaptive contrast enhancement.
\newblock {\em IEEE Transactions on Image Processing}, 31:3997--4010, 2022.

\bibitem{zhang2023waterflow}
Z.~Zhang, Z.~Jiang, J.~Liu, X.~Fan, and R.~Liu.
\newblock Waterflow: heuristic normalizing flow for underwater image
  enhancement and beyond.
\newblock In {\em ACM International Conference on Multimedia}, pages
  7314--7323, 2023.

\bibitem{zhao2023wavelet}
C.~Zhao, W.~Cai, C.~Dong, and C.~Hu.
\newblock Wavelet-based fourier information interaction with frequency
  diffusion adjustment for underwater image restoration.
\newblock {\em arXiv preprint arXiv:2311.16845}, 2023.

\bibitem{zhao2016loss}
H.~Zhao, O.~Gallo, and I.~Frosio.
\newblock Loss functions for image restoration with neural networks.
\newblock {\em IEEE Transactions on computational imaging}, 3(1):47--57, 2016.

\bibitem{zhou2024iacc}
J.~Zhou, Q.~Gai, D.~Zhang, K.-M. Lam, W.~Zhang, and X.~Fu.
\newblock Iacc: Cross-illumination awareness and color correction for
  underwater images under mixed natural and artificial lighting.
\newblock {\em IEEE Transactions on Geoscience and Remote Sensing}, 62:1--15,
  2024.

\bibitem{zhou2023ugif}
J.~Zhou, B.~Li, D.~Zhang, J.~Yuan, W.~Zhang, Z.~Cai, and J.~Shi.
\newblock Ugif-net: An efficient fully guided information flow network for
  underwater image enhancement.
\newblock {\em IEEE Transactions on Geoscience and Remote Sensing}, 2023.

\bibitem{zhou2024hclr}
J.~Zhou, J.~Sun, C.~Li, Q.~Jiang, M.~Zhou, K.-M. Lam, W.~Zhang, and X.~Fu.
\newblock Hclr-net: Hybrid contrastive learning regularization with locally
  randomized perturbation for underwater image enhancement.
\newblock {\em International Journal of Computer Vision}, pages 1--25, 2024.

\bibitem{zhuang2022underwater}
P.~Zhuang, J.~Wu, F.~Porikli, and C.~Li.
\newblock Underwater image enhancement with hyper-laplacian reflectance priors.
\newblock {\em IEEE Transactions on Image Processing}, 31:5442--5455, 2022.

\end{thebibliography}

\clearpage

\appendix
\section{More Visualizations}

\subsection{More Color Enhancement Results}
\begin{figure*}[!b]
    \centering
    \includegraphics[width=1.9cm,height=1.6cm]{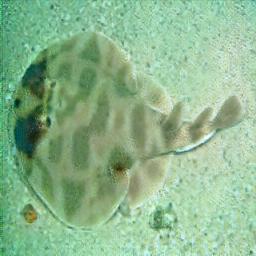}
    \includegraphics[width=1.9cm,height=1.6cm]{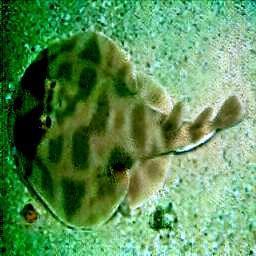}
    \includegraphics[width=1.9cm,height=1.6cm]{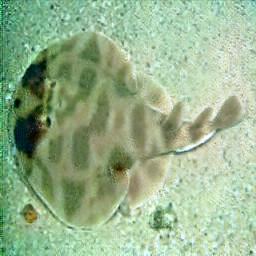}
    \includegraphics[width=1.9cm,height=1.6cm]{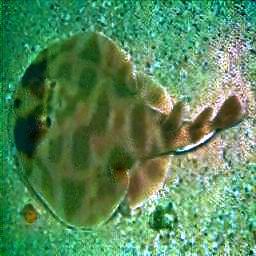}
    \includegraphics[width=1.9cm,height=1.6cm]{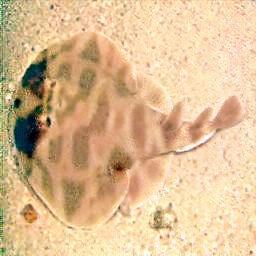}
    \includegraphics[width=1.9cm,height=1.6cm]{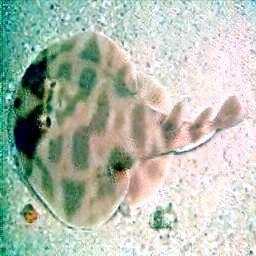}
    \includegraphics[width=1.9cm,height=1.6cm]{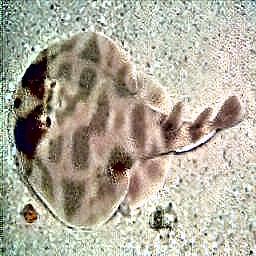}
    \includegraphics[width=1.9cm,height=1.6cm]{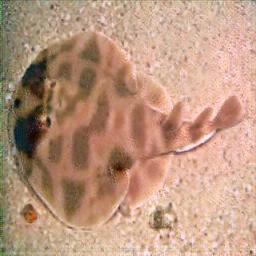}
    \includegraphics[width=1.9cm,height=1.6cm]{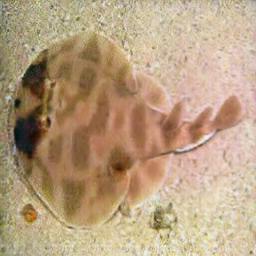}

    \leftline{\hspace{0.1cm} Distortion \hspace{0.7cm} IBLA \hspace{0.9cm} UDCP \hspace{0.9cm}  MIP \hspace{1cm} ULAP \hspace{0.3cm} SMBLOTMOP \hspace{0.3cm} MLLE \hspace{0.45cm} PhysicalNN \hspace{0.05cm} FUnIEGAN}

    \vspace{0.1cm}

    \includegraphics[width=1.9cm,height=1.6cm]{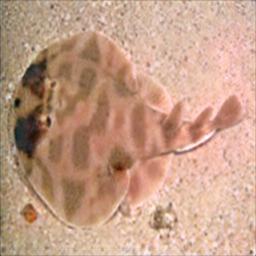}
    \includegraphics[width=1.9cm,height=1.6cm]{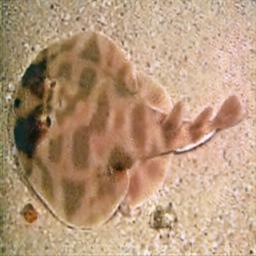}
    \includegraphics[width=1.9cm,height=1.6cm]{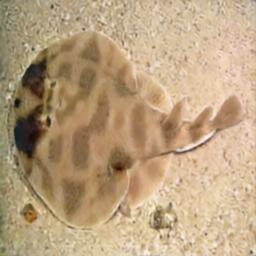}
    \includegraphics[width=1.9cm,height=1.6cm]{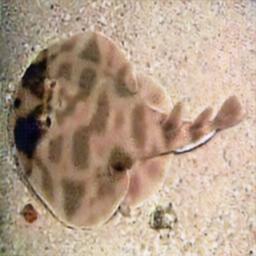}
    \includegraphics[width=1.9cm,height=1.6cm]{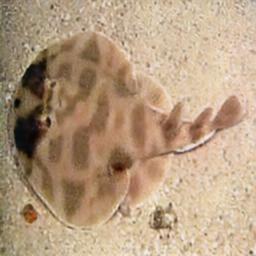}
    \includegraphics[width=1.9cm,height=1.6cm]{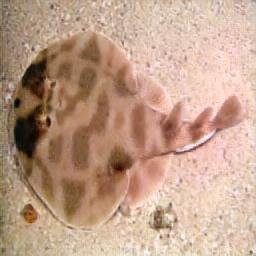}
    \includegraphics[width=1.9cm,height=1.6cm]{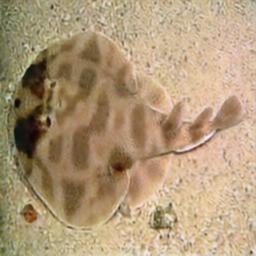}
    \includegraphics[width=1.9cm,height=1.6cm]{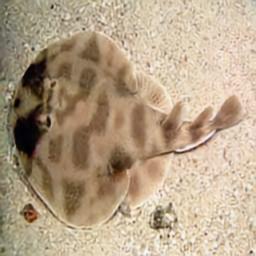}
    \includegraphics[width=1.9cm,height=1.6cm]{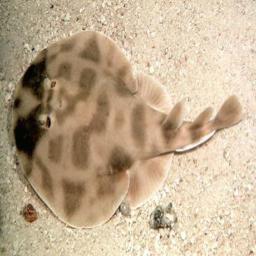}

    \leftline{\hspace{0.2cm} WaterNet \hspace{0.3cm} ADMNNet \hspace{0.5cm} UColor \hspace{0.7cm} UGAN \hspace{0.7cm} U-Trans \hspace{0.5cm} UIE-WD \hspace{0.6cm} SGUIE \hspace{0.6cm} ColorCode \hspace{0.3cm} Reference}

    \rule{17.1cm}{0.4pt}
    \vspace{0.2cm}

    \includegraphics[width=1.9cm,height=1.6cm]{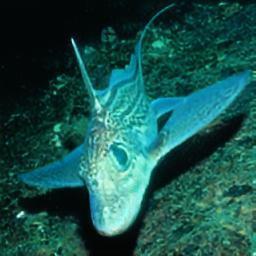}
    \includegraphics[width=1.9cm,height=1.6cm]{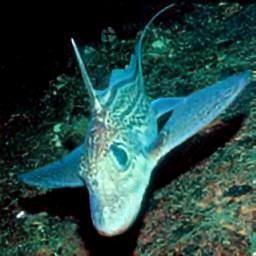}
    \includegraphics[width=1.9cm,height=1.6cm]{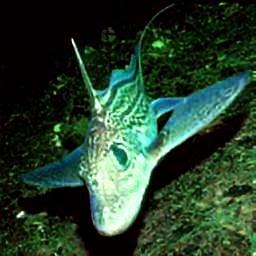}
    \includegraphics[width=1.9cm,height=1.6cm]{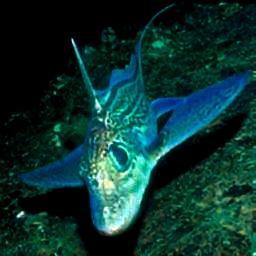}
    \includegraphics[width=1.9cm,height=1.6cm]{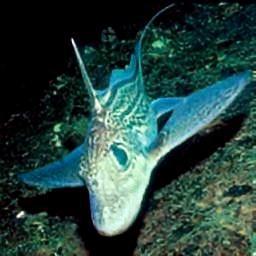}
    \includegraphics[width=1.9cm,height=1.6cm]{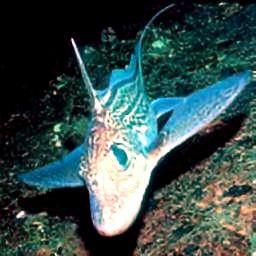}
    \includegraphics[width=1.9cm,height=1.6cm]{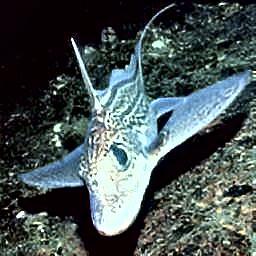}
    \includegraphics[width=1.9cm,height=1.6cm]{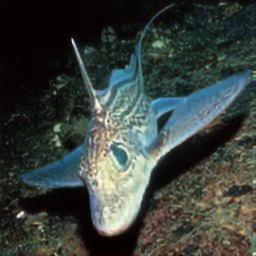}
    \includegraphics[width=1.9cm,height=1.6cm]{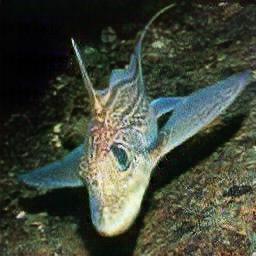}

    \leftline{\hspace{0.1cm} Distortion \hspace{0.7cm} IBLA \hspace{0.9cm} UDCP \hspace{0.9cm}  MIP \hspace{1cm} ULAP \hspace{0.3cm} SMBLOTMOP \hspace{0.3cm} MLLE \hspace{0.45cm} PhysicalNN \hspace{0.05cm} FUnIEGAN}

    \vspace{0.1cm}

    \includegraphics[width=1.9cm,height=1.6cm]{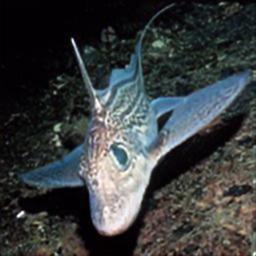}
    \includegraphics[width=1.9cm,height=1.6cm]{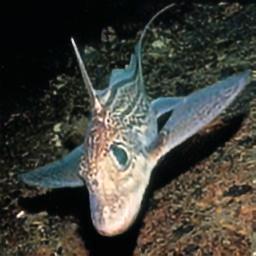}
    \includegraphics[width=1.9cm,height=1.6cm]{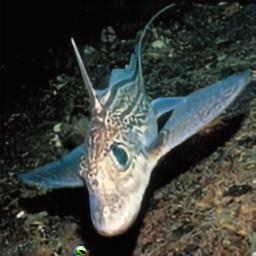}
    \includegraphics[width=1.9cm,height=1.6cm]{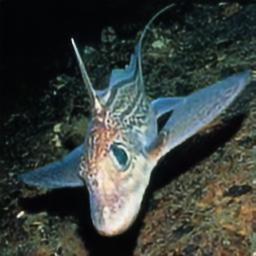}
    \includegraphics[width=1.9cm,height=1.6cm]{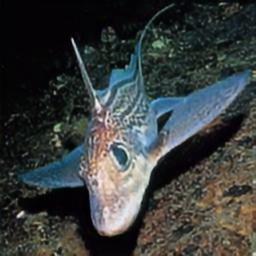}
    \includegraphics[width=1.9cm,height=1.6cm]{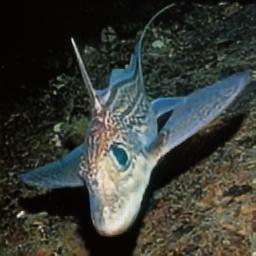}
    \includegraphics[width=1.9cm,height=1.6cm]{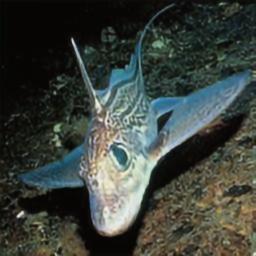}
    \includegraphics[width=1.9cm,height=1.6cm]{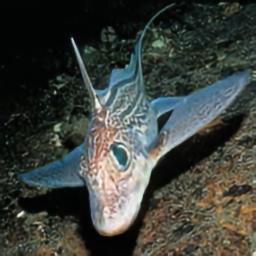}
    \includegraphics[width=1.9cm,height=1.6cm]{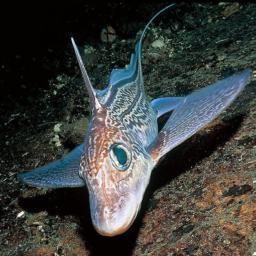}

    \leftline{\hspace{0.2cm} WaterNet \hspace{0.3cm} ADMNNet \hspace{0.5cm} UColor \hspace{0.7cm} UGAN \hspace{0.7cm} U-Trans \hspace{0.5cm} UIE-WD \hspace{0.6cm} SGUIE \hspace{0.6cm} ColorCode \hspace{0.3cm} Reference}

    \caption{Visual results obtained by various UIE algorithms on EUVP-I dataset.}
    \label{fig:visual_results_EUVP_I}
  \end{figure*}

\begin{figure*}
    \centering
    \includegraphics[width=1.9cm,height=1.6cm]{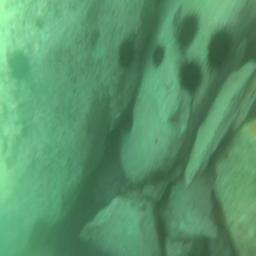}
    \includegraphics[width=1.9cm,height=1.6cm]{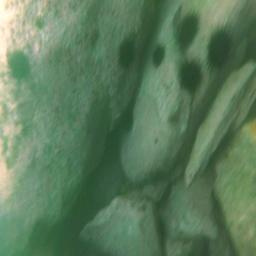}
    \includegraphics[width=1.9cm,height=1.6cm]{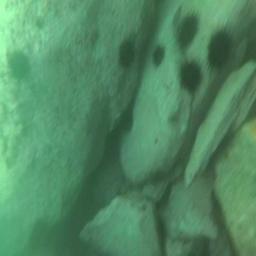}
    \includegraphics[width=1.9cm,height=1.6cm]{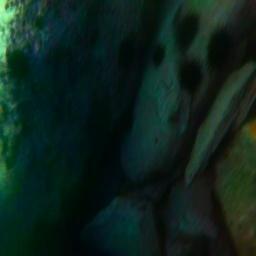}
    \includegraphics[width=1.9cm,height=1.6cm]{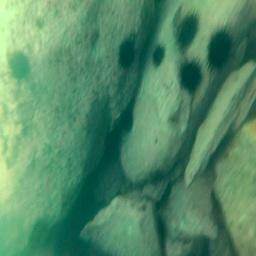}
    \includegraphics[width=1.9cm,height=1.6cm]{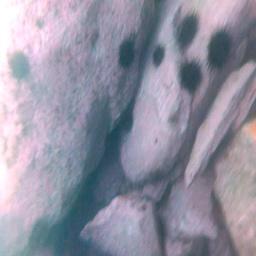}
    \includegraphics[width=1.9cm,height=1.6cm]{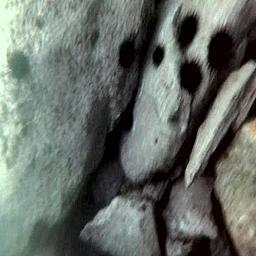}
    \includegraphics[width=1.9cm,height=1.6cm]{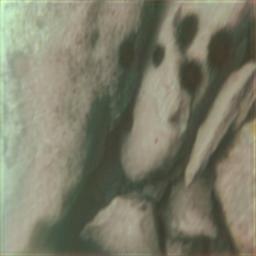}
    \includegraphics[width=1.9cm,height=1.6cm]{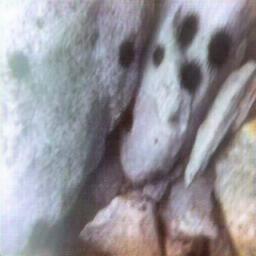}

    \leftline{\hspace{0.1cm} Distortion \hspace{0.7cm} IBLA \hspace{0.9cm} UDCP \hspace{0.9cm}  MIP \hspace{1cm} ULAP \hspace{0.3cm} SMBLOTMOP \hspace{0.3cm} MLLE \hspace{0.45cm} PhysicalNN \hspace{0.05cm} FUnIEGAN}

    \vspace{0.1cm}

    \includegraphics[width=1.9cm,height=1.6cm]{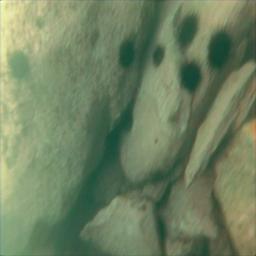}
    \includegraphics[width=1.9cm,height=1.6cm]{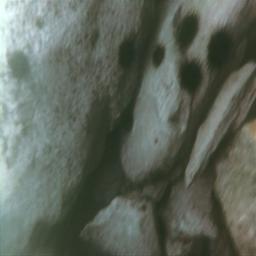}
    \includegraphics[width=1.9cm,height=1.6cm]{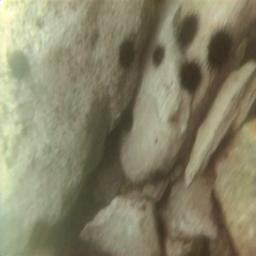}
    \includegraphics[width=1.9cm,height=1.6cm]{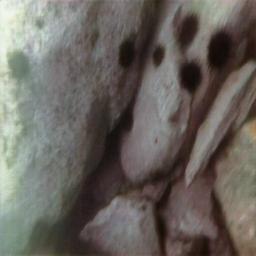}
    \includegraphics[width=1.9cm,height=1.6cm]{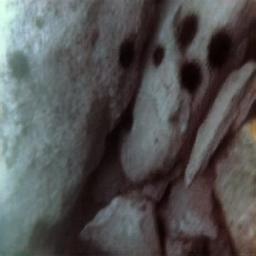}
    \includegraphics[width=1.9cm,height=1.6cm]{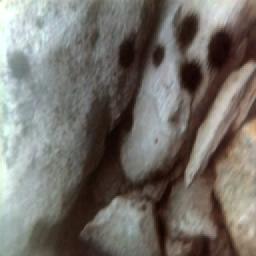}
    \includegraphics[width=1.9cm,height=1.6cm]{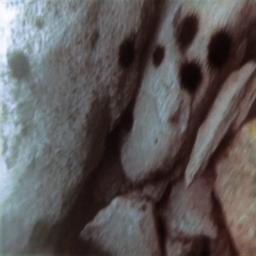}
    \includegraphics[width=1.9cm,height=1.6cm]{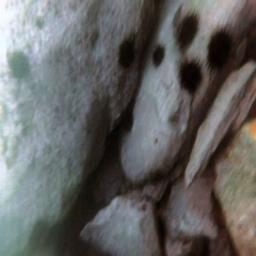}
    \includegraphics[width=1.9cm,height=1.6cm]{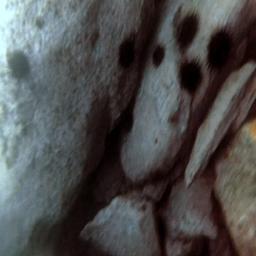}

    \leftline{\hspace{0.2cm} WaterNet \hspace{0.3cm} ADMNNet \hspace{0.5cm} UColor \hspace{0.7cm} UGAN \hspace{0.7cm} U-Trans \hspace{0.5cm} UIE-WD \hspace{0.6cm} SGUIE \hspace{0.6cm} ColorCode \hspace{0.3cm} Reference}

    \rule{17.1cm}{0.4pt}
    \vspace{0.2cm}

    \includegraphics[width=1.9cm,height=1.6cm]{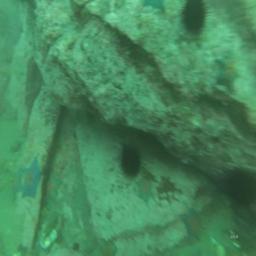}
    \includegraphics[width=1.9cm,height=1.6cm]{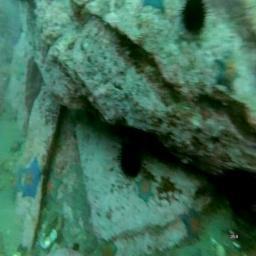}
    \includegraphics[width=1.9cm,height=1.6cm]{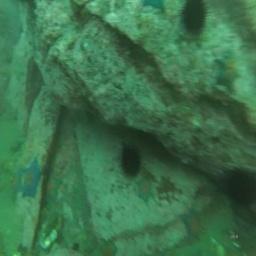}
    \includegraphics[width=1.9cm,height=1.6cm]{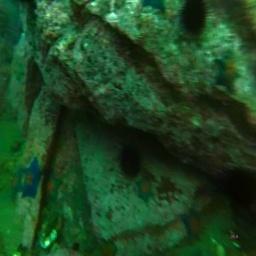}
    \includegraphics[width=1.9cm,height=1.6cm]{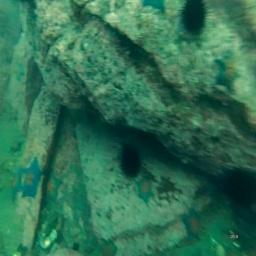}
    \includegraphics[width=1.9cm,height=1.6cm]{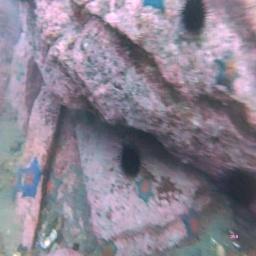}
    \includegraphics[width=1.9cm,height=1.6cm]{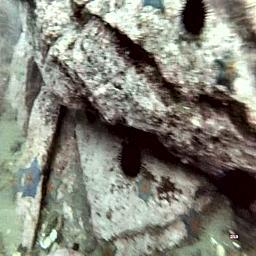}
    \includegraphics[width=1.9cm,height=1.6cm]{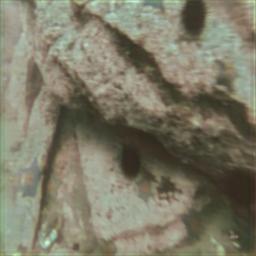}
    \includegraphics[width=1.9cm,height=1.6cm]{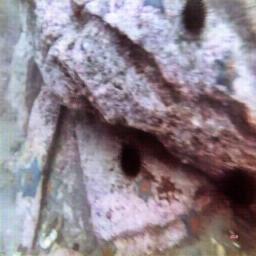}

    \leftline{\hspace{0.1cm} Distortion \hspace{0.7cm} IBLA \hspace{0.9cm} UDCP \hspace{0.9cm}  MIP \hspace{1cm} ULAP \hspace{0.3cm} SMBLOTMOP \hspace{0.3cm} MLLE \hspace{0.45cm} PhysicalNN \hspace{0.05cm} FUnIEGAN}

    \vspace{0.1cm}

    \includegraphics[width=1.9cm,height=1.6cm]{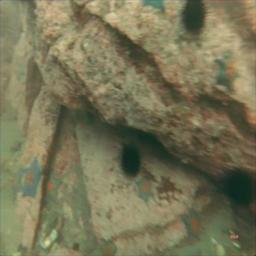}
    \includegraphics[width=1.9cm,height=1.6cm]{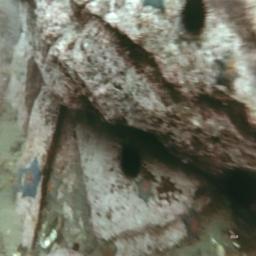}
    \includegraphics[width=1.9cm,height=1.6cm]{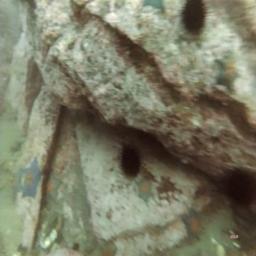}
    \includegraphics[width=1.9cm,height=1.6cm]{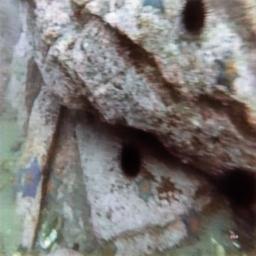}
    \includegraphics[width=1.9cm,height=1.6cm]{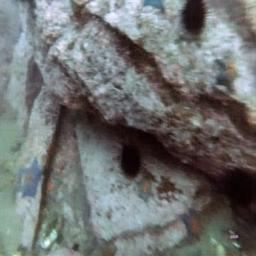}
    \includegraphics[width=1.9cm,height=1.6cm]{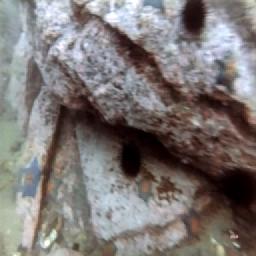}
    \includegraphics[width=1.9cm,height=1.6cm]{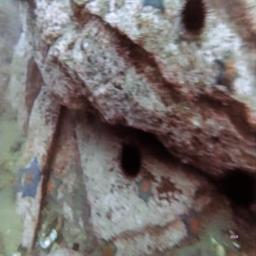}
    \includegraphics[width=1.9cm,height=1.6cm]{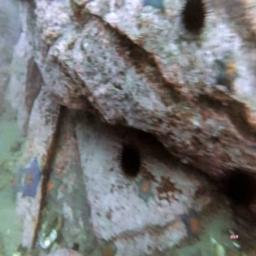}
    \includegraphics[width=1.9cm,height=1.6cm]{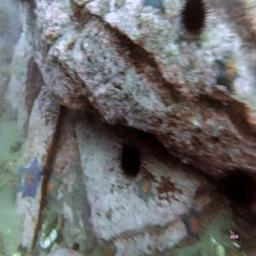}

    \leftline{\hspace{0.2cm} WaterNet \hspace{0.3cm} ADMNNet \hspace{0.5cm} UColor \hspace{0.7cm} UGAN \hspace{0.7cm} U-Trans \hspace{0.5cm} UIE-WD \hspace{0.6cm} SGUIE \hspace{0.6cm} ColorCode \hspace{0.3cm} Reference}

    \caption{Visual results obtained by various UIE algorithms on LSUI dataset.}
    \label{fig:visual_results_LSUI}
  \end{figure*}

\begin{figure*}
    \centering
    \includegraphics[width=1.9cm,height=1.6cm]{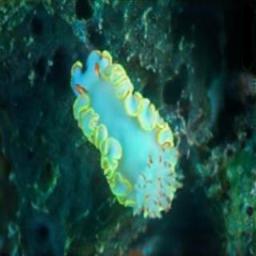}
    \includegraphics[width=1.9cm,height=1.6cm]{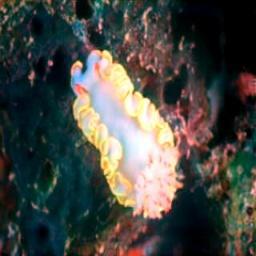}
    \includegraphics[width=1.9cm,height=1.6cm]{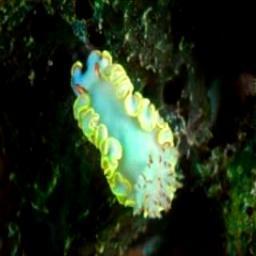}
    \includegraphics[width=1.9cm,height=1.6cm]{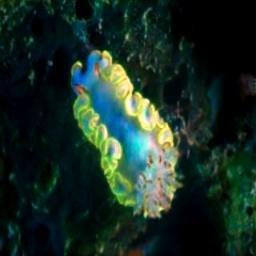}
    \includegraphics[width=1.9cm,height=1.6cm]{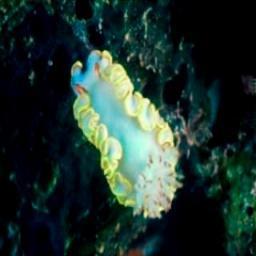}
    \includegraphics[width=1.9cm,height=1.6cm]{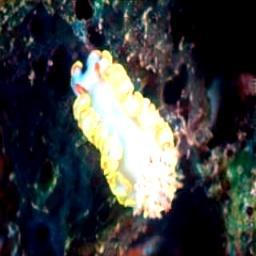}
    \includegraphics[width=1.9cm,height=1.6cm]{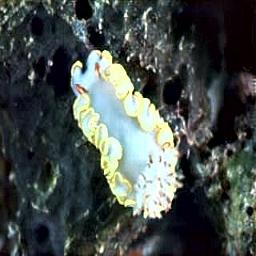}
    \includegraphics[width=1.9cm,height=1.6cm]{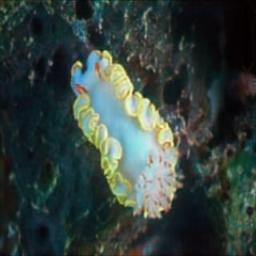}
    \includegraphics[width=1.9cm,height=1.6cm]{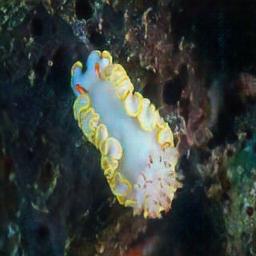}

    \leftline{\hspace{0.1cm} Distortion \hspace{0.7cm} IBLA \hspace{0.9cm} UDCP \hspace{0.9cm}  MIP \hspace{1cm} ULAP \hspace{0.3cm} SMBLOTMOP \hspace{0.3cm} MLLE \hspace{0.45cm} PhysicalNN \hspace{0.05cm} FUnIEGAN}

    \vspace{0.1cm}

    \includegraphics[width=1.9cm,height=1.6cm]{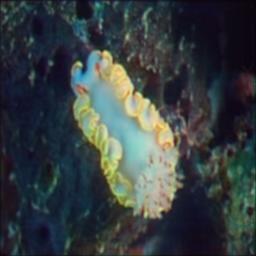}
    \includegraphics[width=1.9cm,height=1.6cm]{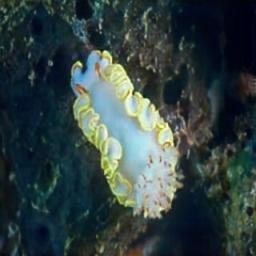}
    \includegraphics[width=1.9cm,height=1.6cm]{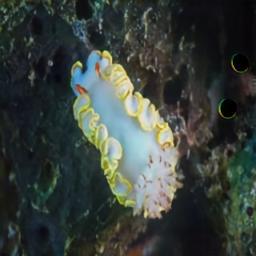}
    \includegraphics[width=1.9cm,height=1.6cm]{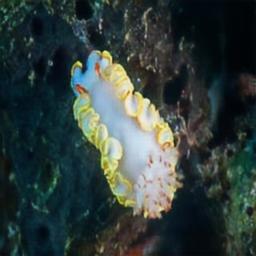}
    \includegraphics[width=1.9cm,height=1.6cm]{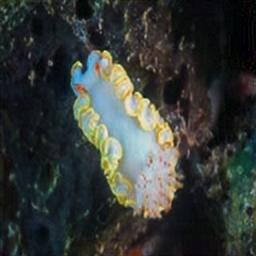}
    \includegraphics[width=1.9cm,height=1.6cm]{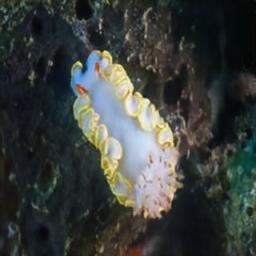}
    \includegraphics[width=1.9cm,height=1.6cm]{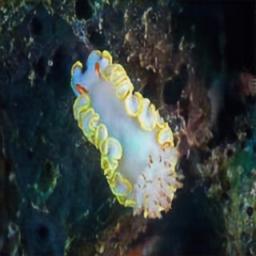}
    \includegraphics[width=1.9cm,height=1.6cm]{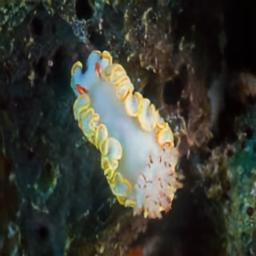}
    \includegraphics[width=1.9cm,height=1.6cm]{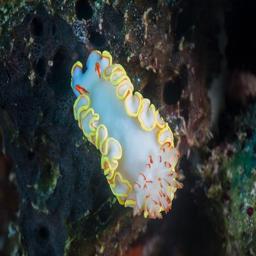}

    \leftline{\hspace{0.2cm} WaterNet \hspace{0.3cm} ADMNNet \hspace{0.5cm} UColor \hspace{0.7cm} UGAN \hspace{0.7cm} U-Trans \hspace{0.5cm} UIE-WD \hspace{0.6cm} SGUIE \hspace{0.6cm} ColorCode \hspace{0.3cm} Reference}

    \rule{17.1cm}{0.4pt}
    \vspace{0.2cm}

    \includegraphics[width=1.9cm,height=1.6cm]{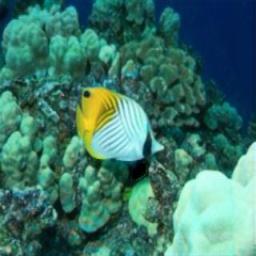}
    \includegraphics[width=1.9cm,height=1.6cm]{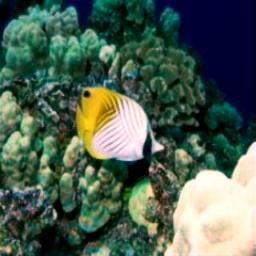}
    \includegraphics[width=1.9cm,height=1.6cm]{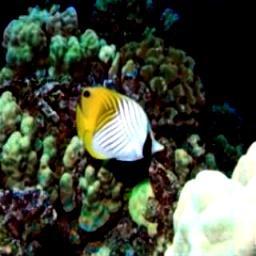}
    \includegraphics[width=1.9cm,height=1.6cm]{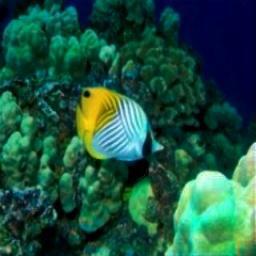}
    \includegraphics[width=1.9cm,height=1.6cm]{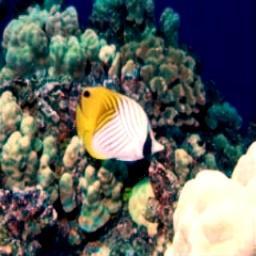}
    \includegraphics[width=1.9cm,height=1.6cm]{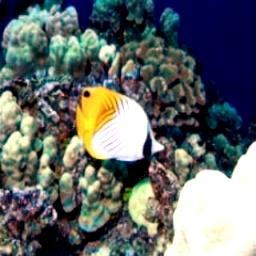}
    \includegraphics[width=1.9cm,height=1.6cm]{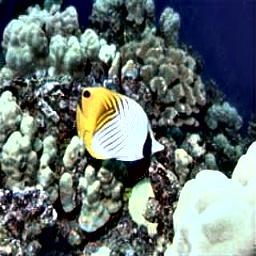}
    \includegraphics[width=1.9cm,height=1.6cm]{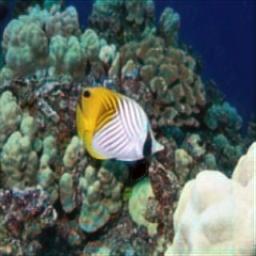}
    \includegraphics[width=1.9cm,height=1.6cm]{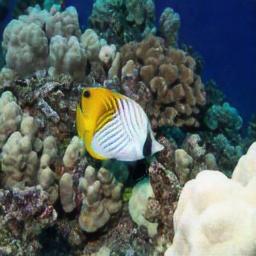}

    \leftline{\hspace{0.1cm} Distortion \hspace{0.7cm} IBLA \hspace{0.9cm} UDCP \hspace{0.9cm}  MIP \hspace{1cm} ULAP \hspace{0.3cm} SMBLOTMOP \hspace{0.3cm} MLLE \hspace{0.45cm} PhysicalNN \hspace{0.05cm} FUnIEGAN}

    \vspace{0.1cm}

    \includegraphics[width=1.9cm,height=1.6cm]{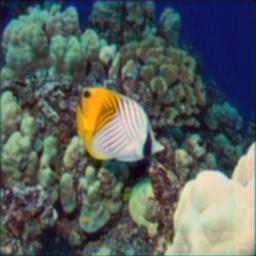}
    \includegraphics[width=1.9cm,height=1.6cm]{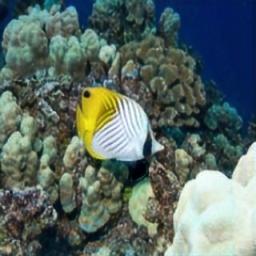}
    \includegraphics[width=1.9cm,height=1.6cm]{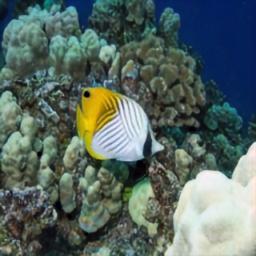}
    \includegraphics[width=1.9cm,height=1.6cm]{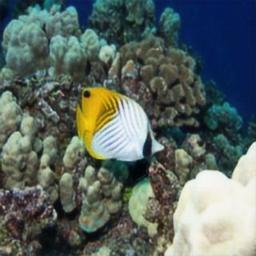}
    \includegraphics[width=1.9cm,height=1.6cm]{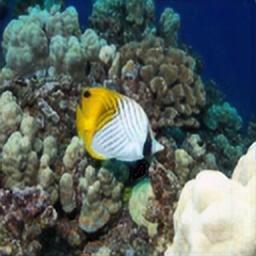}
    \includegraphics[width=1.9cm,height=1.6cm]{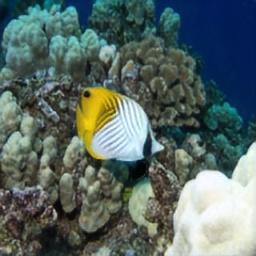}
    \includegraphics[width=1.9cm,height=1.6cm]{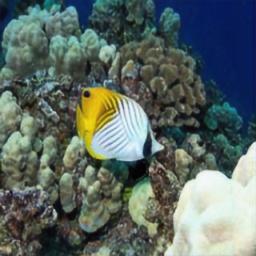}
    \includegraphics[width=1.9cm,height=1.6cm]{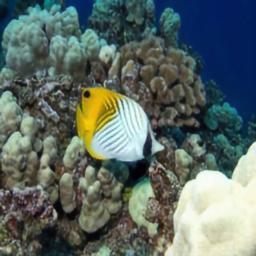}
    \includegraphics[width=1.9cm,height=1.6cm]{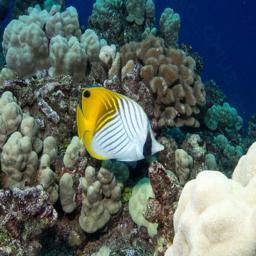}

    \leftline{\hspace{0.2cm} WaterNet \hspace{0.3cm} ADMNNet \hspace{0.5cm} UColor \hspace{0.7cm} UGAN \hspace{0.7cm} U-Trans \hspace{0.5cm} UIE-WD \hspace{0.6cm} SGUIE \hspace{0.6cm} ColorCode \hspace{0.3cm} Reference}

    \caption{Visual results obtained by various UIE algorithms on UFO120 dataset.}
    \label{fig:visual_results_UFO120}
  \end{figure*}

Fig. \ref{fig:visual_results_EUVP_I}, Fig. \ref{fig:visual_results_LSUI} and Fig. \ref{fig:visual_results_UFO120} show the enhanced images on EUVP (image-net) \cite{islam2020fast}, LSUI \cite{peng2023u} and UFO-120 \cite{islam2020simultaneous} datasets. The visual results demonstrate that the proposed ColorCode can achieve competitive performance. The details, textures and colors are effectively enhanced and corrected. The color enhancement capability of ColorCode is further verified.

\subsection{More Color Fine-tuning Results}
\begin{figure*}[!t]
    \centering
    \footnotesize
    \textcolor{black}{\leftline{\hspace{0.4cm} $x \rightarrow \hat{y}_{x}^{m}$ \hspace{0.9cm}  $\alpha = 0.1$ \hspace{0.5cm}  $\alpha = 0.2$  \hspace{0.5cm}   $\alpha = 0.25$ \hspace{0.5cm}  $\alpha = 0.3$  \hspace{0.5cm}  $\alpha = 0.35$  \hspace{0.45cm} $\alpha = 0.4$  \hspace{0.5cm}   $\alpha = 0.45$  \hspace{0.45cm}  $\alpha = 0.5$ \hspace{1.3cm}  $g$}}
    \\

    \vspace{0.02cm}

    \includegraphics[width=1.65cm,height=1.4cm]{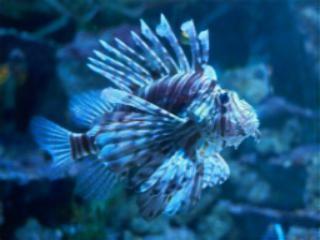}
    \includegraphics[width=0.2cm,height=1.4cm]{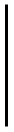}
    \includegraphics[width=1.65cm,height=1.4cm]{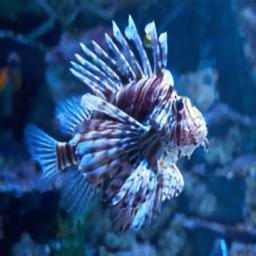}
    \includegraphics[width=1.65cm,height=1.4cm]{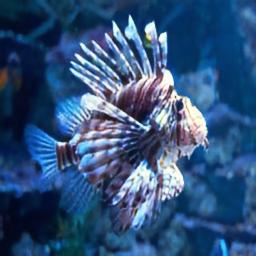}
    \includegraphics[width=1.65cm,height=1.4cm]{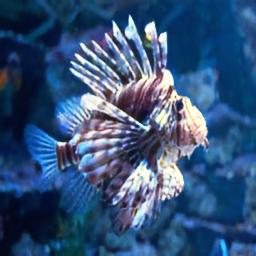}
    \includegraphics[width=1.65cm,height=1.4cm]{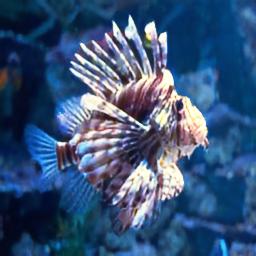}
    \includegraphics[width=1.65cm,height=1.4cm]{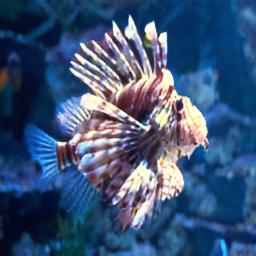}
    \includegraphics[width=1.65cm,height=1.4cm]{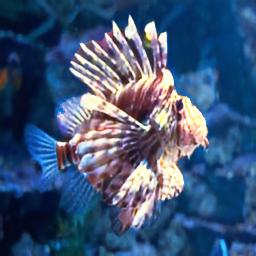}
    \includegraphics[width=1.65cm,height=1.4cm]{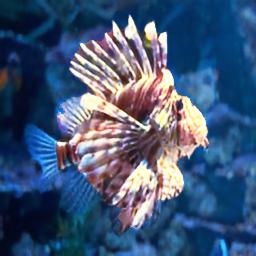}
    \includegraphics[width=1.65cm,height=1.4cm]{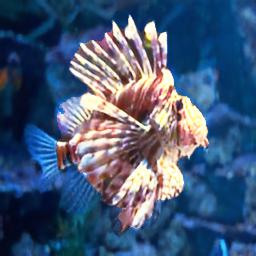}
    \includegraphics[width=0.2cm,height=1.4cm]{00_figures_supp/materials/black_vertical_line.PNG}
    \includegraphics[width=1.65cm,height=1.4cm]{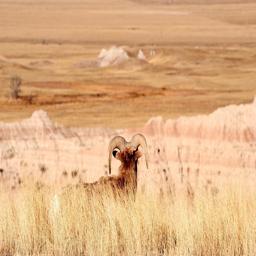}

    \vspace{0.1cm}

    \includegraphics[width=1.65cm,height=1.4cm]{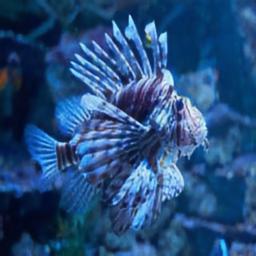}
    \includegraphics[width=0.2cm,height=1.4cm]{00_figures_supp/materials/black_vertical_line.PNG}
    \includegraphics[width=1.65cm,height=1.4cm]{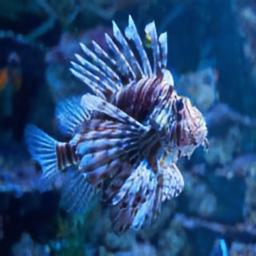}
    \includegraphics[width=1.65cm,height=1.4cm]{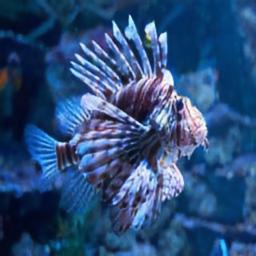}
    \includegraphics[width=1.65cm,height=1.4cm]{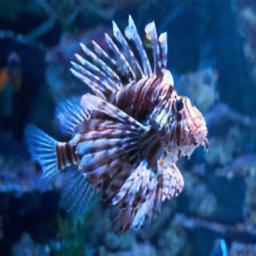}
    \includegraphics[width=1.65cm,height=1.4cm]{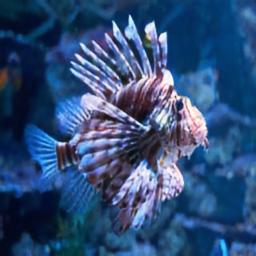}
    \includegraphics[width=1.65cm,height=1.4cm]{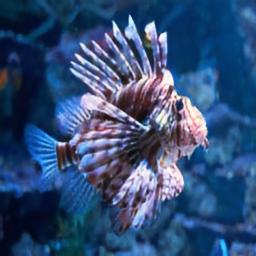}
    \includegraphics[width=1.65cm,height=1.4cm]{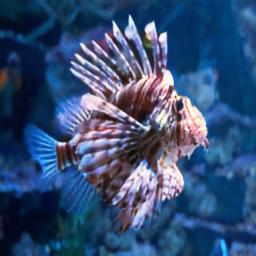}
    \includegraphics[width=1.65cm,height=1.4cm]{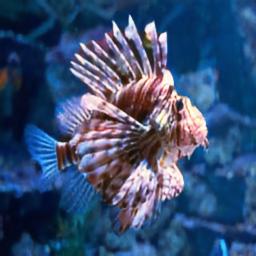}
    \includegraphics[width=1.65cm,height=1.4cm]{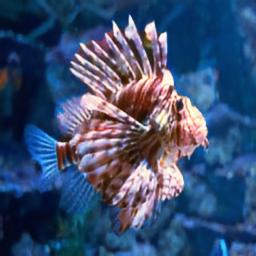}
    \includegraphics[width=0.2cm,height=1.4cm]{00_figures_supp/materials/black_vertical_line.PNG}
    \includegraphics[width=1.65cm,height=1.4cm]{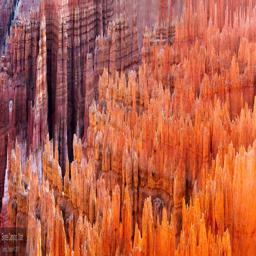}

    \vspace{0.1cm}

    \includegraphics[width=1.65cm,height=1.4cm]{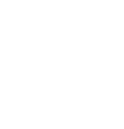}
    \includegraphics[width=0.2cm,height=1.4cm]{00_figures_supp/materials/black_vertical_line.PNG}
    \includegraphics[width=1.65cm,height=1.4cm]{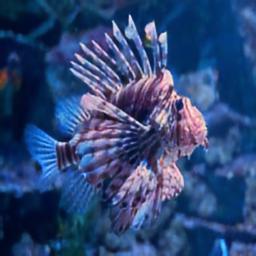}
    \includegraphics[width=1.65cm,height=1.4cm]{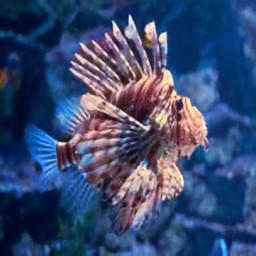}
    \includegraphics[width=1.65cm,height=1.4cm]{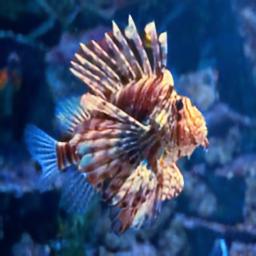}
    \includegraphics[width=1.65cm,height=1.4cm]{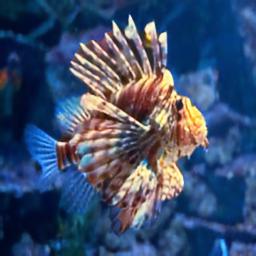}
    \includegraphics[width=1.65cm,height=1.4cm]{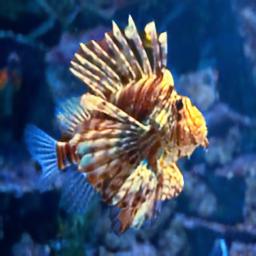}
    \includegraphics[width=1.65cm,height=1.4cm]{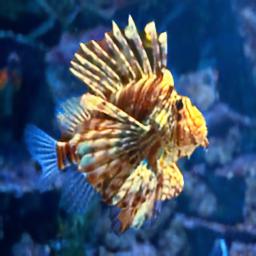}
    \includegraphics[width=1.65cm,height=1.4cm]{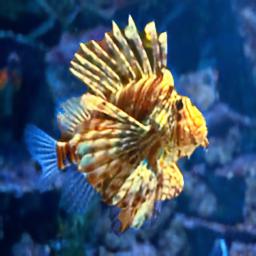}
    \includegraphics[width=1.65cm,height=1.4cm]{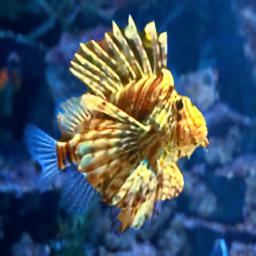}
    \includegraphics[width=0.2cm,height=1.4cm]{00_figures_supp/materials/black_vertical_line.PNG}
    \includegraphics[width=1.65cm,height=1.4cm]{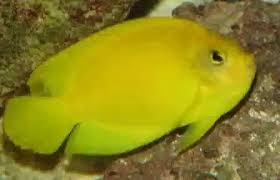}

    \vspace{0.1cm}

    \includegraphics[width=1.65cm,height=1.4cm]{00_figures_supp/materials/white.jpg}
    \includegraphics[width=0.2cm,height=1.4cm]{00_figures_supp/materials/black_vertical_line.PNG}
    \includegraphics[width=1.65cm,height=1.4cm]{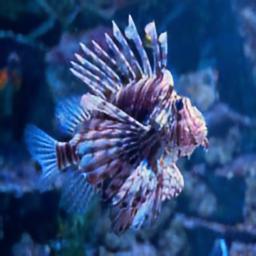}
    \includegraphics[width=1.65cm,height=1.4cm]{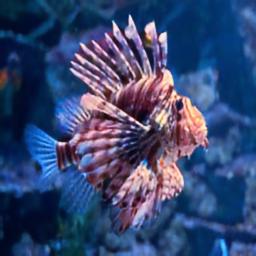}
    \includegraphics[width=1.65cm,height=1.4cm]{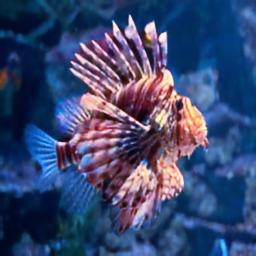}
    \includegraphics[width=1.65cm,height=1.4cm]{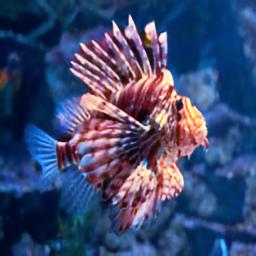}
    \includegraphics[width=1.65cm,height=1.4cm]{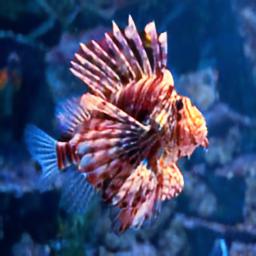}
    \includegraphics[width=1.65cm,height=1.4cm]{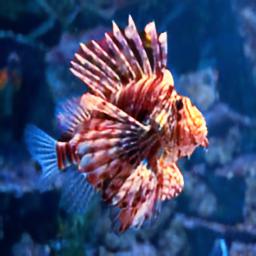}
    \includegraphics[width=1.65cm,height=1.4cm]{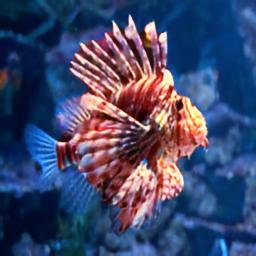}
    \includegraphics[width=1.65cm,height=1.4cm]{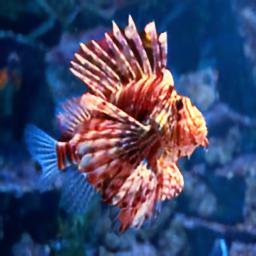}
    \includegraphics[width=0.2cm,height=1.4cm]{00_figures_supp/materials/black_vertical_line.PNG}
    \includegraphics[width=1.65cm,height=1.4cm]{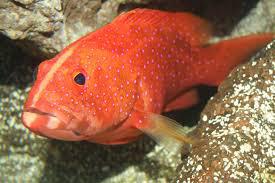}

    \caption{Visualization of ColorCode's color fine-tuning capability. The $\alpha$ represents the weight parameter of the color guidance function $f(x, y, \alpha)$. For the first image of every two rows, the original distorted underwater image $x$ and original enhanced result $\hat{y}_{x}^{m}$ are images at the top and bottom, respectively.}
    \label{fig:guided_by_two_kinds_images_1}
\end{figure*}

\begin{figure*}[!t]
    \centering
    \footnotesize
    \textcolor{black}{\leftline{\hspace{0.4cm} $x \rightarrow \hat{y}_{x}^{m}$ \hspace{0.9cm}  $\alpha = 0.1$ \hspace{0.5cm}  $\alpha = 0.2$  \hspace{0.5cm}   $\alpha = 0.25$ \hspace{0.5cm}  $\alpha = 0.3$  \hspace{0.5cm}  $\alpha = 0.35$  \hspace{0.45cm} $\alpha = 0.4$  \hspace{0.5cm}   $\alpha = 0.45$  \hspace{0.45cm}  $\alpha = 0.5$ \hspace{1.3cm}  $g$}}
    \\

    \vspace{0.02cm}

    \includegraphics[width=1.65cm,height=1.4cm]{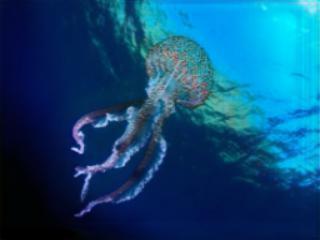}
    \includegraphics[width=0.2cm,height=1.4cm]{00_figures_supp/materials/black_vertical_line.PNG}
    \includegraphics[width=1.65cm,height=1.4cm]{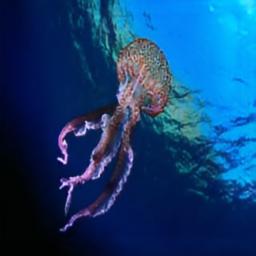}
    \includegraphics[width=1.65cm,height=1.4cm]{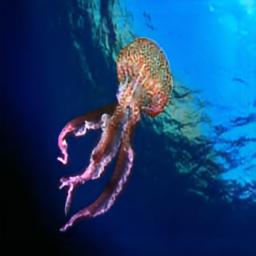}
    \includegraphics[width=1.65cm,height=1.4cm]{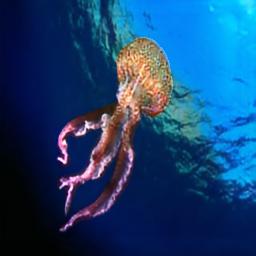}
    \includegraphics[width=1.65cm,height=1.4cm]{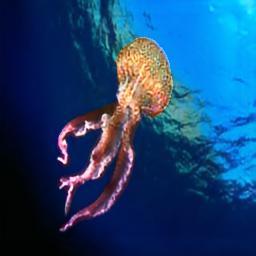}
    \includegraphics[width=1.65cm,height=1.4cm]{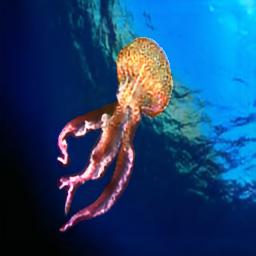}
    \includegraphics[width=1.65cm,height=1.4cm]{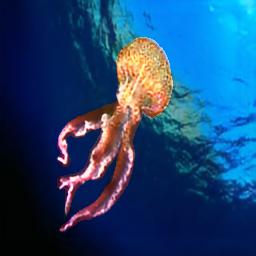}
    \includegraphics[width=1.65cm,height=1.4cm]{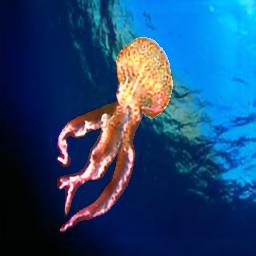}
    \includegraphics[width=1.65cm,height=1.4cm]{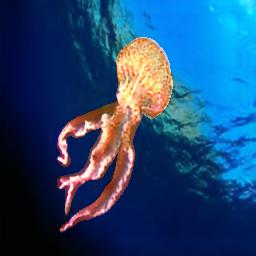}
    \includegraphics[width=0.2cm,height=1.4cm]{00_figures_supp/materials/black_vertical_line.PNG}
    \includegraphics[width=1.65cm,height=1.4cm]{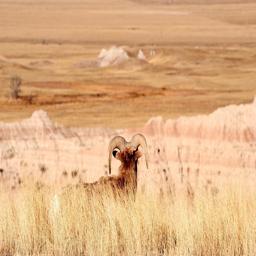}

    \vspace{0.1cm}

    \includegraphics[width=1.65cm,height=1.4cm]{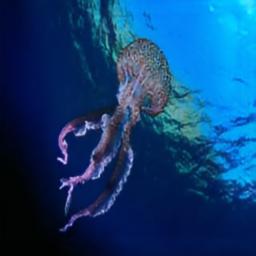}
    \includegraphics[width=0.2cm,height=1.4cm]{00_figures_supp/materials/black_vertical_line.PNG}
    \includegraphics[width=1.65cm,height=1.4cm]{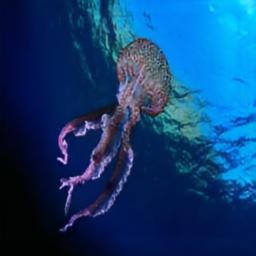}
    \includegraphics[width=1.65cm,height=1.4cm]{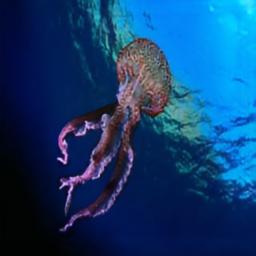}
    \includegraphics[width=1.65cm,height=1.4cm]{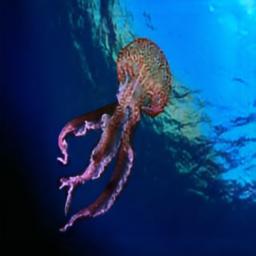}
    \includegraphics[width=1.65cm,height=1.4cm]{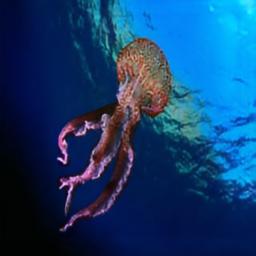}
    \includegraphics[width=1.65cm,height=1.4cm]{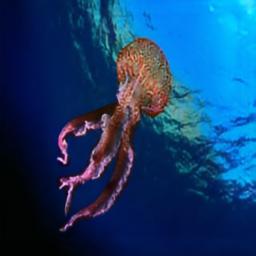}
    \includegraphics[width=1.65cm,height=1.4cm]{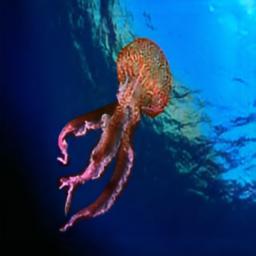}
    \includegraphics[width=1.65cm,height=1.4cm]{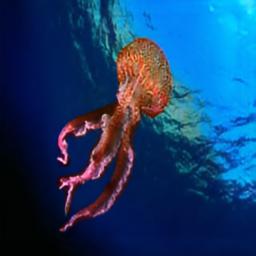}
    \includegraphics[width=1.65cm,height=1.4cm]{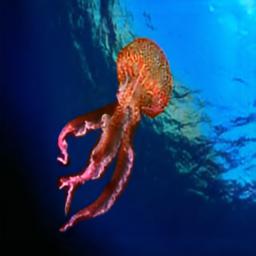}
    \includegraphics[width=0.2cm,height=1.4cm]{00_figures_supp/materials/black_vertical_line.PNG}
    \includegraphics[width=1.65cm,height=1.4cm]{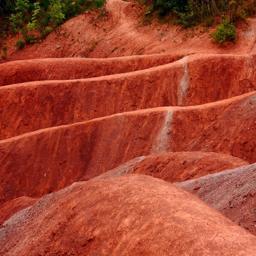}

    \vspace{0.1cm}

    \includegraphics[width=1.65cm,height=1.4cm]{00_figures_supp/materials/white.jpg}
    \includegraphics[width=0.2cm,height=1.4cm]{00_figures_supp/materials/black_vertical_line.PNG}
    \includegraphics[width=1.65cm,height=1.4cm]{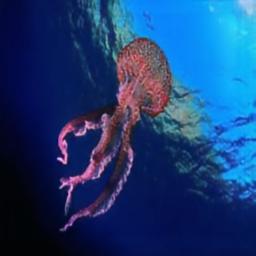}
    \includegraphics[width=1.65cm,height=1.4cm]{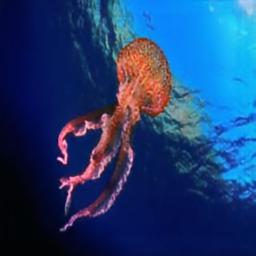}
    \includegraphics[width=1.65cm,height=1.4cm]{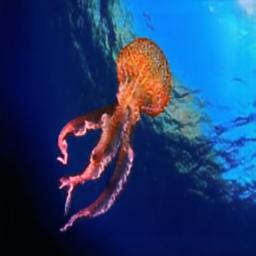}
    \includegraphics[width=1.65cm,height=1.4cm]{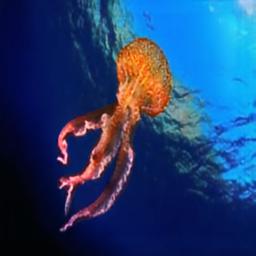}
    \includegraphics[width=1.65cm,height=1.4cm]{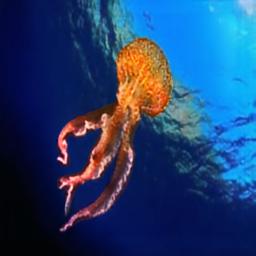}
    \includegraphics[width=1.65cm,height=1.4cm]{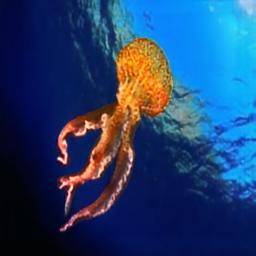}
    \includegraphics[width=1.65cm,height=1.4cm]{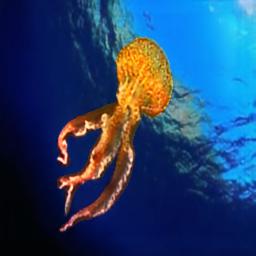}
    \includegraphics[width=1.65cm,height=1.4cm]{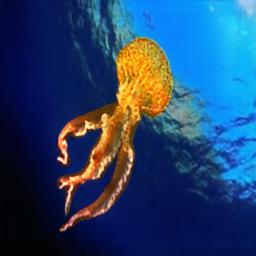}
    \includegraphics[width=0.2cm,height=1.4cm]{00_figures_supp/materials/black_vertical_line.PNG}
    \includegraphics[width=1.65cm,height=1.4cm]{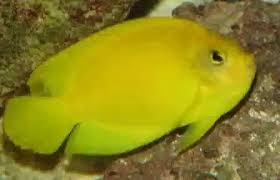}

    \vspace{0.1cm}

    \includegraphics[width=1.65cm,height=1.4cm]{00_figures_supp/materials/white.jpg}
    \includegraphics[width=0.2cm,height=1.4cm]{00_figures_supp/materials/black_vertical_line.PNG}
    \includegraphics[width=1.65cm,height=1.4cm]{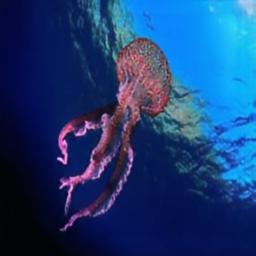}
    \includegraphics[width=1.65cm,height=1.4cm]{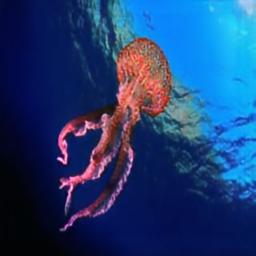}
    \includegraphics[width=1.65cm,height=1.4cm]{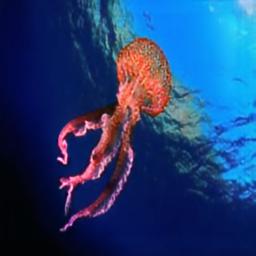}
    \includegraphics[width=1.65cm,height=1.4cm]{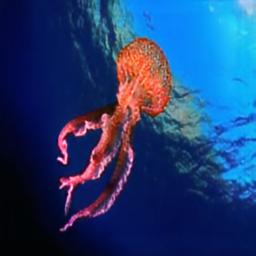}
    \includegraphics[width=1.65cm,height=1.4cm]{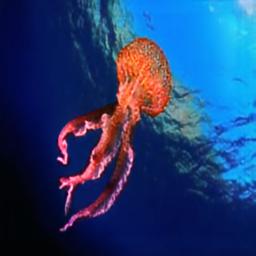}
    \includegraphics[width=1.65cm,height=1.4cm]{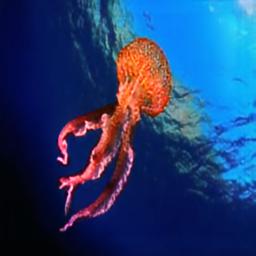}
    \includegraphics[width=1.65cm,height=1.4cm]{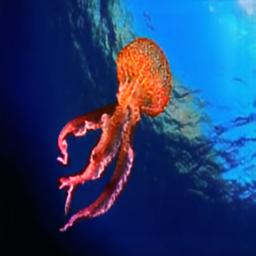}
    \includegraphics[width=1.65cm,height=1.4cm]{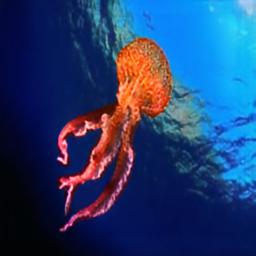}
    \includegraphics[width=0.2cm,height=1.4cm]{00_figures_supp/materials/black_vertical_line.PNG}
    \includegraphics[width=1.65cm,height=1.4cm]{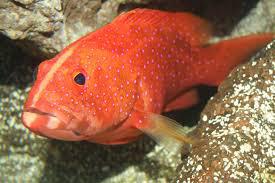}

    \caption{Visualization of ColorCode's color fine-tuning capability. The $\alpha$ represents the weight parameter of the color guidance function $f(x, y, \alpha)$. For the first image of every two rows, the original distorted underwater image $x$ and original enhanced result $\hat{y}_{x}^{m}$ are images at the top and bottom, respectively.}
    \label{fig:guided_by_two_kinds_images_2}
\end{figure*}

\begin{figure*}[!t]
    \centering
    \footnotesize
    \textcolor{black}{\leftline{\hspace{0.4cm} $x \rightarrow \hat{y}_{x}^{m}$ \hspace{0.9cm}  $\alpha = 0.1$ \hspace{0.5cm}  $\alpha = 0.2$  \hspace{0.5cm}   $\alpha = 0.25$ \hspace{0.5cm}  $\alpha = 0.3$  \hspace{0.5cm}  $\alpha = 0.35$  \hspace{0.45cm} $\alpha = 0.4$  \hspace{0.5cm}   $\alpha = 0.45$  \hspace{0.45cm}  $\alpha = 0.5$ \hspace{1.3cm}  $g$}}
    \\

    \vspace{0.02cm}

    \includegraphics[width=1.65cm,height=1.4cm]{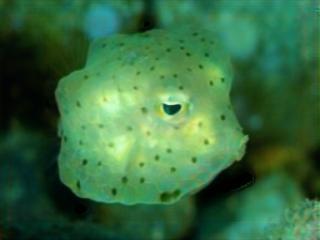}
    \includegraphics[width=0.2cm,height=1.4cm]{00_figures_supp/materials/black_vertical_line.PNG}
    \includegraphics[width=1.65cm,height=1.4cm]{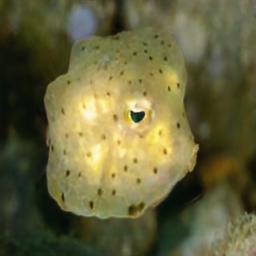}
    \includegraphics[width=1.65cm,height=1.4cm]{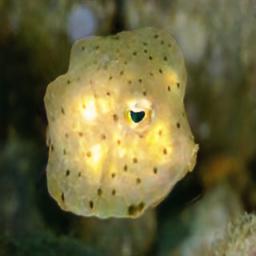}
    \includegraphics[width=1.65cm,height=1.4cm]{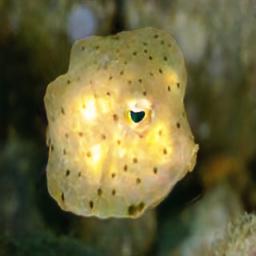}
    \includegraphics[width=1.65cm,height=1.4cm]{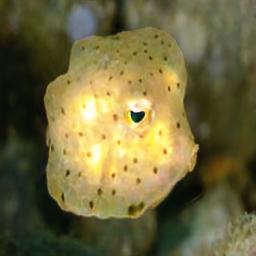}
    \includegraphics[width=1.65cm,height=1.4cm]{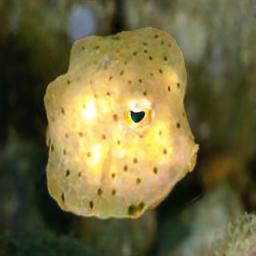}
    \includegraphics[width=1.65cm,height=1.4cm]{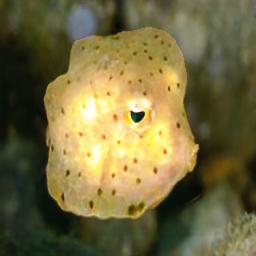}
    \includegraphics[width=1.65cm,height=1.4cm]{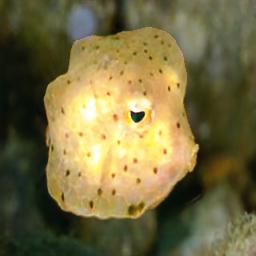}
    \includegraphics[width=1.65cm,height=1.4cm]{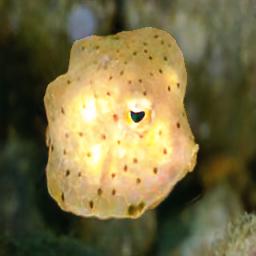}
    \includegraphics[width=0.2cm,height=1.4cm]{00_figures_supp/materials/black_vertical_line.PNG}
    \includegraphics[width=1.65cm,height=1.4cm]{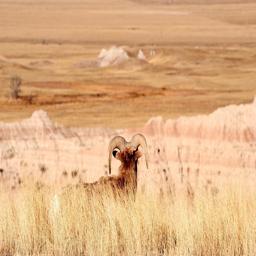}

    \vspace{0.1cm}

    \includegraphics[width=1.65cm,height=1.4cm]{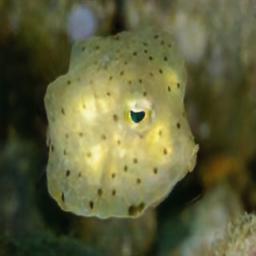}
    \includegraphics[width=0.2cm,height=1.4cm]{00_figures_supp/materials/black_vertical_line.PNG}
    \includegraphics[width=1.65cm,height=1.4cm]{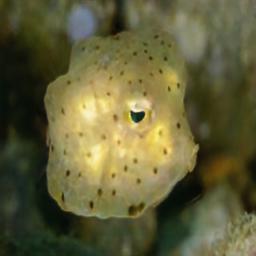}
    \includegraphics[width=1.65cm,height=1.4cm]{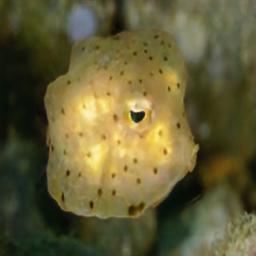}
    \includegraphics[width=1.65cm,height=1.4cm]{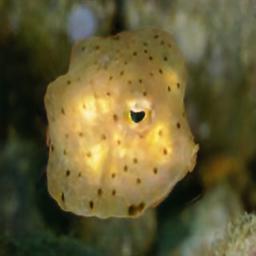}
    \includegraphics[width=1.65cm,height=1.4cm]{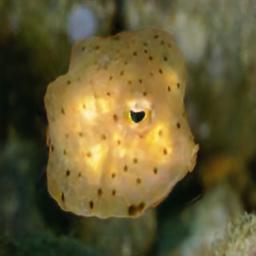}
    \includegraphics[width=1.65cm,height=1.4cm]{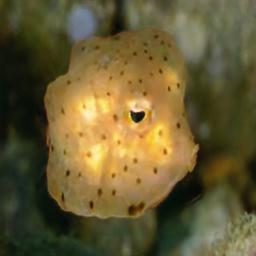}
    \includegraphics[width=1.65cm,height=1.4cm]{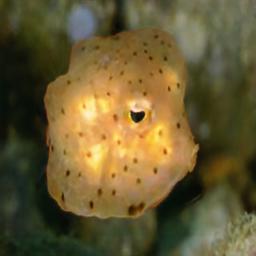}
    \includegraphics[width=1.65cm,height=1.4cm]{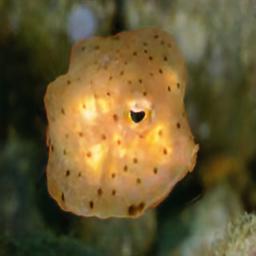}
    \includegraphics[width=1.65cm,height=1.4cm]{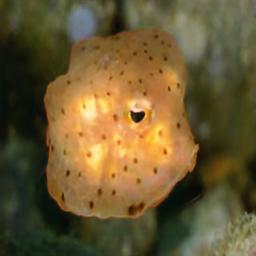}
    \includegraphics[width=0.2cm,height=1.4cm]{00_figures_supp/materials/black_vertical_line.PNG}
    \includegraphics[width=1.65cm,height=1.4cm]{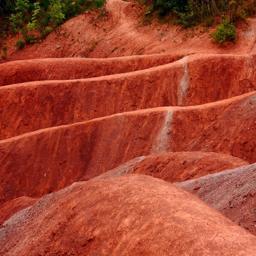}

    \vspace{0.1cm}

    \includegraphics[width=1.65cm,height=1.4cm]{00_figures_supp/materials/white.jpg}
    \includegraphics[width=0.2cm,height=1.4cm]{00_figures_supp/materials/black_vertical_line.PNG}
    \includegraphics[width=1.65cm,height=1.4cm]{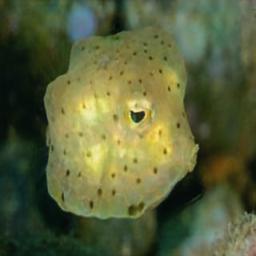}
    \includegraphics[width=1.65cm,height=1.4cm]{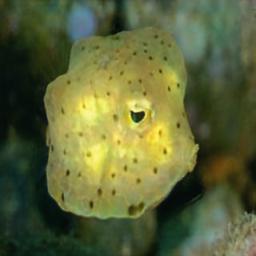}
    \includegraphics[width=1.65cm,height=1.4cm]{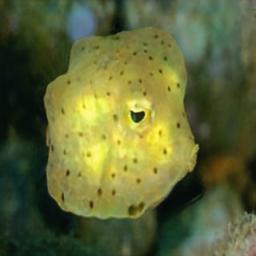}
    \includegraphics[width=1.65cm,height=1.4cm]{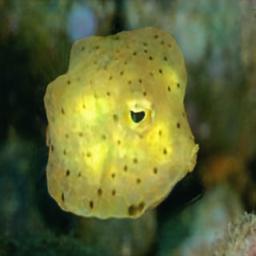}
    \includegraphics[width=1.65cm,height=1.4cm]{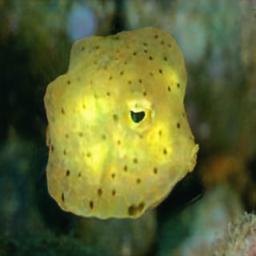}
    \includegraphics[width=1.65cm,height=1.4cm]{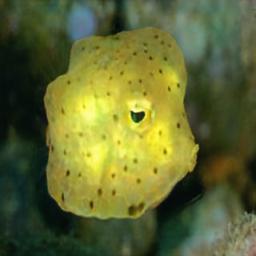}
    \includegraphics[width=1.65cm,height=1.4cm]{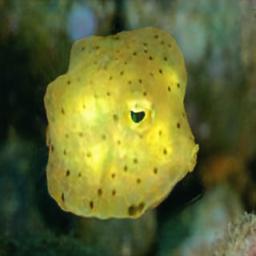}
    \includegraphics[width=1.65cm,height=1.4cm]{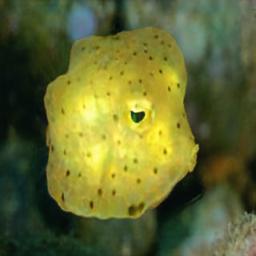}
    \includegraphics[width=0.2cm,height=1.4cm]{00_figures_supp/materials/black_vertical_line.PNG}
    \includegraphics[width=1.65cm,height=1.4cm]{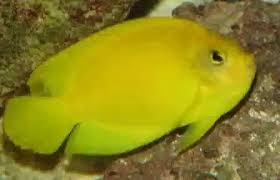}

    \vspace{0.1cm}

    \includegraphics[width=1.65cm,height=1.4cm]{00_figures_supp/materials/white.jpg}
    \includegraphics[width=0.2cm,height=1.4cm]{00_figures_supp/materials/black_vertical_line.PNG}
    \includegraphics[width=1.65cm,height=1.4cm]{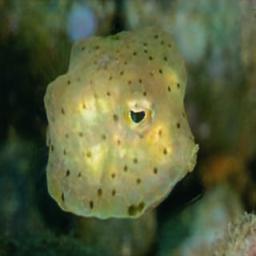}
    \includegraphics[width=1.65cm,height=1.4cm]{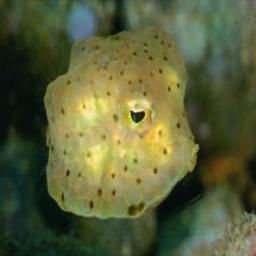}
    \includegraphics[width=1.65cm,height=1.4cm]{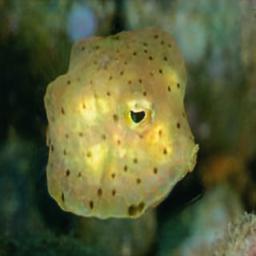}
    \includegraphics[width=1.65cm,height=1.4cm]{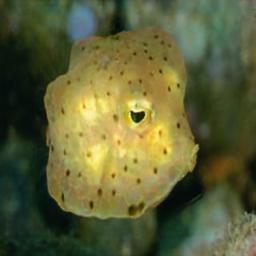}
    \includegraphics[width=1.65cm,height=1.4cm]{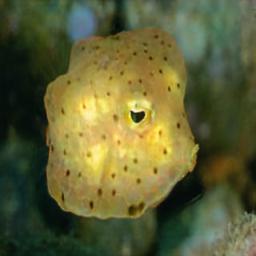}
    \includegraphics[width=1.65cm,height=1.4cm]{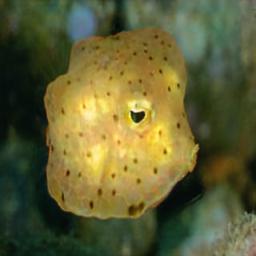}
    \includegraphics[width=1.65cm,height=1.4cm]{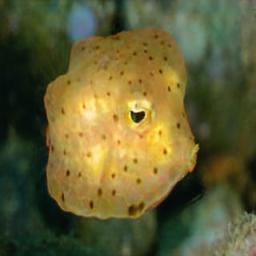}
    \includegraphics[width=1.65cm,height=1.4cm]{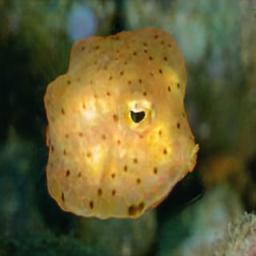}
    \includegraphics[width=0.2cm,height=1.4cm]{00_figures_supp/materials/black_vertical_line.PNG}
    \includegraphics[width=1.65cm,height=1.4cm]{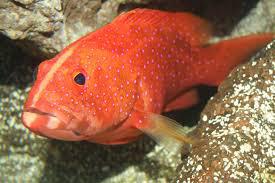}

    \caption{Visualization of ColorCode's color fine-tuning capability. The $\alpha$ represents the weight parameter of the color guidance function $f(x, y, \alpha)$. For the first image of every two rows, the original distorted underwater image $x$ and original enhanced result $\hat{y}_{x}^{m}$ are images at the top and bottom, respectively.}
    \label{fig:guided_by_two_kinds_images_3}
\end{figure*}

\begin{figure*}[!t]
    \centering
    \footnotesize
    \textcolor{black}{\leftline{\hspace{0.4cm} $x \rightarrow \hat{y}_{x}^{m}$ \hspace{0.9cm}  $\alpha = 0.1$ \hspace{0.5cm}  $\alpha = 0.2$  \hspace{0.5cm}   $\alpha = 0.25$ \hspace{0.5cm}  $\alpha = 0.3$  \hspace{0.5cm}  $\alpha = 0.35$  \hspace{0.45cm} $\alpha = 0.4$  \hspace{0.5cm}   $\alpha = 0.45$  \hspace{0.45cm}  $\alpha = 0.5$ \hspace{1.3cm}  $g$}}
    \\

    \vspace{0.02cm}

    \includegraphics[width=1.65cm,height=1.4cm]{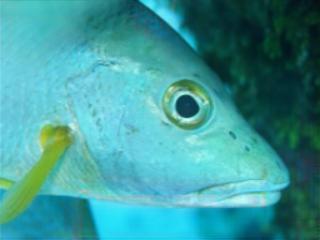}
    \includegraphics[width=0.2cm,height=1.4cm]{00_figures_supp/materials/black_vertical_line.PNG}
    \includegraphics[width=1.65cm,height=1.4cm]{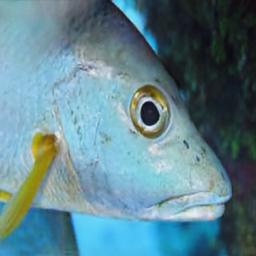}
    \includegraphics[width=1.65cm,height=1.4cm]{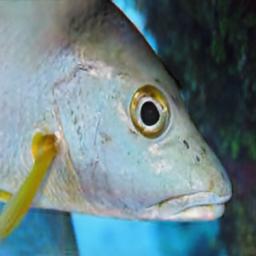}
    \includegraphics[width=1.65cm,height=1.4cm]{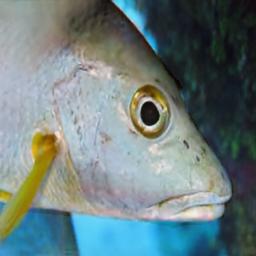}
    \includegraphics[width=1.65cm,height=1.4cm]{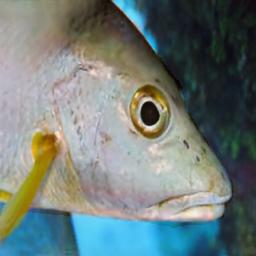}
    \includegraphics[width=1.65cm,height=1.4cm]{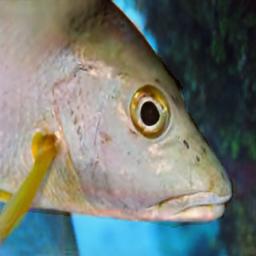}
    \includegraphics[width=1.65cm,height=1.4cm]{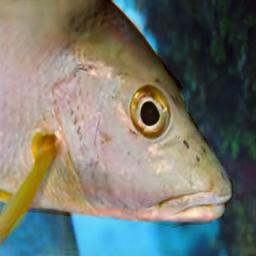}
    \includegraphics[width=1.65cm,height=1.4cm]{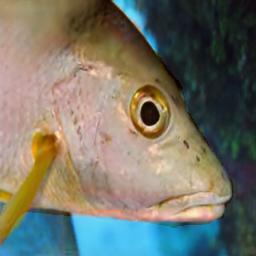}
    \includegraphics[width=1.65cm,height=1.4cm]{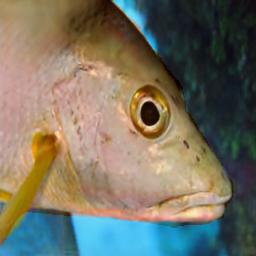}
    \includegraphics[width=0.2cm,height=1.4cm]{00_figures_supp/materials/black_vertical_line.PNG}
    \includegraphics[width=1.65cm,height=1.4cm]{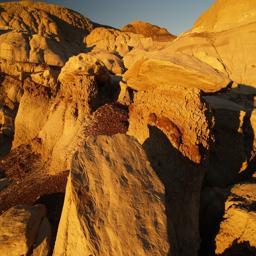}

    \vspace{0.1cm}

    \includegraphics[width=1.65cm,height=1.4cm]{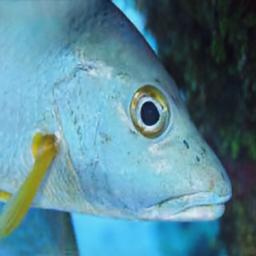}
    \includegraphics[width=0.2cm,height=1.4cm]{00_figures_supp/materials/black_vertical_line.PNG}
    \includegraphics[width=1.65cm,height=1.4cm]{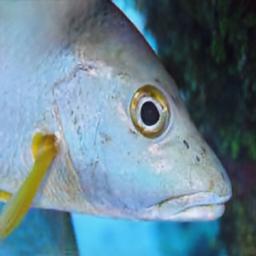}
    \includegraphics[width=1.65cm,height=1.4cm]{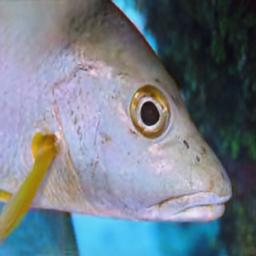}
    \includegraphics[width=1.65cm,height=1.4cm]{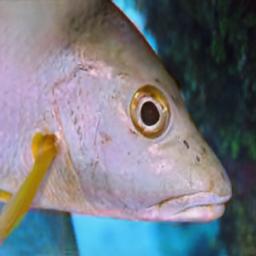}
    \includegraphics[width=1.65cm,height=1.4cm]{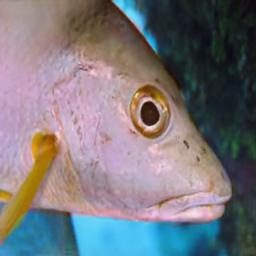}
    \includegraphics[width=1.65cm,height=1.4cm]{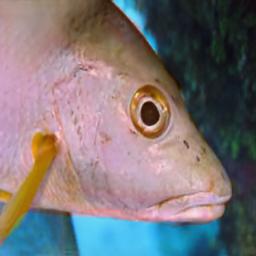}
    \includegraphics[width=1.65cm,height=1.4cm]{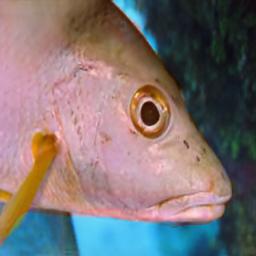}
    \includegraphics[width=1.65cm,height=1.4cm]{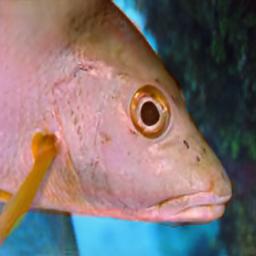}
    \includegraphics[width=1.65cm,height=1.4cm]{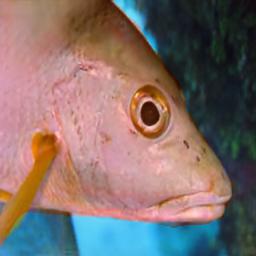}
    \includegraphics[width=0.2cm,height=1.4cm]{00_figures_supp/materials/black_vertical_line.PNG}
    \includegraphics[width=1.65cm,height=1.4cm]{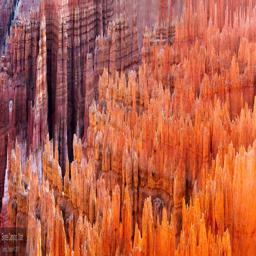}

    \vspace{0.1cm}

    \includegraphics[width=1.65cm,height=1.4cm]{00_figures_supp/materials/white.jpg}
    \includegraphics[width=0.2cm,height=1.4cm]{00_figures_supp/materials/black_vertical_line.PNG}
    \includegraphics[width=1.65cm,height=1.4cm]{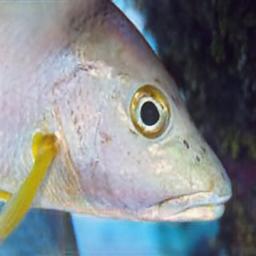}
    \includegraphics[width=1.65cm,height=1.4cm]{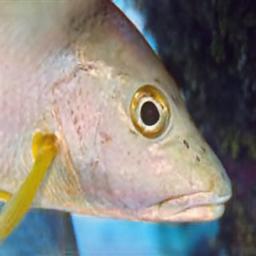}
    \includegraphics[width=1.65cm,height=1.4cm]{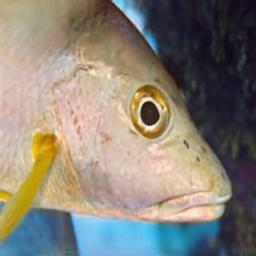}
    \includegraphics[width=1.65cm,height=1.4cm]{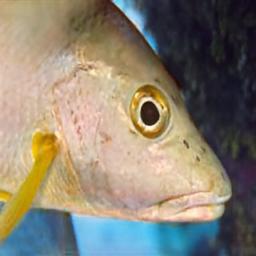}
    \includegraphics[width=1.65cm,height=1.4cm]{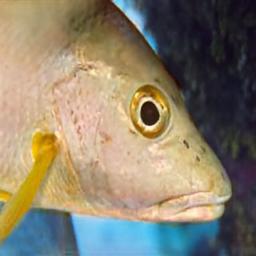}
    \includegraphics[width=1.65cm,height=1.4cm]{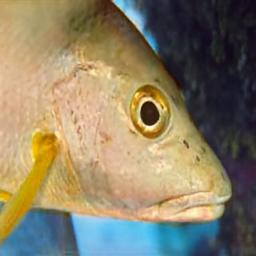}
    \includegraphics[width=1.65cm,height=1.4cm]{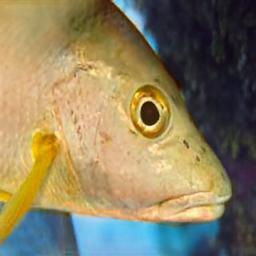}
    \includegraphics[width=1.65cm,height=1.4cm]{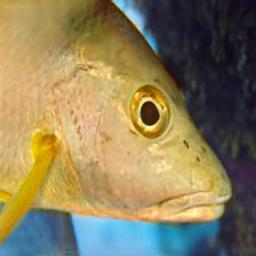}
    \includegraphics[width=0.2cm,height=1.4cm]{00_figures_supp/materials/black_vertical_line.PNG}
    \includegraphics[width=1.65cm,height=1.4cm]{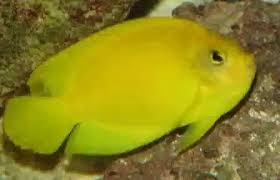}

    \vspace{0.1cm}

    \includegraphics[width=1.65cm,height=1.4cm]{00_figures_supp/materials/white.jpg}
    \includegraphics[width=0.2cm,height=1.4cm]{00_figures_supp/materials/black_vertical_line.PNG}
    \includegraphics[width=1.65cm,height=1.4cm]{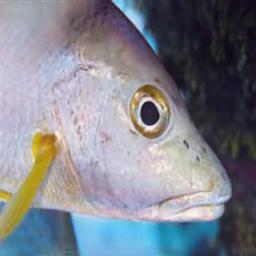}
    \includegraphics[width=1.65cm,height=1.4cm]{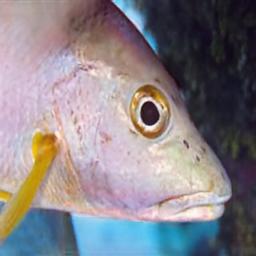}
    \includegraphics[width=1.65cm,height=1.4cm]{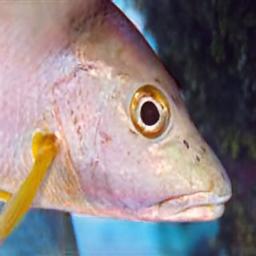}
    \includegraphics[width=1.65cm,height=1.4cm]{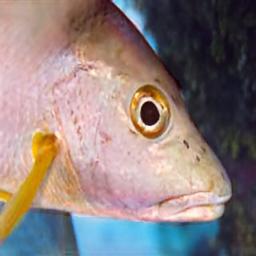}
    \includegraphics[width=1.65cm,height=1.4cm]{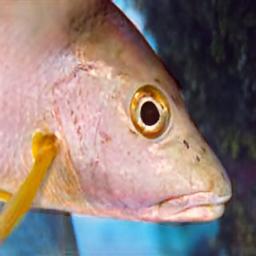}
    \includegraphics[width=1.65cm,height=1.4cm]{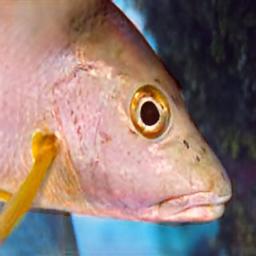}
    \includegraphics[width=1.65cm,height=1.4cm]{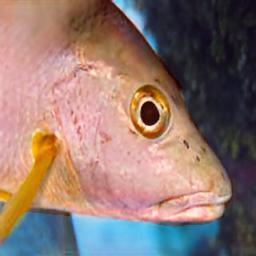}
    \includegraphics[width=1.65cm,height=1.4cm]{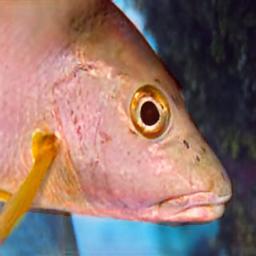}
    \includegraphics[width=0.2cm,height=1.4cm]{00_figures_supp/materials/black_vertical_line.PNG}
    \includegraphics[width=1.65cm,height=1.4cm]{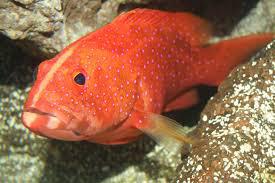}

    \caption{Visualization of ColorCode's color fine-tuning capability. The $\alpha$ represents the weight parameter of the color guidance function $f(x, y, \alpha)$. For the first image of every two rows, the original distorted underwater image $x$ and original enhanced result $\hat{y}_{x}^{m}$ are images at the top and bottom, respectively.}
    \label{fig:guided_by_two_kinds_images_4}
\end{figure*}

\begin{figure*}[!t]
    \centering
    \footnotesize
    \textcolor{black}{\leftline{\hspace{0.4cm} $x \rightarrow \hat{y}_{x}^{m}$ \hspace{0.9cm}  $\alpha = 0.1$ \hspace{0.5cm}  $\alpha = 0.2$  \hspace{0.5cm}   $\alpha = 0.25$ \hspace{0.5cm}  $\alpha = 0.3$  \hspace{0.5cm}  $\alpha = 0.35$  \hspace{0.45cm} $\alpha = 0.4$  \hspace{0.5cm}   $\alpha = 0.45$  \hspace{0.45cm}  $\alpha = 0.5$ \hspace{1.3cm}  $g$}}
    \\

    \vspace{0.02cm}

    \includegraphics[width=1.65cm,height=1.4cm]{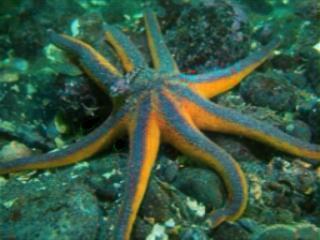}
    \includegraphics[width=0.2cm,height=1.4cm]{00_figures_supp/materials/black_vertical_line.PNG}
    \includegraphics[width=1.65cm,height=1.4cm]{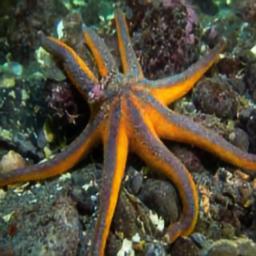}
    \includegraphics[width=1.65cm,height=1.4cm]{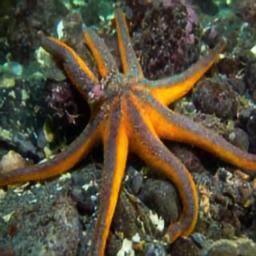}
    \includegraphics[width=1.65cm,height=1.4cm]{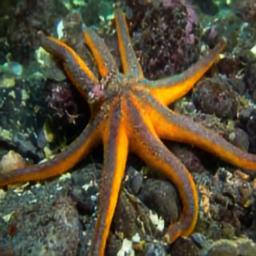}
    \includegraphics[width=1.65cm,height=1.4cm]{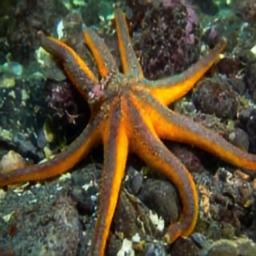}
    \includegraphics[width=1.65cm,height=1.4cm]{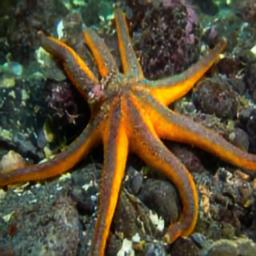}
    \includegraphics[width=1.65cm,height=1.4cm]{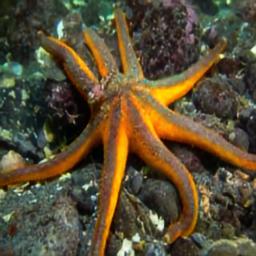}
    \includegraphics[width=1.65cm,height=1.4cm]{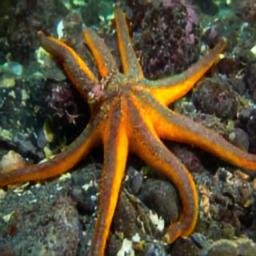}
    \includegraphics[width=1.65cm,height=1.4cm]{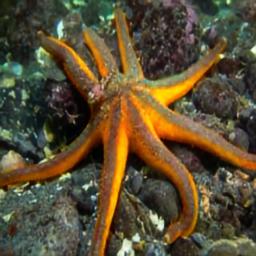}
    \includegraphics[width=0.2cm,height=1.4cm]{00_figures_supp/materials/black_vertical_line.PNG}
    \includegraphics[width=1.65cm,height=1.4cm]{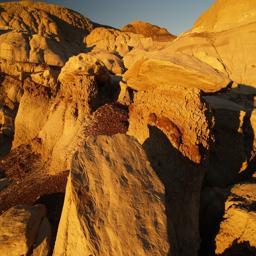}

    \vspace{0.1cm}

    \includegraphics[width=1.65cm,height=1.4cm]{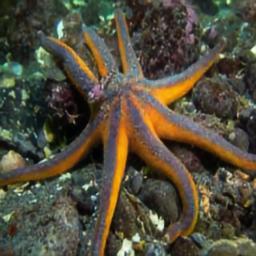}
    \includegraphics[width=0.2cm,height=1.4cm]{00_figures_supp/materials/black_vertical_line.PNG}
    \includegraphics[width=1.65cm,height=1.4cm]{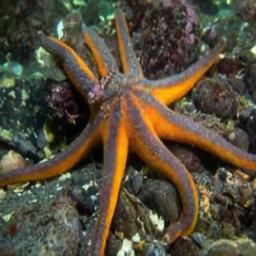}
    \includegraphics[width=1.65cm,height=1.4cm]{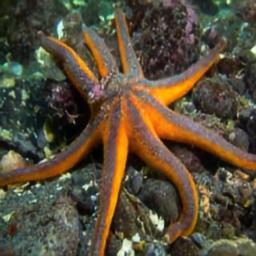}
    \includegraphics[width=1.65cm,height=1.4cm]{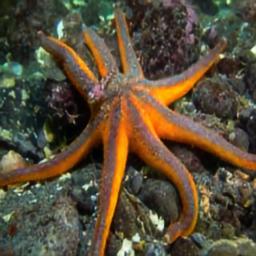}
    \includegraphics[width=1.65cm,height=1.4cm]{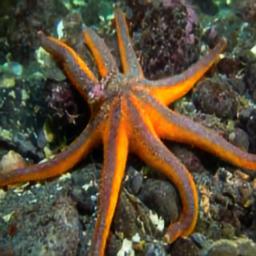}
    \includegraphics[width=1.65cm,height=1.4cm]{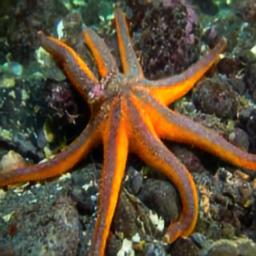}
    \includegraphics[width=1.65cm,height=1.4cm]{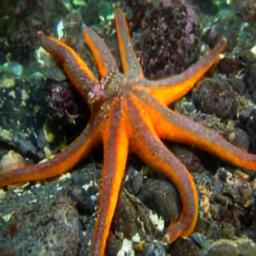}
    \includegraphics[width=1.65cm,height=1.4cm]{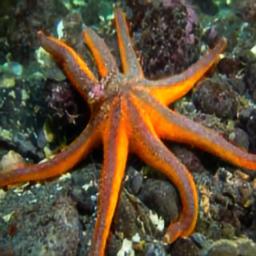}
    \includegraphics[width=1.65cm,height=1.4cm]{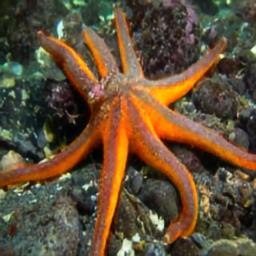}
    \includegraphics[width=0.2cm,height=1.4cm]{00_figures_supp/materials/black_vertical_line.PNG}
    \includegraphics[width=1.65cm,height=1.4cm]{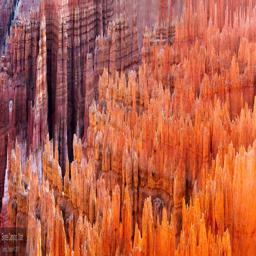}

    \vspace{0.1cm}

    \includegraphics[width=1.65cm,height=1.4cm]{00_figures_supp/materials/white.jpg}
    \includegraphics[width=0.2cm,height=1.4cm]{00_figures_supp/materials/black_vertical_line.PNG}
    \includegraphics[width=1.65cm,height=1.4cm]{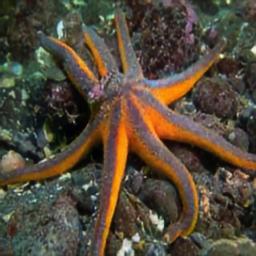}
    \includegraphics[width=1.65cm,height=1.4cm]{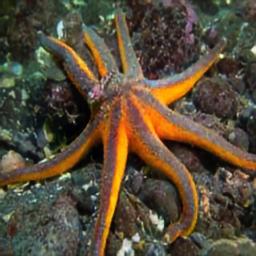}
    \includegraphics[width=1.65cm,height=1.4cm]{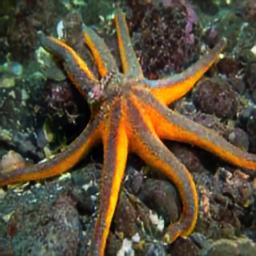}
    \includegraphics[width=1.65cm,height=1.4cm]{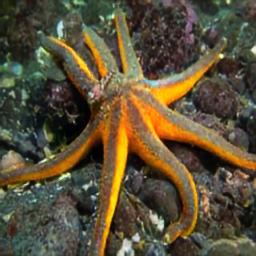}
    \includegraphics[width=1.65cm,height=1.4cm]{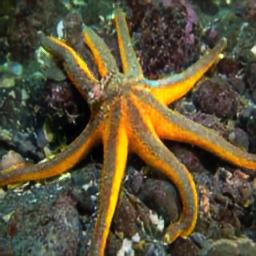}
    \includegraphics[width=1.65cm,height=1.4cm]{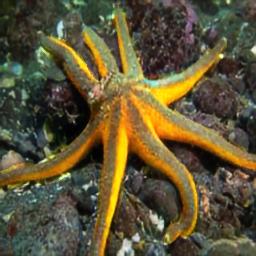}
    \includegraphics[width=1.65cm,height=1.4cm]{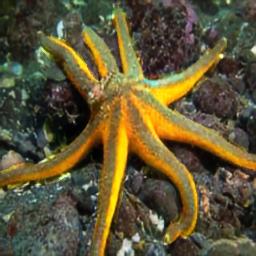}
    \includegraphics[width=1.65cm,height=1.4cm]{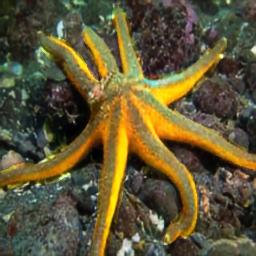}
    \includegraphics[width=0.2cm,height=1.4cm]{00_figures_supp/materials/black_vertical_line.PNG}
    \includegraphics[width=1.65cm,height=1.4cm]{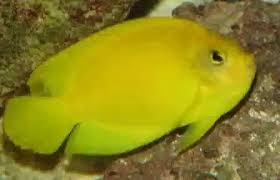}

    \vspace{0.1cm}

    \includegraphics[width=1.65cm,height=1.4cm]{00_figures_supp/materials/white.jpg}
    \includegraphics[width=0.2cm,height=1.4cm]{00_figures_supp/materials/black_vertical_line.PNG}
    \includegraphics[width=1.65cm,height=1.4cm]{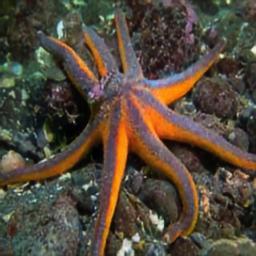}
    \includegraphics[width=1.65cm,height=1.4cm]{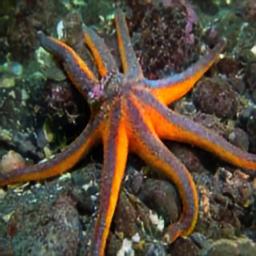}
    \includegraphics[width=1.65cm,height=1.4cm]{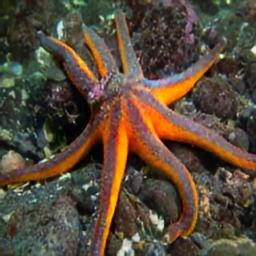}
    \includegraphics[width=1.65cm,height=1.4cm]{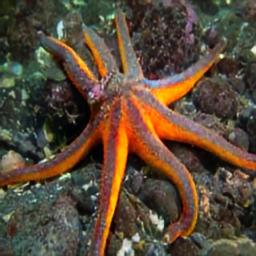}
    \includegraphics[width=1.65cm,height=1.4cm]{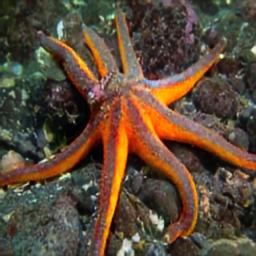}
    \includegraphics[width=1.65cm,height=1.4cm]{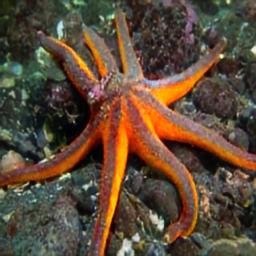}
    \includegraphics[width=1.65cm,height=1.4cm]{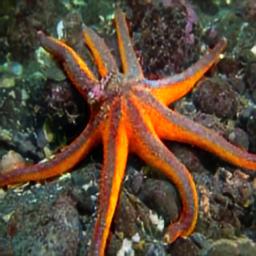}
    \includegraphics[width=1.65cm,height=1.4cm]{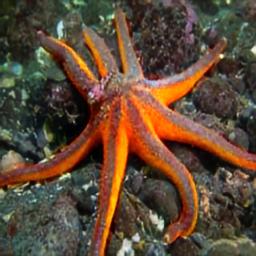}
    \includegraphics[width=0.2cm,height=1.4cm]{00_figures_supp/materials/black_vertical_line.PNG}
    \includegraphics[width=1.65cm,height=1.4cm]{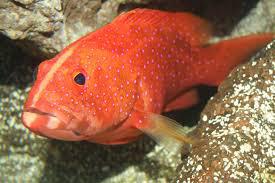}

    \caption{Visualization of ColorCode's color fine-tuning capability. The $\alpha$ represents the weight parameter of the color guidance function $f(x, y, \alpha)$. For the first image of every two rows, the original distorted underwater image $x$ and original enhanced result $\hat{y}_{x}^{m}$ are images at the top and bottom, respectively.}
    \label{fig:guided_by_two_kinds_images_5}
\end{figure*}

\begin{figure*}[!t]
    \centering
    \footnotesize
    \textcolor{black}{\leftline{\hspace{0.4cm} $x \rightarrow \hat{y}_{x}^{m}$ \hspace{0.9cm}  $\alpha = 0.1$ \hspace{0.5cm}  $\alpha = 0.2$  \hspace{0.5cm}   $\alpha = 0.25$ \hspace{0.5cm}  $\alpha = 0.3$  \hspace{0.5cm}  $\alpha = 0.35$  \hspace{0.45cm} $\alpha = 0.4$  \hspace{0.5cm}   $\alpha = 0.45$  \hspace{0.45cm}  $\alpha = 0.5$ \hspace{1.3cm}  $g$}}
    \\

    \vspace{0.02cm}

    \includegraphics[width=1.65cm,height=1.4cm]{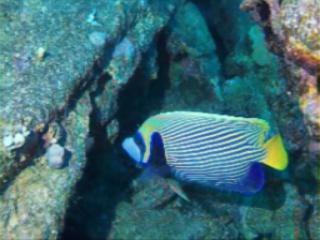}
    \includegraphics[width=0.2cm,height=1.4cm]{00_figures_supp/materials/black_vertical_line.PNG}
    \includegraphics[width=1.65cm,height=1.4cm]{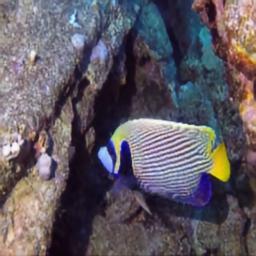}
    \includegraphics[width=1.65cm,height=1.4cm]{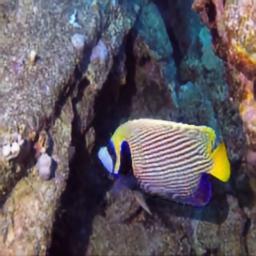}
    \includegraphics[width=1.65cm,height=1.4cm]{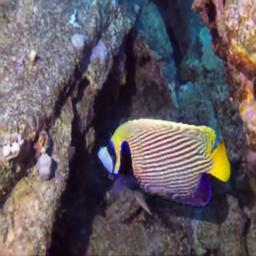}
    \includegraphics[width=1.65cm,height=1.4cm]{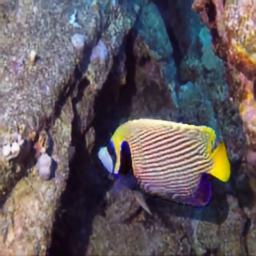}
    \includegraphics[width=1.65cm,height=1.4cm]{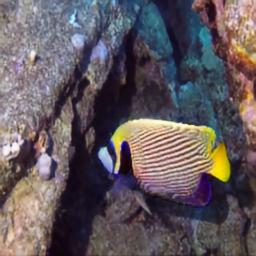}
    \includegraphics[width=1.65cm,height=1.4cm]{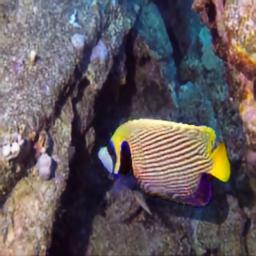}
    \includegraphics[width=1.65cm,height=1.4cm]{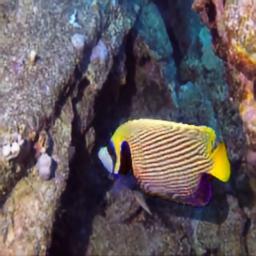}
    \includegraphics[width=1.65cm,height=1.4cm]{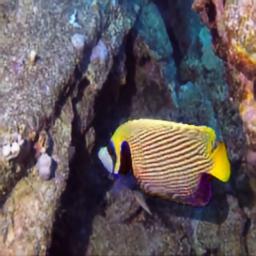}
    \includegraphics[width=0.2cm,height=1.4cm]{00_figures_supp/materials/black_vertical_line.PNG}
    \includegraphics[width=1.65cm,height=1.4cm]{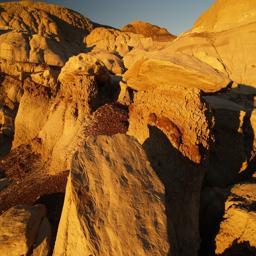}

    \vspace{0.1cm}

    \includegraphics[width=1.65cm,height=1.4cm]{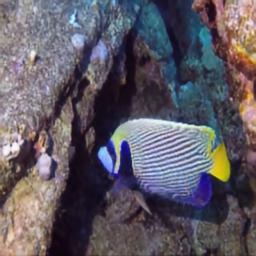}
    \includegraphics[width=0.2cm,height=1.4cm]{00_figures_supp/materials/black_vertical_line.PNG}
    \includegraphics[width=1.65cm,height=1.4cm]{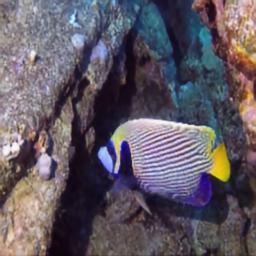}
    \includegraphics[width=1.65cm,height=1.4cm]{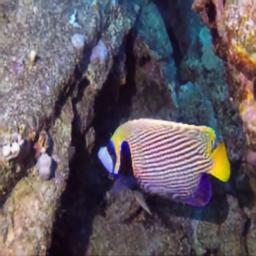}
    \includegraphics[width=1.65cm,height=1.4cm]{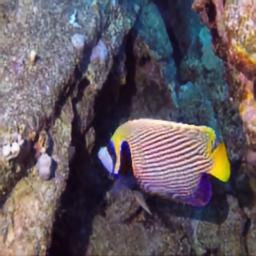}
    \includegraphics[width=1.65cm,height=1.4cm]{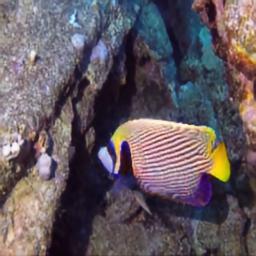}
    \includegraphics[width=1.65cm,height=1.4cm]{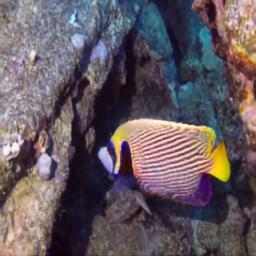}
    \includegraphics[width=1.65cm,height=1.4cm]{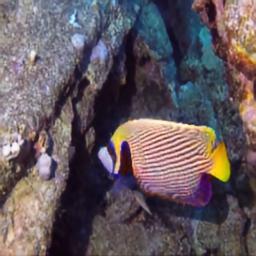}
    \includegraphics[width=1.65cm,height=1.4cm]{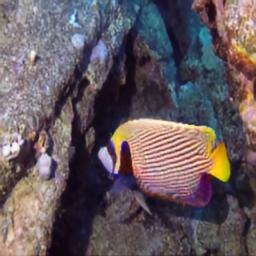}
    \includegraphics[width=1.65cm,height=1.4cm]{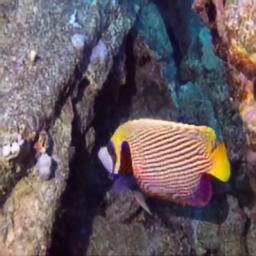}
    \includegraphics[width=0.2cm,height=1.4cm]{00_figures_supp/materials/black_vertical_line.PNG}
    \includegraphics[width=1.65cm,height=1.4cm]{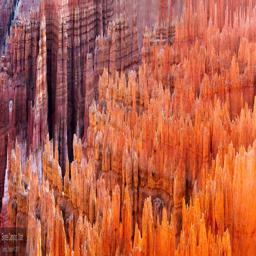}

    \vspace{0.1cm}

    \includegraphics[width=1.65cm,height=1.4cm]{00_figures_supp/materials/white.jpg}
    \includegraphics[width=0.2cm,height=1.4cm]{00_figures_supp/materials/black_vertical_line.PNG}
    \includegraphics[width=1.65cm,height=1.4cm]{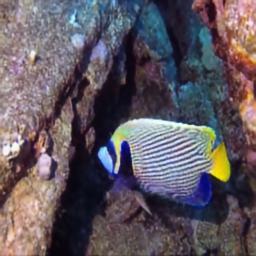}
    \includegraphics[width=1.65cm,height=1.4cm]{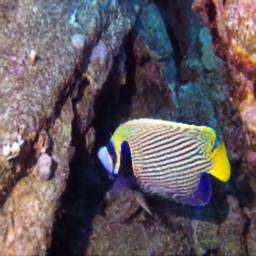}
    \includegraphics[width=1.65cm,height=1.4cm]{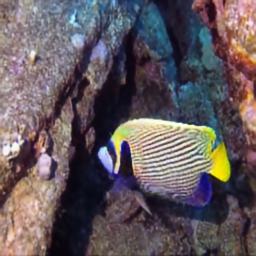}
    \includegraphics[width=1.65cm,height=1.4cm]{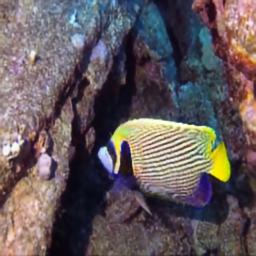}
    \includegraphics[width=1.65cm,height=1.4cm]{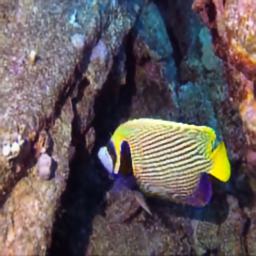}
    \includegraphics[width=1.65cm,height=1.4cm]{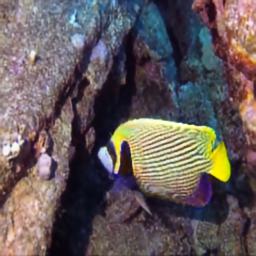}
    \includegraphics[width=1.65cm,height=1.4cm]{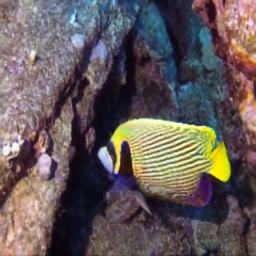}
    \includegraphics[width=1.65cm,height=1.4cm]{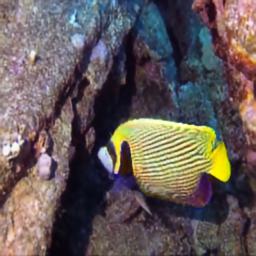}
    \includegraphics[width=0.2cm,height=1.4cm]{00_figures_supp/materials/black_vertical_line.PNG}
    \includegraphics[width=1.65cm,height=1.4cm]{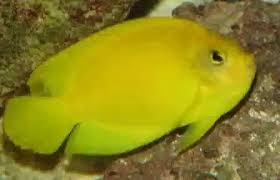}

    \vspace{0.1cm}

    \includegraphics[width=1.65cm,height=1.4cm]{00_figures_supp/materials/white.jpg}
    \includegraphics[width=0.2cm,height=1.4cm]{00_figures_supp/materials/black_vertical_line.PNG}
    \includegraphics[width=1.65cm,height=1.4cm]{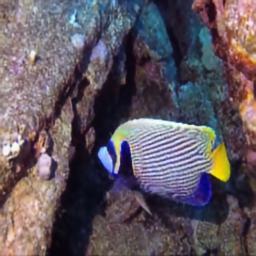}
    \includegraphics[width=1.65cm,height=1.4cm]{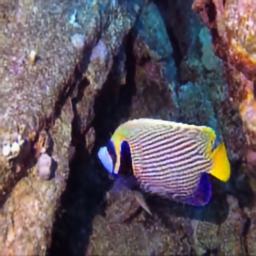}
    \includegraphics[width=1.65cm,height=1.4cm]{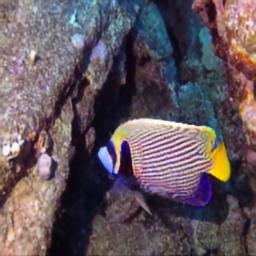}
    \includegraphics[width=1.65cm,height=1.4cm]{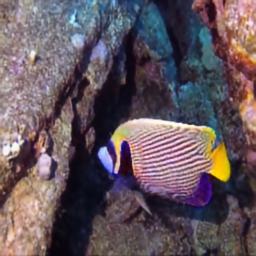}
    \includegraphics[width=1.65cm,height=1.4cm]{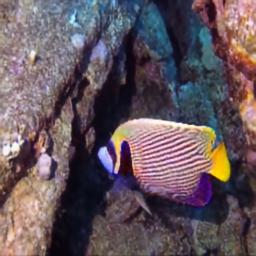}
    \includegraphics[width=1.65cm,height=1.4cm]{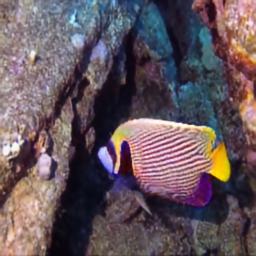}
    \includegraphics[width=1.65cm,height=1.4cm]{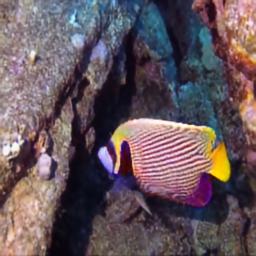}
    \includegraphics[width=1.65cm,height=1.4cm]{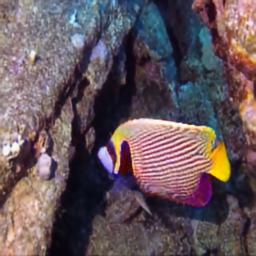}
    \includegraphics[width=0.2cm,height=1.4cm]{00_figures_supp/materials/black_vertical_line.PNG}
    \includegraphics[width=1.65cm,height=1.4cm]{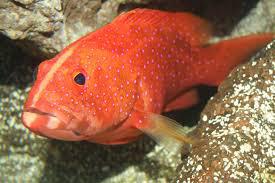}

    \caption{Visualization of ColorCode's color fine-tuning capability. The $\alpha$ represents the weight parameter of the color guidance function $f(x, y, \alpha)$. For the first image of every two rows, the original distorted underwater image $x$ and original enhanced result $\hat{y}_{x}^{m}$ are images at the top and bottom, respectively.}
    \label{fig:guided_by_two_kinds_images_6}
\end{figure*}

Fig. \ref{fig:guided_by_two_kinds_images_1}, Fig. \ref{fig:guided_by_two_kinds_images_2}, Fig. \ref{fig:guided_by_two_kinds_images_3}, Fig. \ref{fig:guided_by_two_kinds_images_4}, Fig. \ref{fig:guided_by_two_kinds_images_5} and Fig. \ref{fig:guided_by_two_kinds_images_6} show the color fine-tuning results on multiple scenes. The results show that the colors of underwater organisms can be fine-tuned when images with long-wavelength colors are used as guidance. The guidance can be underwater images or natural images. The color fine-tuning capability of ColorCode is further verified.

\subsection{More Color Interpolation Results}

\begin{figure*}
    \small
    \centering
    \includegraphics[width=18cm,height=10cm]{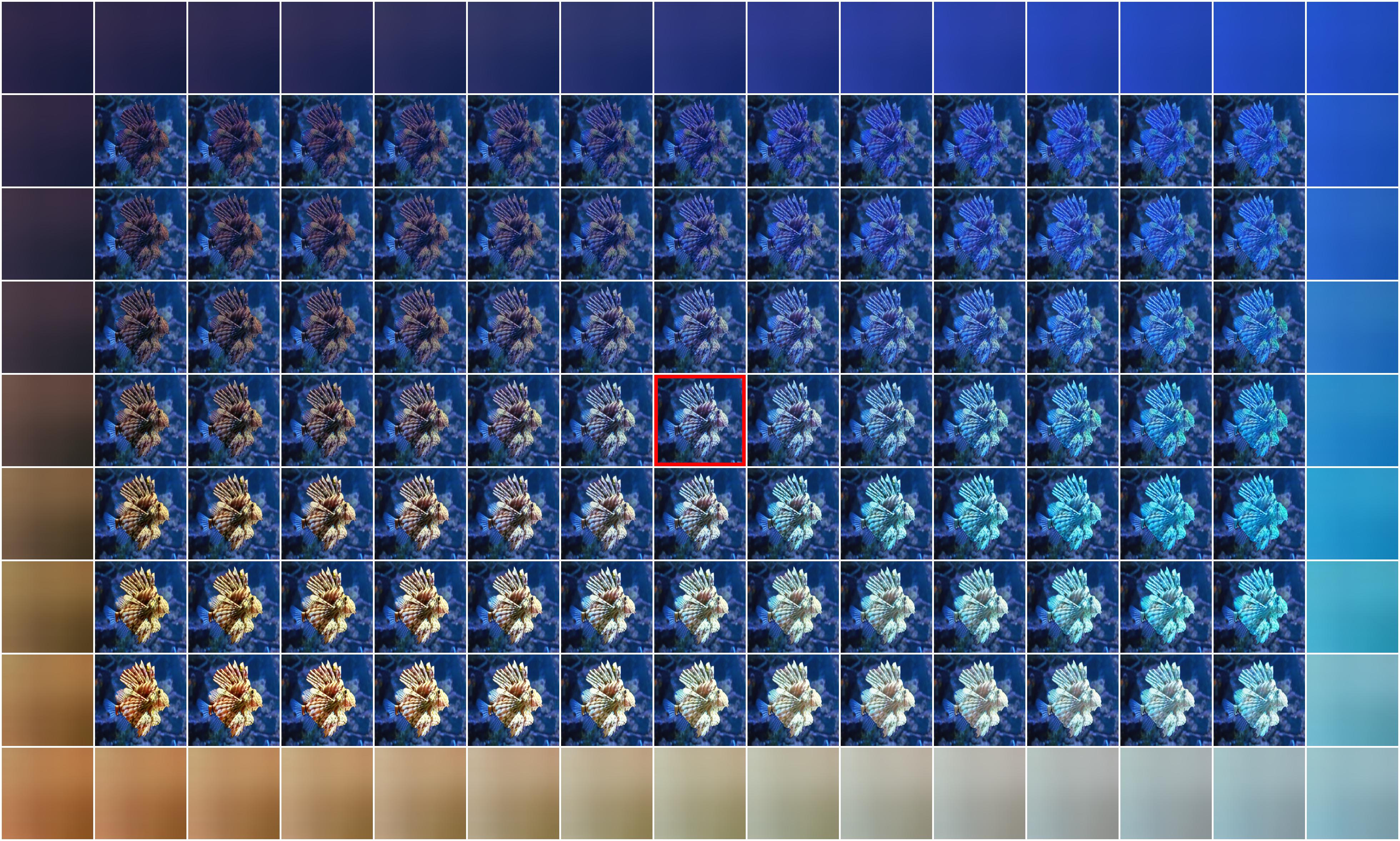}

    \vspace{0.5cm}

    \includegraphics[width=18cm,height=10cm]{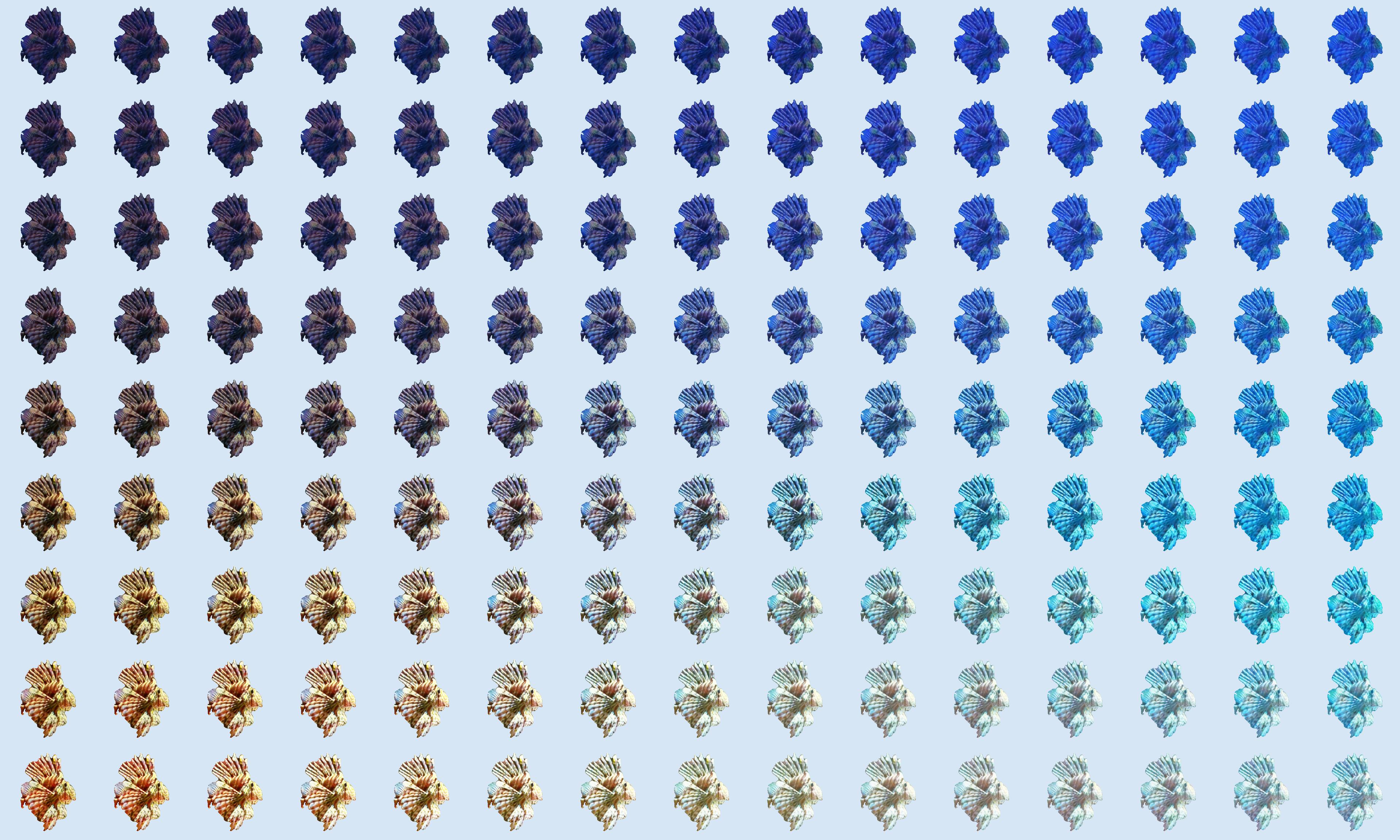}
    \caption{Diversification results obtained by sampling color codes.}
    \label{fig:color_code_interpolation_supp_0}
  \end{figure*}

\begin{figure*}
    \small
    \centering
    \includegraphics[width=18cm,height=10cm]{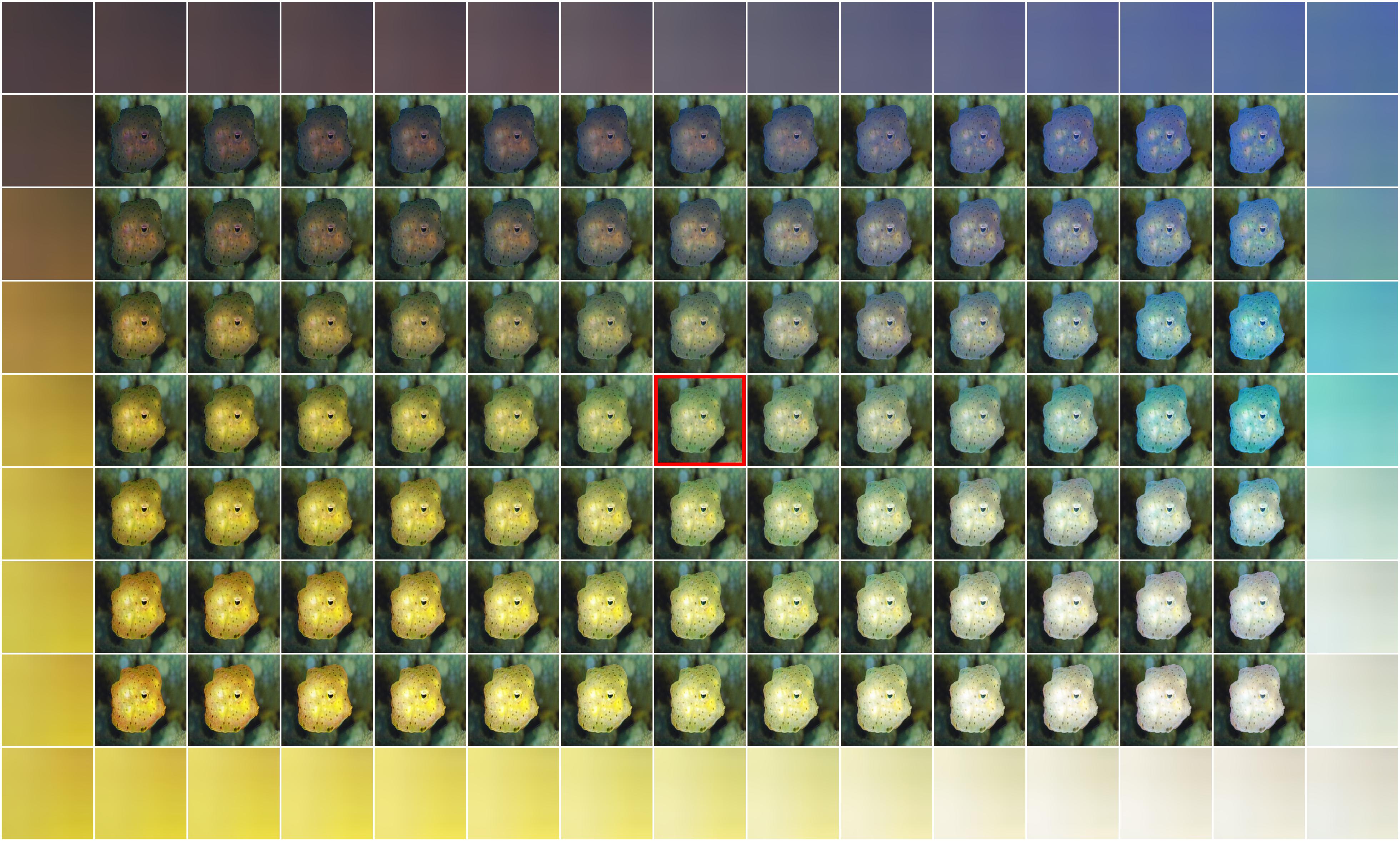}

    \vspace{0.5cm}

    \includegraphics[width=18cm,height=10cm]{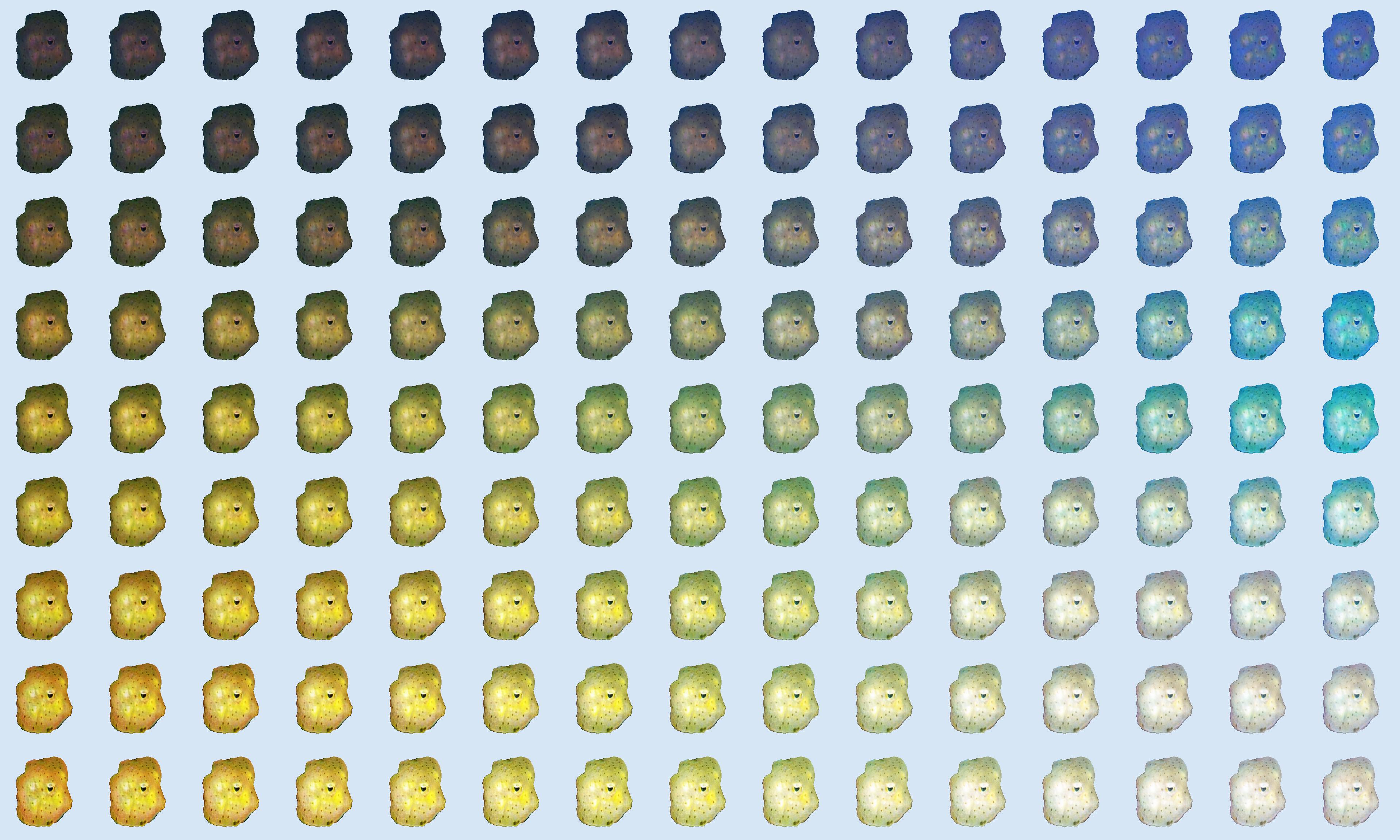}
    \caption{Diversification results obtained by sampling color codes.}
    \label{fig:color_code_interpolation_supp_1}
  \end{figure*}

\begin{figure*}
    \small
    \centering
    \includegraphics[width=18cm,height=10cm]{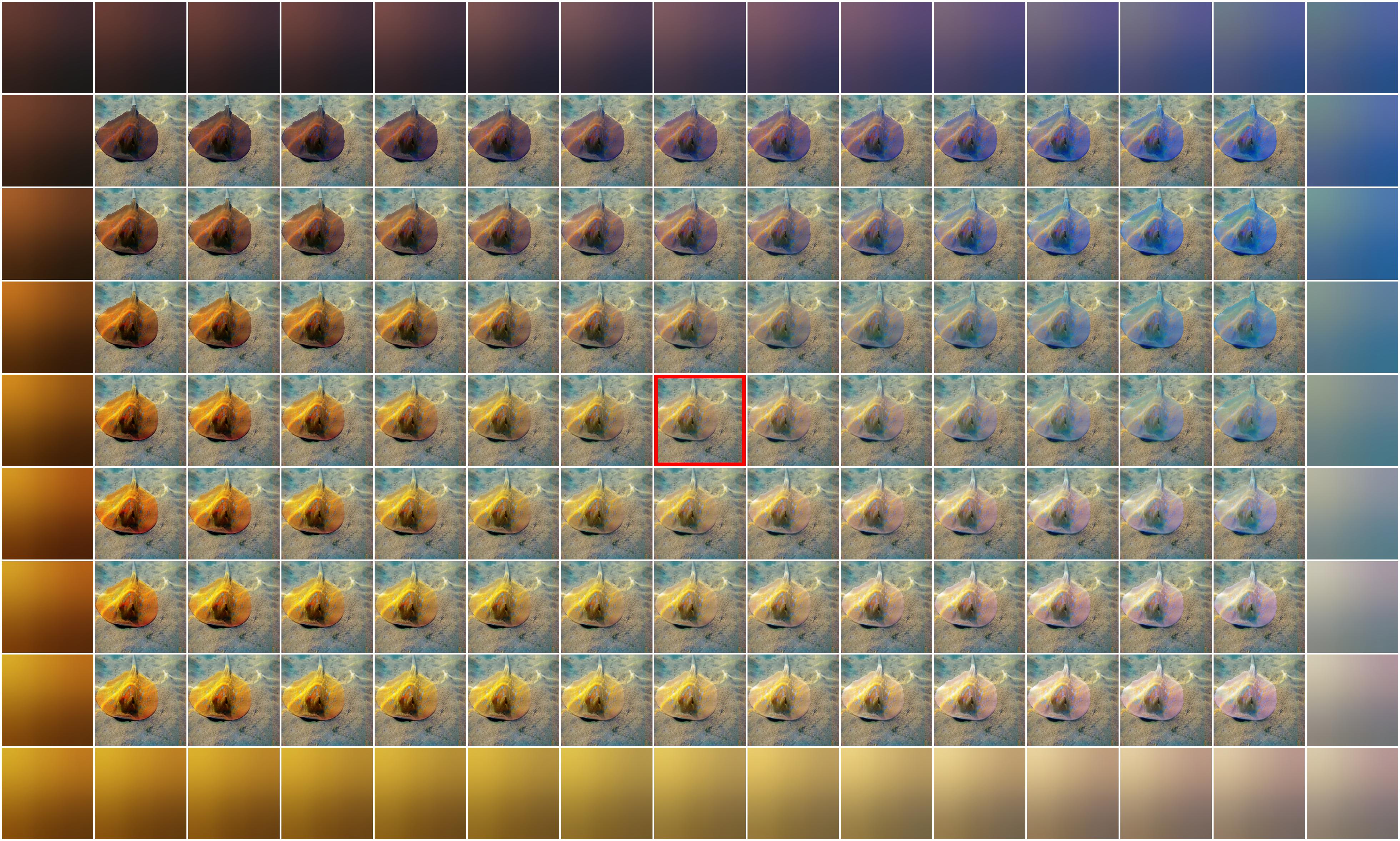}

    \vspace{0.5cm}

    \includegraphics[width=18cm,height=10cm]{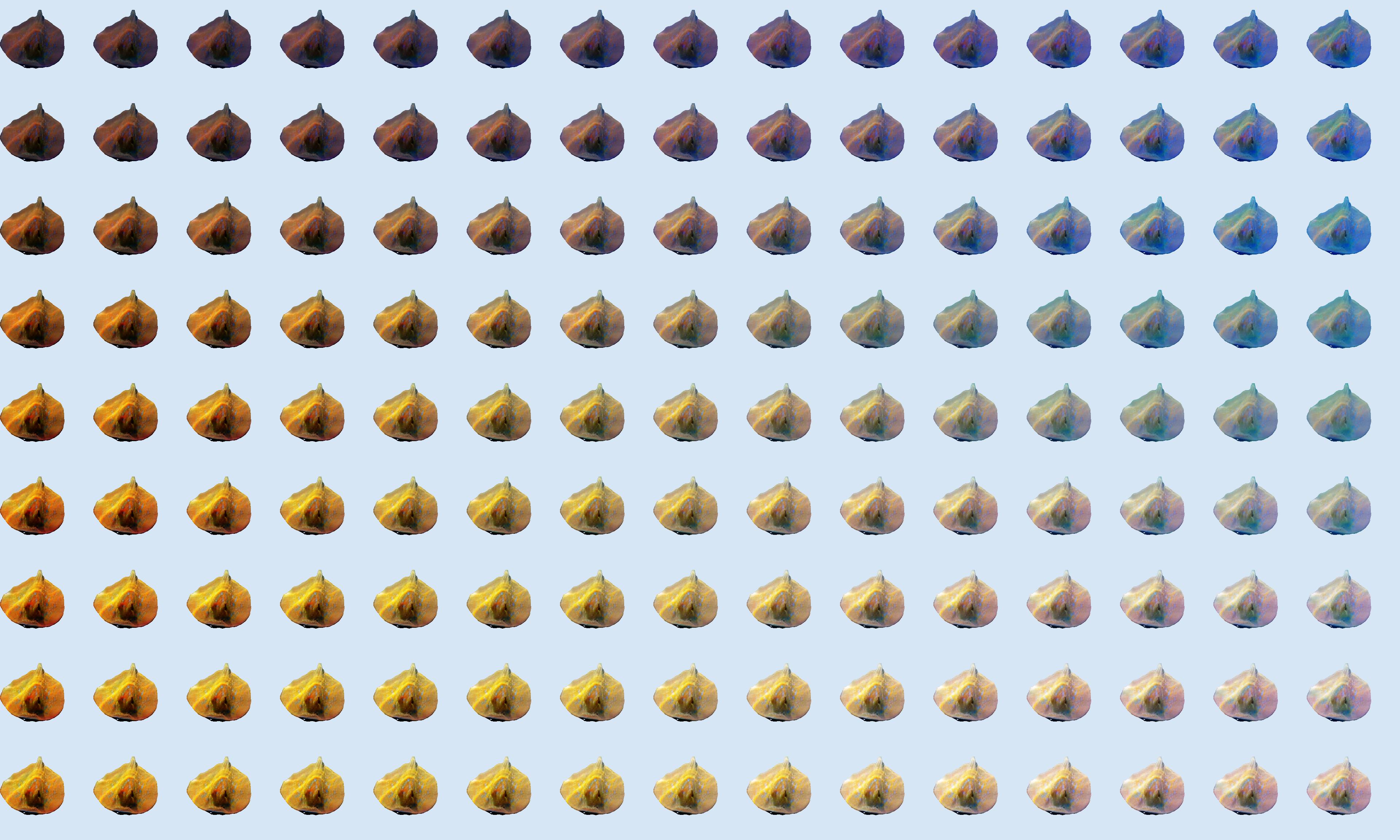}
    \caption{Diversification results obtained by sampling color codes.}
    \label{fig:color_code_interpolation_supp_2}
  \end{figure*}

\begin{figure*}
    \small
    \centering
    \includegraphics[width=18cm,height=10cm]{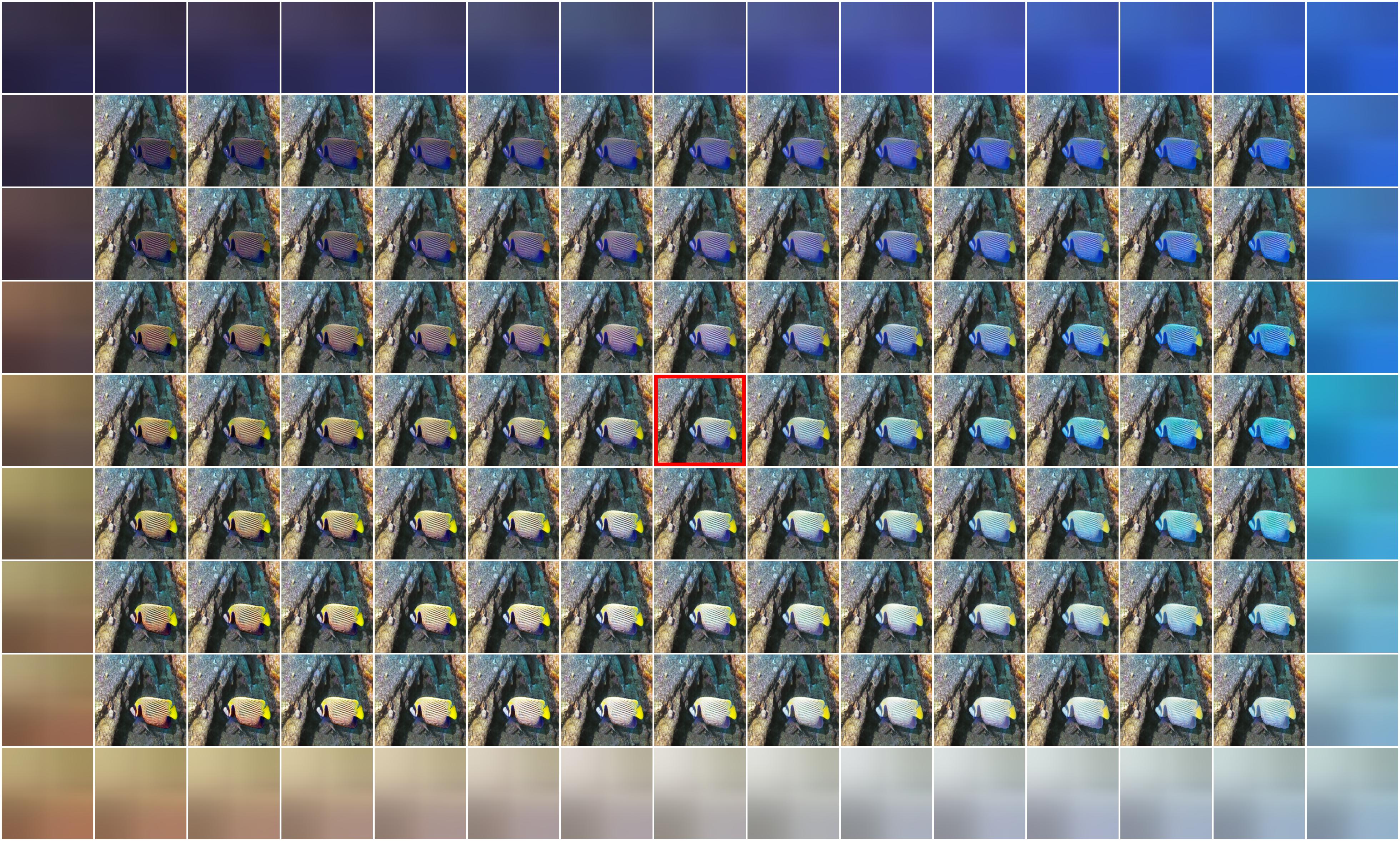}

    \vspace{0.5cm}

    \includegraphics[width=18cm,height=10cm]{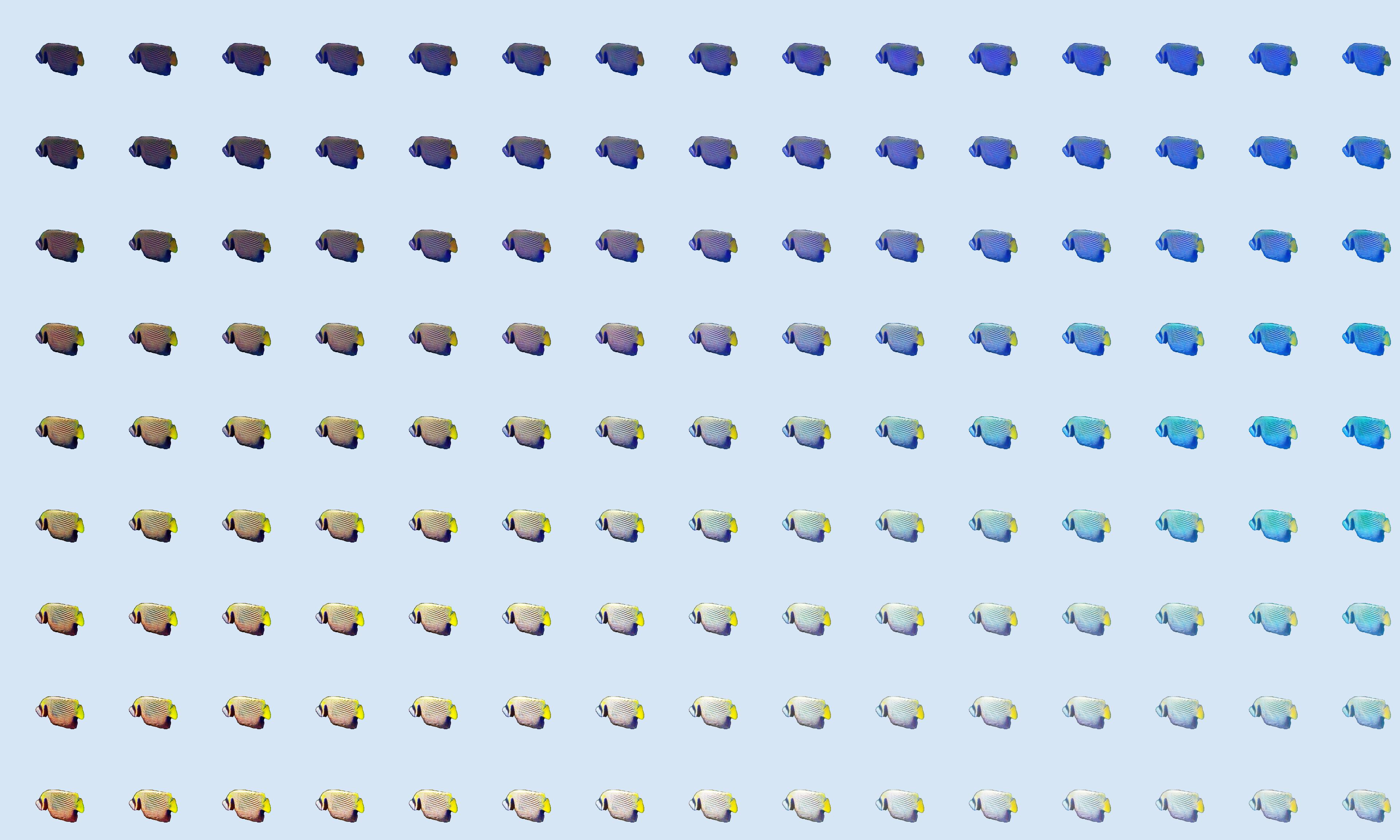}
    \caption{Diversification results obtained by sampling color codes.}
    \label{fig:color_code_interpolation_supp_3}
  \end{figure*}

Fig. \ref{fig:color_code_interpolation_supp_0}, Fig. \ref{fig:color_code_interpolation_supp_1}, Fig. \ref{fig:color_code_interpolation_supp_2} and Fig. \ref{fig:color_code_interpolation_supp_3} show the color interpolation results for multiple underwater organisms. The interpolated image shows that color diversity can be effectively enriched through color code sampling. The color interpolation capability of ColorCode is further verified.

\end{document}